\documentclass[letterpaper, 9.5 pt, journal]{IEEEtran}  

\IEEEoverridecommandlockouts 

\usepackage[numbers]{natbib}
\usepackage{multicol}

\usepackage{graphics} 
\usepackage{times} 
\usepackage{amsmath} 
\usepackage{amssymb} 
\usepackage[svgnames]{xcolor}
\usepackage{graphicx}
\usepackage{booktabs}
\usepackage{tabularx}
\usepackage{amsfonts}
\usepackage{mathtools}
\usepackage{caption}
\usepackage[T1]{fontenc} 
\usepackage{mathtools, cuted} 
\usepackage[ruled,vlined]{algorithm2e} 
\usepackage{enumerate}  
\usepackage{balance} 
\usepackage{tikz}
\usepackage[hidelinks]{hyperref}
\usepackage[T1]{fontenc} 
\usepackage[hyphenbreaks]{breakurl} 
\usepackage{fancyvrb} 
\usepackage{makecell} 

\usepackage{subcaption}
\DeclareCaptionLabelSeparator{periodspace}{.\quad}
\usepackage{amsthm}

\newtheorem{lemma}{Lemma}
\newtheorem{remark}{Remark}
\newtheorem{prop}{Proposition}

\theoremstyle{definition}
\newtheorem{example}{Example}
\newtheorem{problem}{Problem}
\newtheorem{definition}{Definition}

\usepackage{letltxmacro}
\LetLtxMacro\orgvdots\vdots
\LetLtxMacro\orgddots\ddots

\makeatletter
\DeclareRobustCommand\vdots{%
	\mathpalette\@vdots{}%
}
\newcommand*{\@vdots}[2]{%
	\sbox0{$#1\cdotp\cdotp\cdotp\m@th$}%
	\sbox2{$#1.\m@th$}%
	\vbox{%
		\dimen@=\wd0 %
		\advance\dimen@ -3\ht2 %
		\kern.5\dimen@
		\dimen@=\wd2 %
		\advance\dimen@ -\ht2 %
		\dimen2=\wd0 %
		\advance\dimen2 -\dimen@
		\vbox to \dimen2{%
			\offinterlineskip
			\copy2 \vfill\copy2 \vfill\copy2 %
		}%
	}%
}
\DeclareRobustCommand\ddots{%
	\mathinner{%
		\mathpalette\@ddots{}%
		\mkern\thinmuskip
	}%
}
\newcommand*{\@ddots}[2]{%
	\sbox0{$#1\cdotp\cdotp\cdotp\m@th$}%
	\sbox2{$#1.\m@th$}%
	\vbox{%
		\dimen@=\wd0 %
		\advance\dimen@ -3\ht2 %
		\kern.5\dimen@
		\dimen@=\wd2 %
		\advance\dimen@ -\ht2 %
		\dimen2=\wd0 %
		\advance\dimen2 -\dimen@
		\vbox to \dimen2{%
			\offinterlineskip
			\hbox{$#1\mathpunct{.}\m@th$}%
			\vfill
			\hbox{$#1\mathpunct{\kern\wd2}\mathpunct{.}\m@th$}%
			\vfill
			\hbox{$#1\mathpunct{\kern\wd2}\mathpunct{\kern\wd2}\mathpunct{.}\m@th$}%
		}%
	}%
}
\makeatother

\def\bn{\mathbb N}
\def\bz{\mathbb Z}
\def\br{\mathbb R}

\def\bp{\mathbb P}

\def\bv{\mathbb V}
\def\bu{\mathbb U}

\def\sa{\mathcal A}

\def\sv{\mathcal V}
\def\se{\mathcal E}

\def\sg{\mathcal G}

\def\ss{\mathcal S}

\def\tr{\mathrm{tr}}
\def\eig{\mathrm{eig}}

\def\diag{\mathrm{diag}}

\def\Ln{L_{\text{\normalfont nrm}}}
\def\Lnt{\tilde{L}_{\text{\normalfont nrm}}}

\newcommand{\tc}[2][red,fill=red]{\tikz[baseline=-0.5ex]\draw[#1,radius=#2] (0,0) circle ;}%

\newcommand\eqdef{\mathrel{\overset{\makebox[0pt]{\mbox{\normalfont\tiny def}}}{=}}}

\begin{document}

\title{\LARGE \bf
	CLEAR: A Consistent Lifting, Embedding, and Alignment Rectification Algorithm for Multi-View Data Association
}
\author{Kaveh Fathian, Kasra Khosoussi, Yulun Tian,  Parker Lusk,  Jonathan P. How
	\thanks{K.\ Fathian, K.\ Khosousi, Y.\ Tian, P.\ Lusk, and J.\ P.\ How are with
	  the Department of Aeronautics and Astronautics at the Massachusetts Institute of
	  Technology. Email: \{kavehf, kasra, plusk, yulun, jhow\}@mit.edu.}	     
}%

\newcommand{\XX}[1]{{\color{red} {
			\bf XX #1 XX\ }}}

\maketitle

\begin{abstract}
	
Many robotics applications require alignment and
fusion of observations obtained at multiple views 
to form a global model of the environment. Multi-way data
association methods provide a mechanism to improve alignment
accuracy of pairwise associations and ensure their consistency.
However, existing methods that solve this computationally challenging problem
are often too slow for real-time applications. Furthermore, some of the existing techniques can violate
the cycle consistency principle, thus drastically reducing the fusion
accuracy. This work presents the CLEAR (Consistent Lifting,
Embedding, and Alignment Rectification) algorithm to address
these issues. By leveraging insights from the multi-way matching
and spectral graph clustering literature, CLEAR provides cycle
consistent and accurate solutions in a computationally efficient
manner. Numerical experiments on both synthetic and real datasets
are carried out to demonstrate the scalability and superior performance of our algorithm
in real-world problems.
This algorithmic framework can provide significant improvement
in the accuracy and efficiency of existing discrete assignment
problems, which traditionally use pairwise (but potentially inconsistent) correspondences. An implementation of CLEAR is
made publicly available online.
\end{abstract}

\IEEEpeerreviewmaketitle

\section*{Supplementary Material}

CLEAR source code and the code for generating comparison results: 
{{\href{https://github.com/mit-acl/clear}{\color{blue}https://github.com/mit-acl/clear}}}. Video of paper summary:  
{{\href{https://youtu.be/RBxq9KYcgTY}{\color{blue}https://youtu.be/RBxq9KYcgTY}}}.

\section{Introduction}
\begin{figure}[t]
\centering
\includegraphics[trim = 0mm 0mm 0mm 0mm, clip, width=\columnwidth] {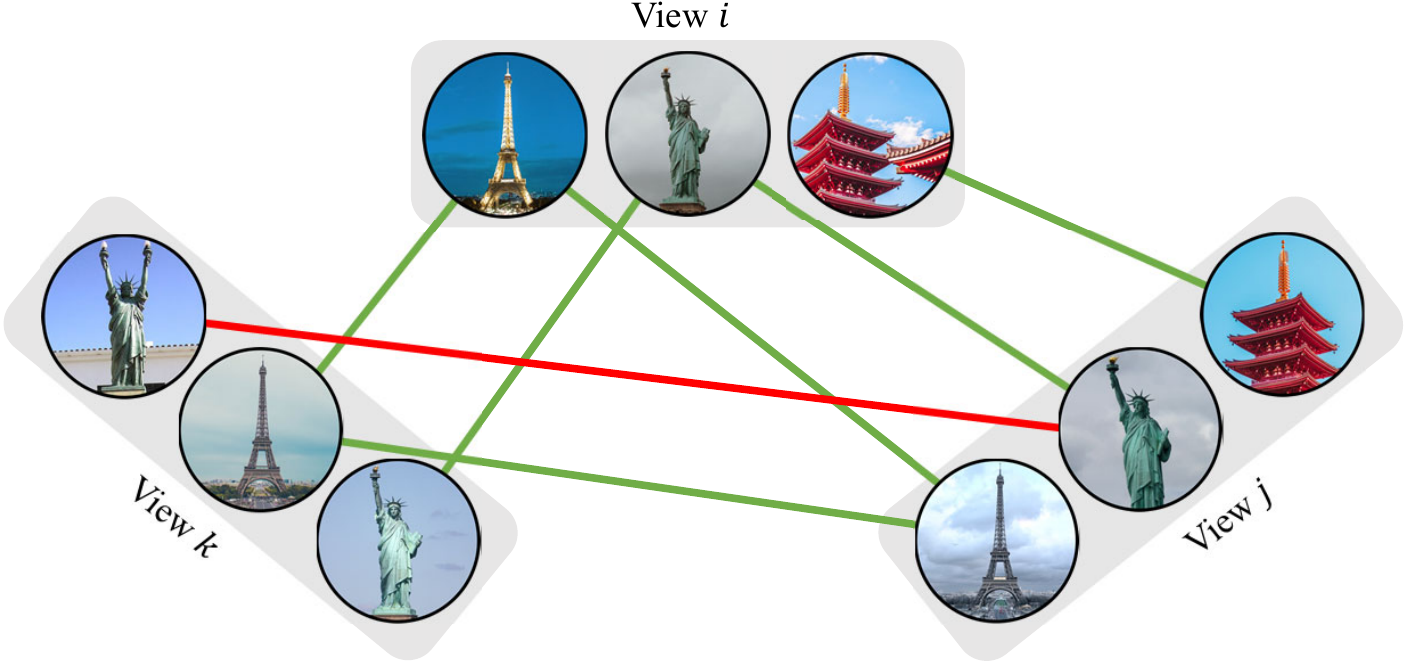}	
\caption{An illustrative example of cycle consistency for the association of images observed in views $i,\, j, \,k$. Associations of ``Eiffel tower'' are cycle consistent. On the other hand, the ``statue of liberty'' associations are inconsistent since the images matched between views $i$ and $j$ and views $i$ and $k$ are not matched between views $j$ and $k$ (i.e., violation of transitivity).}
\label{fig:ExSemantics}
\end{figure}

Data association across \emph{multiple} views, known as the
multi-view or multi-way \cite{Pachauri2013} matching, is a fundamental problem in
robotic perception and computer vision. Conceptually, the goal in this problem is to establish
correct associations between the sightings of ``items'' across multiple ``views'' (see Fig.~\ref{fig:ExSemantics}).
Examples include feature matching across multiple frames
\cite{Pachauri2013,Zhou2015,Tron2017}, and associating landmarks across
multiple maps for map fusion in single/multi-robot simultaneous
localization and mapping (SLAM) \cite{aragues2011consistent}.

The traditional approach treats the multi-view data association problem as a
sequence of decoupled \emph{pairwise} matching subproblems, each of which 
can be formulated and solved, e.g., as a linear
assignment problem \cite{Burkard2012}.
Such techniques, however, cannot leverage
the redundancy in the observations and, furthermore, often result
in \emph{non-transitive} (a.k.a., cycle \emph{inconsistent}) associations; see Fig.~\ref{fig:ExSemantics}.
One can address these issues by \emph{synchronizing} all noisy
pairwise associations via enforcing the cycle consistency constraint.
Cycle consistency serves two crucial purposes: 1) it provides a natural
mechanism for the discovery and correction of wrong (or missing) associations obtained
through pairwise matching; and 2) it establishes an equivalence
relation on the set of observations, which is \emph{necessary} for global fusion in the
so-called clique-centric applications such as map merging (Section~\ref{sec:apps}).
%
%

Synchronizing pairwise associations is a
combinatorial optimization problem with an exponentially large search space.
This problem has been extensively studied in recent years (see \cite{Pachauri2013,Zhou2015,Leonardos2017,Tron2017} and references therein).
These efforts have resulted in several algorithms that can improve the
erroneous initial set of pairwise associations. However, providing solutions that are computationally tractable for real-time applications remains a fundamental challenge.
Further, the rounding techniques used by some of relaxation-based methods may violate
the cycle consistency and \textit{distinctness} constraints (distinctness
implies that the items observed in each view are unique, and thus must not be associated with each other).

This work aims to address these critical challenges via a novel spectral
graph-theoretic approach. Specifically, we leverage the natural graphical representation of the
problem and propose a spectral graph clustering technique uniquely tailored for
producing accurate solutions to the multi-way data association problem in a computationally tractable manner.
Our solutions, by construction, are guaranteed to satisfy the cycle consistency and distinctness constraints under any noise regime. 
These are demonstrated in our extensive empirical evaluations based on synthetic and
real datasets in the context of feature matching and map fusion in
landmark-based SLAM.
%

\subsection{Related Work}

With the exception of combinatorial methods \cite{Zach2010,Nguyen2011} that do
not scale well to large problems, and a recent deep leaning approach in
\cite{Phillips2019}, the majority of permutation synchronization algorithms that aim to solve this computationally challenging problem can be classified
as (i) convex relaxation; (ii) spectral relaxation; and (iii) graph clustering.

Methods in the first category include \cite{Chen2014}, which uses a semidefinite programming relaxation of the problem and solves it via ADMM \cite{Boyd2011}. 
A distributed variation of this method with a similar formulation has been
recently presented in \cite{Hu2018}. Toward the same goal, \cite{Zhou2015} uses
a low-rank matrix factorization to improve the computational complexity. 
Works such as \cite{Yu2016} and \cite{Leonardos2017} require full observability, whereas methods such as \cite{Leonardos2018a} can perform in a partially observable setting, where only a subset of overall items is observed at each view.  
The aforementioned algorithms often return solutions that have the highest
accuracy; however, 
due lifting to high dimensional spaces,
they are slow and not suitable for real-time applications.

Methods in the second category are based on a spectral relaxation of the problem, with prominent works including \cite{Pachauri2013} and \cite{Maset2017}.
The method proposed in \cite{Pachauri2013} returns consistent solutions from noisy pairwise associations using a spectral relaxation in the fully observable setting.
The work done by \cite{Maset2017} proposes an eigendecomposition approach that works in a partially observable setting, however cycle consistency is lost in higher noise regimes.
The recent work of \cite{Bernard2018} leverages a non-negative matrix factorization approach to solve the problem. This method works in a partially observable setting and preserves cycle consistency. 
Algorithms that use spectral relaxation are relatively fast and return solutions that have comparably high accuracy.

Methods in the third category use a graph representation of the problem.  
In \cite{Yan2016} and \cite{Tron2017}, the authors have elegantly observed the equivalence relation between cycle consistency and cluster structure of the association graph.
This observation is used to find approximate solutions to the problem based on existing graph clustering algorithms.
The work done in \cite{Yan2016} has considered a constrained clustering approach using a method similar to $k$-means.
In \cite{Tron2017}, the existing density-based graph clustering algorithm in \cite{Vedaldi2008} is leveraged to solve the problem.
Methods in this category could be very fast, however, the accuracy may be compromised.

Lastly, we point out that the multi-way data association problem can be viewed and solved from a graph matching perspective \cite{Yan2015, Swoboda2019}.
Unlike all previously discussed methods (and the present work), which only leverage the
association information across views, graph matching additionally incorporates
geometrical information between the items in each view. The additional
complexity, in general, results in significantly slower algorithms. 
%

\subsection{Our Contributions}

Our work provides new insights into connections between the multi-way data association problem and the spectral graph clustering literature. 
We leverage these insights to push the boundaries of accuracy and speed---which
are crucial for real-time robotics applications---to solve 
the multi-way data association problem.
The main contributions of this work are as follows:
\begin{enumerate}
	\item To our knowledge, this work provides the first approach that formulates and solves the
	  multi-way association problem using a normalized objective function. This
	  normalization is crucial to recover the correct
	  solution when the association	graph is a mixture of large and small clusters (Remark~\ref{rem:normalizedObj}).
	
	\item We leverage the natural graphical structure of the problem to estimate
	  the unknown universe size\footnote{By definition, universe size is the total number of unique items in all
	  views.} from erroneous associations. Specifically, we prove that our
	  technique is guaranteed to recover the correct universe size under certain
	  bounded noise regimes (Proposition 2). Moreover, we empirically demonstrate that the
	  proposed approach is more robust to noise than the standard eigengap
	  heuristic \cite{VonLuxburg2007}
	  used in the spectral graph clustering literature
	  (Remark~\ref{rem:eigengap}).
	
	\item We propose a projection (rounding) method that, by construction, is guaranteed to
	  produce solutions that satisfy the cycle consistency and distinctness
	  constraints, whereas these constraints can be violated by some of the
	  state-of-the-art algorithms in high-noise regimes
	  (Section~\ref{sec:Simulations}).
\end{enumerate}
In addition, we address an important subtlety regarding the choice of suitable
metrics for evaluating the performance of multi-way matching algorithms in
applications such as map fusion (Section~\ref{sec:apps}). Finally, we provide extensive numerical experiments on both synthetic and
real datasets in the context of feature matching and map fusion
(Sections~\ref{sec:Simulations} and \ref{sec:Experiments}). Our empirical results
demonstrate the superior performance of our algorithm in comparison to the state-of-the-art methods in terms of both accuracy and speed.

\subsection*{Outline}
The organization of the paper is as follows. The notation and definitions are introduced in Section~\ref{sec:Notation}, followed by the problem formulation in Section~\ref{sec:ProblemFormulation}. 
The CLEAR algorithm is presented in Section~\ref{sec:CLEAR}, followed by a numerical example in Section\ref{sec:Example}.
The theoretical justifications behind the algorithm are discussed in Section~\ref{sec:Theory} with proofs presented in the Appendix.
Application domains for the CLEAR algorithm are discussed in Section~\ref{sec:apps}.
CLEAR is benchmarked against the state-of-the-art algorithms using synthetic data in Section~\ref{sec:Simulations}.Finally, experimental evaluations of CLEAR on real-world datasets are presented in Section~\ref{sec:Experiments}.

\section{Notation and Definitions} \label{sec:Notation}

We denote the set of natural numbers by $\bn$, integers by $\bz$, $\bn_0 \eqdef \{0\} \cup \bn$, and define
$\bn_n \eqdef \{1,\, 2,\, \cdots, n \}$.
Scalars and vectors are denoted by lower case (e.g., $a$), matrices by uppercase (e.g., $A$), and sets by script letters (e.g., $\sa$). Cardinality of set $\sa$ is denoted by $|\sa|$.
The element on row $i$ and column $j$ of matrix $A$ is denoted by $(A)_{ij}$.
The Frobenius inner product is defined as $\langle A,B \rangle \eqdef \tr(A^\top
B)$, where $A$ and $B$ are matrices of the same size. Finally, $\| \cdot \|$
denotes the (induced) 2-norm.
Table~\ref{tab:notation} lists the key variables used throughout the paper.

\subsection{Permutation Matrices} \label{sec:}

Matrix $P^i_j \in \{0,1\}^{m_i \times m_j}$ is said to be a \textit{partial permutation} matrix if and only if each row and column of $P^i_j$ \textit{at most} contains a single 1 entry.
Matrix $P$ is called a \textit{full permutation} matrix if and only if each row and column has \textit{exactly} a single 1 entry.
We denote the space of all (partial or full) permutation matrices by $\bp$. 
Matrix $P^i \in \bp$ is said to be a \textit{lifting permutation matrix} if and only if each row of $P^i$ contains a single 1 entry (however, column entries could be all zero). 
We denote the space of all lifting permutation matrices by $\bp^{L}$.
The \textit{aggregate association matrix} consisting of matrices $P^i_j \in \{0,1\}^{m_i \times m_j}$, $i,j \in \bn_n$, is defined as 
%
\begin{gather} \label{eq:P}
P \eqdef \begin{bsmallmatrix}
I     ~& P_2^1  ~& \cdots ~& P_n^1 \\
P_1^2 ~& I      ~& \cdots ~& P_n^2 \\
\vdots~& \vdots ~& \ddots ~& \vdots \\
P_1^n ~& P_2^n  ~& \cdots ~& I 
\end{bsmallmatrix} \, \in \br^{l\times l},
\end{gather}
%
where $I$ is the identity matrix with appropriate size, and ${l \eqdef \sum_{i=1}^{n}{m_i}}$.

\subsection{Graph Theory} \label{sec:}

We denote a graph with $l$ vertices by $\sg = (\sv,\se)$, where $\sv$ is the set
of vertices, and $\se$ is the set of undirected edges. 
The adjacency matrix $A \in \{0,1\}^{l \times l}$ of $\sg$ is defined by $(A)_{ij} = a_{ij}$, where $a_{ij} = 1$ if there is an edge between vertices ${v_i, v_j \in \sv}$, otherwise $a_{ij} = 0$. We assume $a_{ii} = 0$, i.e., graph has no self-loops.
The degree of a vertex $v_i \in \sv$ is defined as ${d_i \eqdef \sum_{j= 1}^{l}{a_{ij}}}$, and the $l \times l$ degree matrix $D$ is defined as a diagonal matrix with $d_1, \dots , d_l$ on the diagonal. We define $C \eqdef D + I$, where $I$ is identity matrix. 
If $c_i$'s denote the diagonal entries of $C$, then $C^{-\frac{1}{2}}$ is a diagonal matrix with diagonal entries $1/\sqrt{c_i}$. 
The Laplacian matrix of $\sg$ is defined as $L \eqdef D - A$.
A \textit{cluster graph} $\sg$ is a disjoint union of cliques (i.e., complete subgraphs).
That is, $\sg$ can be partitioned into subgraphs $\sa_1, \sa_2, \dots, \sa_m$, where each $\sa_i$ is a complete graph and there is no edge between any two $\sa_i,\, \sa_j$. The cliques in a cluster graph are called clusters.


\begin{table} [t] 
\renewcommand{\arraystretch}{1.5}
\caption{Summary of important nomenclature used throughout the paper.}
\begin{tabular}{ p{0.8cm}  p{1.5cm}  p{5.2cm}  } 
    \Xhline{1pt} 
	Notation & Domain &  Definition and properties \\
	\Xhline{1pt} 
	$n$ 	& $\bn$ & Total number of views \\ \hline
	$m$		& $\bn$ & Size of universe; number of unique items; number of cliques in the association graph\\ \hline
	$m_i$   & $\bn$ & Number of items observed in view $i$ \\ \hline 
	$l$		& $\bn$ & Total number of items observed across all views; $l \eqdef \sum_i m_i$ \\ \hline
	\textasciitilde	& - & Accent used for variables corresponding to the noisy input \\ \Xhline{1.0pt}

	$P^i_j$ & $\{0,1\}^{m_i \times m_j}$ &Partial permutation matrix; association matrix between items at views $i$ and $j$ \\ \hline
	$P$ 	& $\{0,1\}^{l \times l}$ & Aggregate association matrix consisting of $P^i_j$'s; see \eqref{eq:P} \\ \hline	
	
	$A$ 	& $\{0,1\}^{l \times l}$ & Adjacency matrix of the association graph; ${A = P - I}$ \\ \hline
	$D$ 	& $\bn_0^{l \times l}$ & Degree matrix of the association graph\\ \hline	
	$C$ 	& $\bn_0^{l \times l}$ & Diagonal matrix with entries $c_i \eqdef \sum_j (P)_{ij}$; $C = D + I$ \\ \hline
	$L$		& $\bz^{l \times l}$ & Laplacian matrix of $\sg$; $L \eqdef D - A = C - P$ \\ \hline  
	$L_{\text{nrm}}$ & $\br^{l \times l}$ & Normalized Laplacian matrix; ${L_{\text{nrm}} \eqdef C^{-\frac{1}{2}} \, L \, C^{-\frac{1}{2}}}$ \\ \hline	
	$P_{\text{nrm}}$ & $\br^{l \times l}$  & Normalized association matrix; ${P_{\text{nrm}} \eqdef C^{-\frac{1}{2}} \, P \, C^{-\frac{1}{2}}}$ \\ \Xhline{1.0pt}

	$P^i$   & $\{0,1\}^{m_i \times m}$ & Lifting permutation matrix; association of items observed at views $i$ to items of the universe \\ \hline	
	$V$ 	& $\{0,1\}^{l \times m}$ & Aggregate lifting permutation matrix consisting of $P^i$'s; see \eqref{eq:V} \\ \hline	
	$U$		& $\br^{l \times m}$ & Normalized aggregate lifting permutation; $U \eqdef C^{-\frac{1}{2}} \, V$; eigenvectors  associated to $m$ smallest eigenvalues of $\Ln$\\ \hline
	$u_i$	& $\br^m$ & Row of $U$ \\ \hline
	$u'_i$	& $\br^m$ & Pivot row of $U$ \\ \Xhline{1.0pt}
\end{tabular}
\label{tab:notation}
\end{table}
\section{Problem Formulation} \label{sec:ProblemFormulation}

Simply put, the objective of this paper is to reconstruct a set of \textit{cycle consistent} associations from a set of pairwise associations, which may contain error and lack cycle consistency. 
This problem can be approached from either an optimization or a graph-theoretic viewpoint.
In what follows, we will first describe each formulation separately, and then
shed light on their connections.

\subsection{Optimization-Based Formulation} \label{sec:}

\begin{figure}[t]
	\centering
	\includegraphics[trim = 0mm 0mm 0mm 0mm, clip, width=0.45\textwidth] {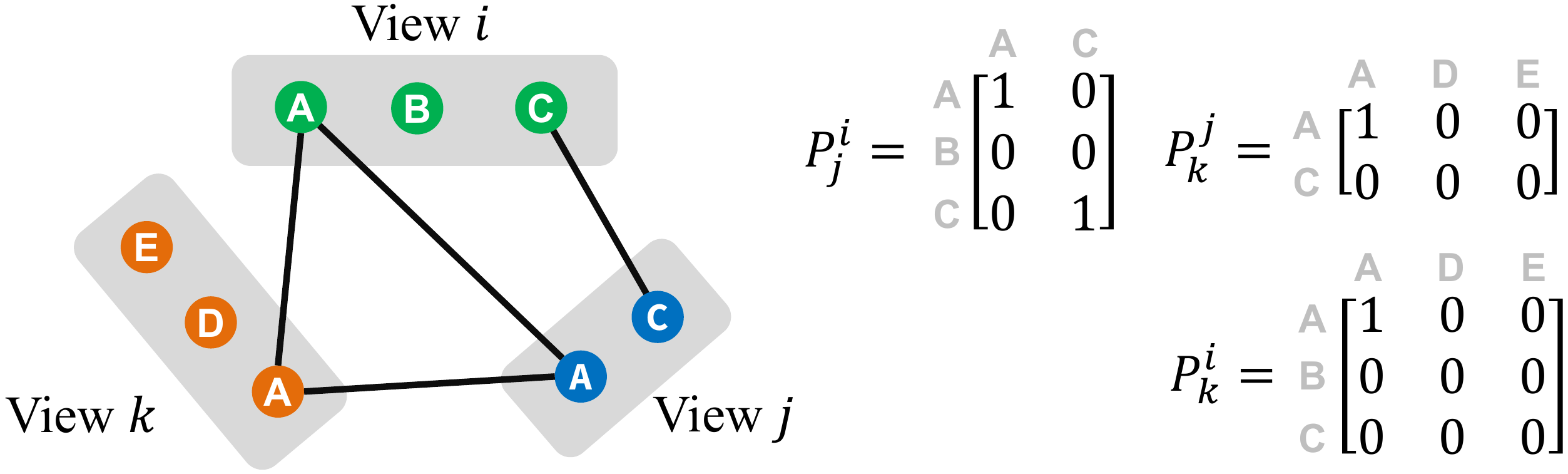}	
	\caption{Association of items labeled as A, B,..., E observed at three views.}
	\label{fig:ExPij}
\end{figure}

\begin{figure}[t]
	\centering
	\includegraphics[trim = 0mm 0mm 0mm 0mm, clip, width=0.45\textwidth] {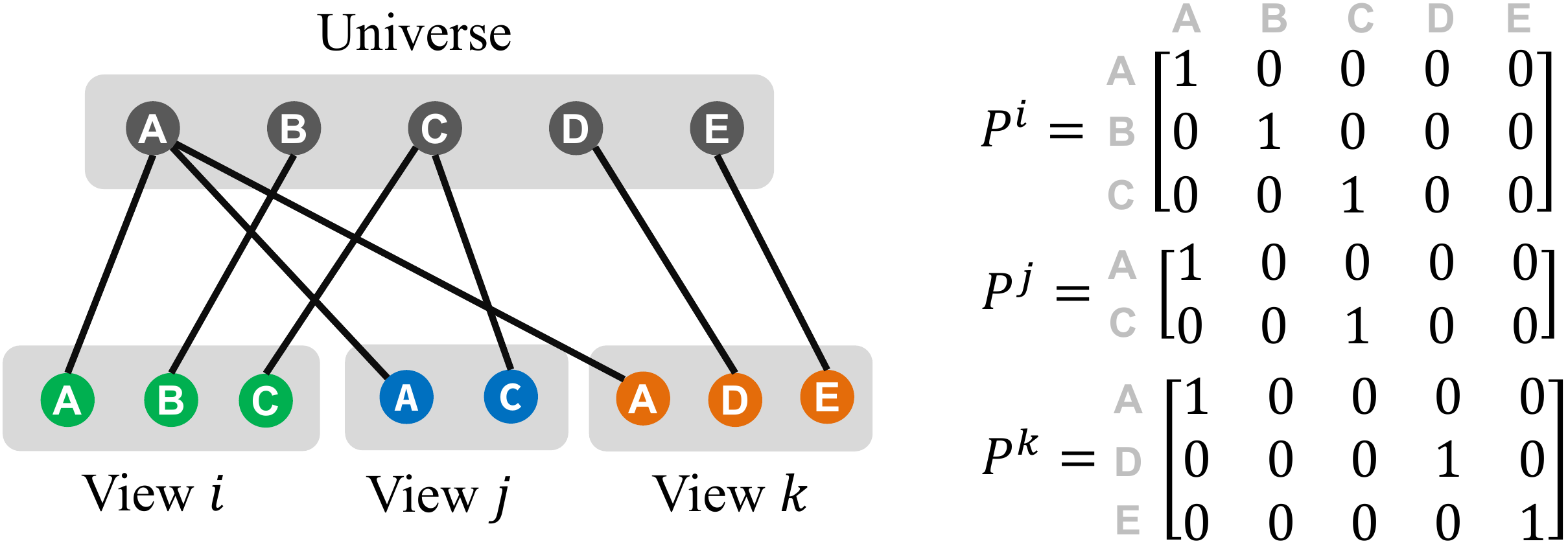}	
	\caption{Lifting permutation matrices associating observations at views $i,\, j,\, k$ to the universe, which consists of items labeled as A, B, ..., E.}
	\label{fig:ExPj}
\end{figure}

We consider $n$ views and assume that view $i$ contains $m_i$ items.
Associations (or matchings) between items in views $i$ and $j$ can be represented by a binary matrix ${P^i_j \in \{0,1\}^{m_i \times m_j}}$, in which the one entries indicate the associations.
An example of pairwise associations among three views is shown in Fig.~\ref{fig:ExPij}.
A lifting permutation represents the association between items observed in a view and the universe (which by definition consists of all items). An example is provided in Fig.~\ref{fig:ExPj}.

\begin{definition} \label{def:CycleCons}
Pairwise associations $P^i_j$ are \textit{cycle consistent} if and only if there exist lifting permutations ${P^i \in \bp^L}$ such that 
%
\begin{gather} \label{eq:Consistency}
P^i_j = P^i \, P^{j\,\top}, \qquad  \forall i,j \in \bn_n.
\end{gather}
%
%
\end{definition}

The cycle consistency condition \eqref{eq:Consistency} can be presented more concisely as $P = V \, V^\top$, where $P$ is the aggregate association matrix defined in \eqref{eq:P}, and 
%
\begin{equation} \label{eq:V}
V \eqdef \begin{bmatrix}
P^{1\, \top} & P^{2\, \top} & \cdots & P^{n\, \top} 
\end{bmatrix}^\top \in \{0,1\}^{l\times m},
\end{equation}
%
where $l \eqdef \sum_i{m_i}$. Here, $m \in \bn_l$ is the number of columns of lifting permutations that is referred to
as the \textit{size of universe}.

Throughout this paper, we use the accent \textasciitilde\ to distinguish the variables that are associated with the noisy input. 
Therefore, ${\tilde{P}^i_j \in \{0, 1\}^{m_i \times m_j}}$ denotes the noisy association between items in views $i$ and $j$, where $\tilde{P}^i_i = I$ by definition. Note that $\tilde{P}^i_j$'s can be erroneous and inconsistent.
Let $\tilde{P} \in \br^{l\times l}$, defined via~\eqref{eq:P}, denote the noisy aggregate association matrix. Further, let $\tilde{C}$ be an $l\times l$ diagonal matrix with positive diagonal entries $\tilde{c}_1,\dots, \tilde{c}_l$ defined as the sum of corresponding rows of $\tilde{P}$, i.e.,  $\tilde{c}_i \eqdef \sum_{j = 1}^{l}{(\tilde{P})_{ij}}$.
Using definitions above, we now formulate the main problem.

\begin{problem} \label{prob:Main}
Given noisy associations $\tilde{P}^i_j$, find cycle consistent associations $P^i_j$ that solve the program 
%
\begin{gather} \label{eq:optim}
\begin{array}{ll}
\underset{P^i_j \in \bp}{\text{maximize}} & ~ \langle P_{\text{nrm}},\, \tilde{P}_{\text{nrm}} \rangle  \\
\text{subject to}   & ~ P = V \, V^\top,
\end{array}
\end{gather}
%
where $P_{\text{nrm}} \eqdef C^{-\frac{1}{2}} \, P \, C^{-\frac{1}{2}}$, $\tilde{P}_{\text{nrm}} \eqdef \tilde{C}^{-\frac{1}{2}} \, \tilde{P} \, \tilde{C}^{-\frac{1}{2}}$.
\end{problem}

In Problem~\ref{prob:Main}, diagonal matrices $C$ and $\tilde{C}$ are used to normalize the aggregate association matrices. 
The justification behind this normalization will be explained in Remark~\ref{rem:normalizedObj} after the graph formulation of the problem is introduced.
The constraint $P^i_j \in \bp$ enforces the permutation structure, preventing the rows and columns of $P^i_j$ from having more than a single one entry. This enforces the \textit{distinctness constraint}, which implies that items in the same view are unique, thus should not be associated with each other. 
The constraint $P = V \, V^\top$ imposes cycle consistency, capturing the fact
that correct associations should be transitive (i.e., if item $i$ is associated to item $j$, and item $j$ is associated to item $k$, then item $i$ must also be associated to item $k$).

\subsection{Graph-Based Formulation} \label{sec:GraphFrom}

The problem of data association has a graph representation. This representation provides the key insights that are leveraged by our algorithm to improve accuracy and runtime.
A set of pairwise associations $P^i_j$ can be represented as a colored graph, where items in each view are denoted by vertices with identical color, and each nonzero entry of $P^i_j$ represents an edge between the corresponding vertices (e.g.,  Fig.~\ref{fig:ExPij}). 
Formally, an \textit{association graph} is defined as $\sg = (\sv,\se)$ with the coloring map $g: \sv \rightarrow \bn_n$.
The set of vertices $\sv$ consists of subsets $\sv_i$ corresponding to items in view $i$, where $g(v_j) \eqdef i$ for all $v_j \in \sv_i$. The set of edges $\se$ consists of subsets  $\se_{ij}$, $i \neq j \in \bn_n$, corresponding to associations, where $\{v_r, v_s\} \in \se_{ij}$ if and only if $(P^i_j)_{rs} = 1$.

The variables $P,\, C$ and $V$ defined previously in the optimization formulation \eqref{eq:optim} also have graph interpretations. Specifically, the adjacency matrix of the association graph $\sg$ is given by $A = P - I$. Further, we have that $C = D + I$, where $D$ is the degree matrix of the graph. To understand the graph interpretation of $V$, we first note the following relation between the cycle consistency and the graph representation.

\begin{prop} \label{prop:CycleCons}
A set of pairwise associations is cycle consistent if and only if the corresponding association graph is a cluster graph (i.e., a disjoint union of complete subgraphs).
\end{prop}

Proof of Proposition~\ref{prop:CycleCons} is given in \cite[Prop. 2]{Tron2017} and hence omitted here. The proof reveals the connection between the algebraic definition of cycle consistency, $P = V \, V^\top$, and clusters of the association graph, denoted by $\sa_1, \dots, \sa_m$.  
In particular, row $i$ of the aggregate lifting permutation matrix $V \in \{0,1\}^{l\times m}$ represents vertex $v_i$ of the association graph. The one entries in $j$-th column of $V$ indicate the vertices that belong to cluster $\sa_j$ of $\sg$. 
That is, if $(V)_{ij} = (V)_{kj} = 1$, then vertices $v_i$ and $v_k$ are connected by an edge and belong to cluster $\sa_j$. 
We will leverage this observation in the theoretical analysis of the algorithm.

\begin{figure*}[t!]
	\centering
	\begin{subfigure}[b]{0.85\textwidth} \includegraphics[trim = 0mm 0mm 0mm 0mm, clip, width=1.0\textwidth] {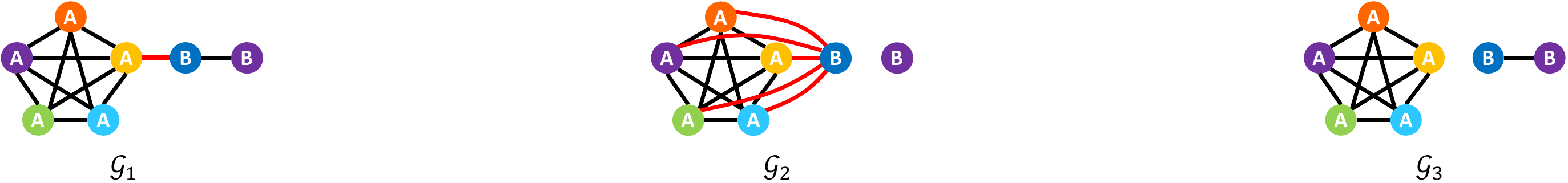}
	\end{subfigure}%
	\caption{(Best viewed in color) Graph $\sg_1$ indicates the association of two items labeled as A, B, in six views identified by colors \tc[white, fill=violet]{3pt}, \tc[white, fill=blue]{3pt}, \tc[white, fill=Goldenrod]{3pt}, \tc[white, fill=Turquoise]{3pt}, \tc[white, fill=LimeGreen]{3pt}, \tc[white, fill=orange]{3pt}.
		The incorrect association, which connects A and B, is indicated by the red edge.	
		If in \eqref{eq:optim} the unnormalized objective $\langle P,\, \tilde{P} \rangle$ is used instead, $\sg_2$ (and also $\sg_3$) would be the optimal solution (with optimal values of $29$).
		On the other hand, by using the proposed normalized objective $\langle P_{\text{nrm}},\, \tilde{P}_{\text{nrm}} \rangle$, the correct association graph $\sg_3$ is the only optimal solution (with optimal value of $1.79$; the value for $\sg_2$ is $1.43$).
	}
	\label{fig:GraphExample}	
\end{figure*}

Given a noisy association graph $\tilde{\sg}$ with adjacency matrix $\tilde{A}$, degree matrix  $\tilde{D}$, and $\tilde{C} = \tilde{D}+I$, the graph-based formulation of the multi-way association problem is as follows.

\begin{problem} \label{prob:Main2}
Given the noisy association graph $\tilde{\sg}$, find undirected graph $\sg$ with adjacency matrix $A$ that solves 
%
\begin{gather} \label{eq:optim2}
%
\begin{array}{ll}
\underset{A}{\text{maximize}} &  ~ \langle A_{\text{nrm}},\, \tilde{A}_{\text{nrm}} \rangle  \\
\text{subject to}   & \sg  \text{~consists of clusters~} \sa_1, \dots, \sa_m  \\ 
& g(v_i) \neq g(v_j), \quad \forall v_i, v_j \in \sa_k \\
\end{array}
\end{gather}
%
where $A_{\text{nrm}} \eqdef C^{-\frac{1}{2}} \, A \, C^{-\frac{1}{2}}$ and $\tilde{A}_{\text{nrm}} \eqdef \tilde{C}^{-\frac{1}{2}} \, \tilde{A} \, \tilde{C}^{-\frac{1}{2}}$.
\end{problem}

Note that Problems~\ref{prob:Main} and \ref{prob:Main2} are equivalent. As
elaborated above, the indices of the vertices belonging to clusters  $\sa_1,\, \dots, \sa_m$ uniquely determine $V$ in Problem~\ref{prob:Main}. Further, since $A = P - I$, both objective functions have the same optimizer. 
In \eqref{eq:optim2}, the first two constraints respectively correspond to the
cycle consistency and distinctness of associations, where the latter is achieved
by the fact that the colors of vertices in each cluster must be distinct.
%

\begin{remark} \label{rem:normalizedObj}
The normalized objective function in \eqref{eq:optim} is a key distinction from several state-of-the-art methods \cite{Pachauri2013, Maset2017, Bernard2018} that consider the unnormalized objective $\langle P,\, \tilde{P} \rangle$. 
By weighting edges based on the degrees of their adjacent vertices, the normalized objective provides a measure to ``balance'' the number of edges that are removed from or added to the noisy association graph $\tilde{\sg}$ to obtain $\sg$. The unnormalized objective, on the other hand, is indifferent to the number of added edges. This can lead to (undesired) optimal solutions that consist of many additional edges.
This point is illustrated in Fig.~\ref{fig:GraphExample}, where, in contrast to the normalized objective, the optimal solution with an unnormalized objective could fail to recover the ground truth even in a relatively low-noise regime.		
\end{remark}

We point out that the example shown in Fig.~\ref{fig:GraphExample} is only one of countless scenarios in which the optimal solution of an \textit{unnormalized} objective could fail to recover the desired association in a low-noise regime. Such examples can be constructed by (wrongly) associating clusters with small and large number of vertices.

\section{The Consistent Lifting, Embedding, and Alignment Rectification (CLEAR) Algorithm} \label{sec:CLEAR}

In this section, we present a concise description of the CLEAR algorithm used for solving the permutation synchronization problem, followed by a numerical example to further illustrate the steps of the algorithm in the next section. Theoretical justifications of the algorithm will be discussed in Section~\ref{sec:Theory}.
The pseudocode of  CLEAR is given in Algorithm~\ref{alg:CLEAR}, where each step is discussed in details below.

\vspace{0.2cm}
\noindent $\bullet$ \textbf{Step 1}: 
Let $\tilde{\sg}$ denote the association graph corresponding to a set of noisy pairwise associations $\tilde{P}^i_j$. Define the normalized Laplacian of $\tilde{\sg}$ as 
%
\begin{gather} \label{eq:Lhat2}
\Lnt \eqdef \tilde{C}^{-\frac{1}{2}} \, \tilde{L} \, \tilde{C}^{-\frac{1}{2}},
\end{gather}
%
where $\tilde{L} = \tilde{D} - \tilde{A}$, $\tilde{C} \eqdef \tilde{D} + I$, and $\tilde{D},\, \tilde{A}$ are respectively the degree and adjacency matrix of $\tilde{\sg}$. 
Compute the eigenvalues and eigenvectors of $\Lnt$.

To reduce the computational complexity, eigendecomposition of $\Lnt$ is done by first finding the connected components of $\tilde{\sg}$ using the breadth-first search (BFS) algorithm \cite{Skiena1998}. 
Eigenvalues of $\Lnt$ are then given as the disjoint union of each component's normalized Laplacian eigenvalues. Similarly, eigenvectors are given by
appropriately padding the eigenvectors of connected components with zeros. 

We point out that if $\Lnt$ is not symmetric, its symmetric component $(\Lnt + \Lnt^\top)/2$ should be used instead in the eigendecomposition (the skew-symmetric component does not contribute to the optimal answer; see Remark~\ref{rem:symmetry}). Note that the symmetry implies that all eigenvalues and eigenvectors are real.

\vspace{0.2cm}
\noindent $\bullet$ \textbf{Step 2}: 
Obtain an estimate for the \textit{size of universe} as
%
\begin{equation} \label{eq:mEstimate}
\hat{m} \eqdef \max \,  \{ \tilde{m},\, m_1,\, m_2, \dots, m_n\},
\end{equation}
%
where $m_i$ is the number of items in view $i$, and $\tilde{m}$ is defined as
%
\begin{equation} \label{eq:mEstimate2}
\tilde{m} \eqdef \left| \{\lambda \in \eig(\Lnt) \,:\, \lambda < 0.5  \} \right|,
\end{equation}
%
i.e., the number of eigenvalues of $\Lnt$ that are less than $0.5$.

\vspace{0.2cm}
\noindent $\bullet$ \textbf{Step 3}: 
Define matrix $U \in \br^{l\times m}$ as the $\hat{m}$ first eigenvectors of $\Lnt$, that is, the eigenvectors associated with the smallest eigenvalues.

\vspace{0.2cm}
\noindent $\bullet$
\textbf{Step 4}: 
Normalize rows of $U$ to have unit norm, i.e., the $i$-th row of $U$, denoted by $u_i$, is replaced by $u_i/\|u_i\|$.
Choose the $\hat{m}$ most orthogonal rows as \textit{pivots}.                     

This can be done based on a greedy strategy where the first row of $U$ is chosen as the first pivot. To find the remaining pivots, the row with the smallest inner product magnitude with previously chosen pivots is picked consecutively.
That is, if $u'_k$ denotes the $k$-th pivot, $u'_{k+1}$ is selected such that $\sum_{i=1}^k{| \langle u'_i, u'_{k+1} \rangle |}$ is minimized.

For each view $i$, define matrix ${F^i \in \br^{m_i \times m}}$ by ${(F^i)_{jk} \eqdef \| u_j - u'_k \|^2}$, where $u_j$ denotes the rows of $U$ associated to items in view $i$, and $u'_k$ denotes the pivot rows.\footnote{Specifically, $u_j$ denotes rows $\sum_{k = 1}^{i-1} m_k + 1$ through $\sum_{k = 1}^{i} m_k$ of $U$.}
Solve a linear assignment problem based on $F^i$ as the cost matrix to obtain a lifting permutation $P^i \in \bp^L$ that associates the items in view $i$ (rows $u_j$) to the items of the universe (pivot rows $u'_k$). 
The Hungarian algorithm~\cite{Burkard2012} can be used to solve the linear
assignment problem and find the optimal answer. However, to reduce the
computational complexity, faster (suboptimal) algorithms can be used instead
while the distinctness constraint is preserved by ensuring that 
each $u'_k$ is associated to at most one $u_j$,
and each $u_j$ is associated to exactly one $u'_k$.

\vspace{0.2cm}
\noindent $\bullet$
\textbf{Step 5}:
Compute pairwise associations as $P^i_j = P^i P^{j\,\top}$.
From Definition~\ref{def:CycleCons}, pairwise associations are cycle consistent by construction.

\begin{algorithm}[t] 
%
\DontPrintSemicolon 
\SetKwData{Left}{left}\SetKwData{This}{this}\SetKwData{Up}{up}
\SetKwFunction{Union}{Union}\SetKwFunction{FindCompress}{FindCompress}
\SetKwInOut{Input}{Input}\SetKwInOut{Output}{Output}

\SetKwInput{StepA}{$\bullet$ Step 1}
\SetKwInput{StepB}{$\bullet$ Step 2}
\SetKwInput{StepC}{$\bullet$ Step 3}
\SetKwInput{StepD}{$\bullet$ Step 4}
\SetKwInput{StepE}{$\bullet$ Step 5}
\SetKwInput{StepF}{Step 6}
\SetKwInput{StepG}{step 7}
\SetKwInput{StepH}{step 8}
\SetKwInput{Notation}{notation}

\caption{CLEAR (pseudocode)}

\Input{Noisy pairwise associations $\tilde{P}^i_j$.}
\Output{Cycle consistent associations $P^i_j$.} 
\BlankLine

\BlankLine

\StepA{Compute $\Lnt$ from \eqref{eq:Lhat2} and find its eigendecomposition.} 
\StepB{Estimate size of universe $\hat{m}$ from \eqref{eq:mEstimate}.}
\StepC{Set $U$ as the $\hat{m}$ first eigenvectors of $\Lnt$.}
\StepD{Normalize rows of $U$ and chose $\hat{m}$ most orthogonal rows as pivots. Find lifting permutations $P^i$ by assigning rows to pivots based on distance.}
\StepE{Set $P^i_j \leftarrow P^i P^{j\,\top}$.}
%
%
\label{alg:CLEAR}
\end{algorithm}


\section{Numerical Example} \label{sec:Example}

We present an example to illustrate the steps of the CLEAR algorithm and show how pivot rows are chosen. 

\begin{example} \label{ex:graph}
In this example, we use the CLEAR algorithm to recover cycle-consistent associations from the (noisy) association graph $\tilde{\sg}$ shown in Fig.~\ref{fig:example1}.
Note that $\tilde{\sg}$ is identical to $\sg_1$ in Fig.~\ref{fig:GraphExample}, where the correct associations and the labels $A, B$ are unknown and should be recovered. 
The aggregate association matrix (which is equal to the adjacency matrix plus identity) is given by
%
\begin{gather} \label{eq:Ptilde}
\tilde{P} = \begin{bsmallmatrix}
1 &    0 &    1  &  0 &   0  &   0  &  0 \\
0 &    1 &    0  &  1 &   1  &   1  &  1 \\
1 &    0 &    1  &  1 &   0  &   0  &  0 \\
0 &    1 &    1  &  1 &   1  &   1  &  1 \\
0 &    1 &    0  &  1 &   1  &   1  &  1 \\ 
0 &    1 &    0  &  1 &   1  &   1  &  1 \\
0 &    1 &    0  &  1 &   1  &   1  &  1 \\
\end{bsmallmatrix}.
\end{gather}
%
The first two rows of $P$ correspond to items in view 1 and the remaining rows successively correspond to views 2 through 6.

\begin{figure} [t]
	\centering
	\includegraphics[trim = 0mm 0mm 0mm 0mm, clip, width=0.15\textwidth] {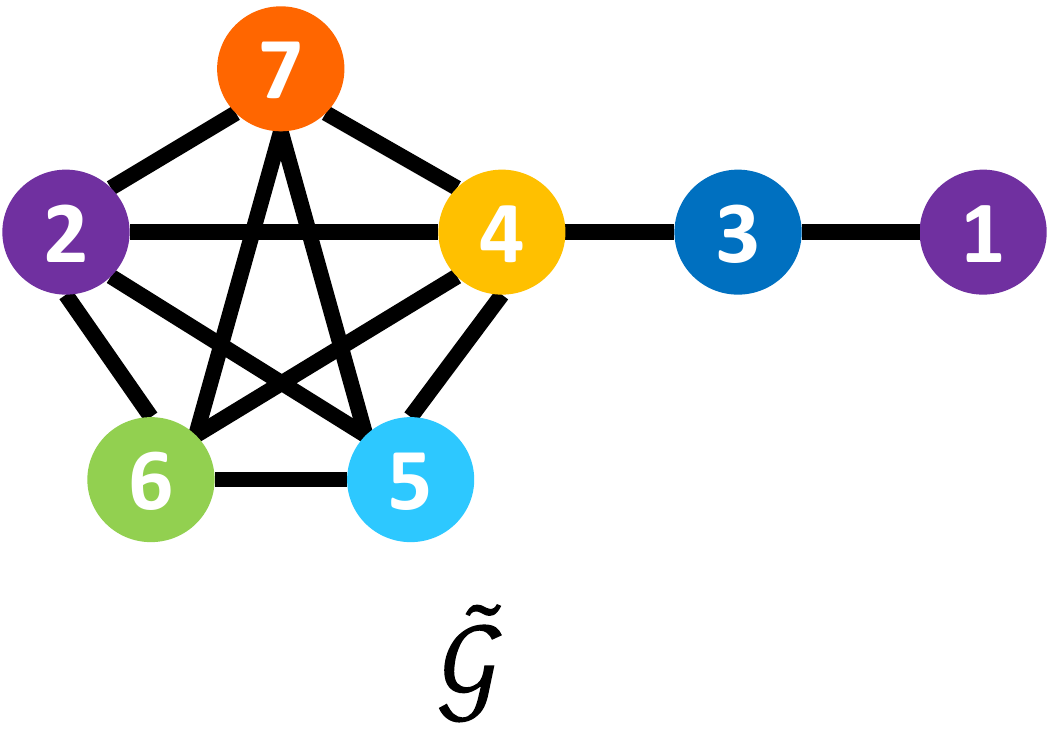}
	\caption{The association graph corresponding to observations in six views distinguished by color. View 1 is colored as \tc[white, fill=violet]{3pt}, and views 2 through 6 are successively colored as  \tc[white, fill=blue]{3pt}, \tc[white, fill=Goldenrod]{3pt}, \tc[white, fill=Turquoise]{3pt}, \tc[white, fill=LimeGreen]{3pt}, \tc[white, fill=orange]{3pt}. Vertices are numbered from 1 to 7.}
	\label{fig:example1}
\end{figure}

\vspace{0.2cm}
\noindent $\bullet$ \textbf{Step 1}:
From \eqref{eq:Ptilde}, the Laplacian matrix is computed as $\tilde{L} = \tilde{C} - \tilde{P}$, where $\tilde{C} = \diag(2,5,3,6,5,5,5)$ and $\diag$ creates a diagonal matrix from input arguments. The normalized Laplacian matrix is given by $\Lnt = \tilde{C}^{-\frac{1}{2}} \, \tilde{L} \, \tilde{C}^{-\frac{1}{2}}$, which has eigenvalues $\{ 1.18,\, 1,\, 1,\, 1,\, 0.85,\, 0.17,\, 0 \}$.

\vspace{0.2cm}
\noindent $\bullet$ \textbf{Step 2}: 
The number of eigenvalues of $\Lnt$ that are less than $0.5$ are two. Hence, $\tilde{m} = 2$. The number of items in views is either two (for view 1) or one (for the rest of views). Thus, the estimated size of universe is obtained as $\hat{m} = 2$.

\vspace{0.2cm}
\noindent $\bullet$ \textbf{Step 3}: 
Matrix $U$ consisting of the first two eigenvectors of $\Lnt$ is computed. 

\vspace{0.2cm}
\noindent $\bullet$ \textbf{Step 4}: 
Rows of $U$ are normalized to obtain (up to two decimals)
%
\begin{gather} \label{eq:Uexample}
U = \begin{bsmallmatrix}
\mbox{-}0.94 &~& 0.34 \\ 
~0.47 & & 0.88 \\ 
\mbox{-}0.88 &~ & 0.48 \\ 
~0.07 &~ & 0.99 \\ 
~0.47 &~ & 0.88 \\  
~0.47 &~ & 0.88 \\ 
~0.47 &~ & 0.88 \\ 
\end{bsmallmatrix}.
\end{gather}
%
Fig.~\ref{fig:points1} depicts rows of \eqref{eq:Uexample} as vectors, where the endpoint of each vector is colored based on the view that it corresponds to, and the unit circle is drawn to indicate that rows have unit norm.
The pivots are chosen by taking the first row as the first pivot $u'_1 = [-0.94, \, 0.34]$. The second pivot is chosen as the row of $U$ that has the smallest (absolute value of) inner product with $u'_1$, which gives $u'_2 = [0.47, \, 0.88]$.

\begin{figure} [t]
	\centering
	\includegraphics[trim = 0mm 0mm 0mm 0mm, clip, width=0.4\textwidth] {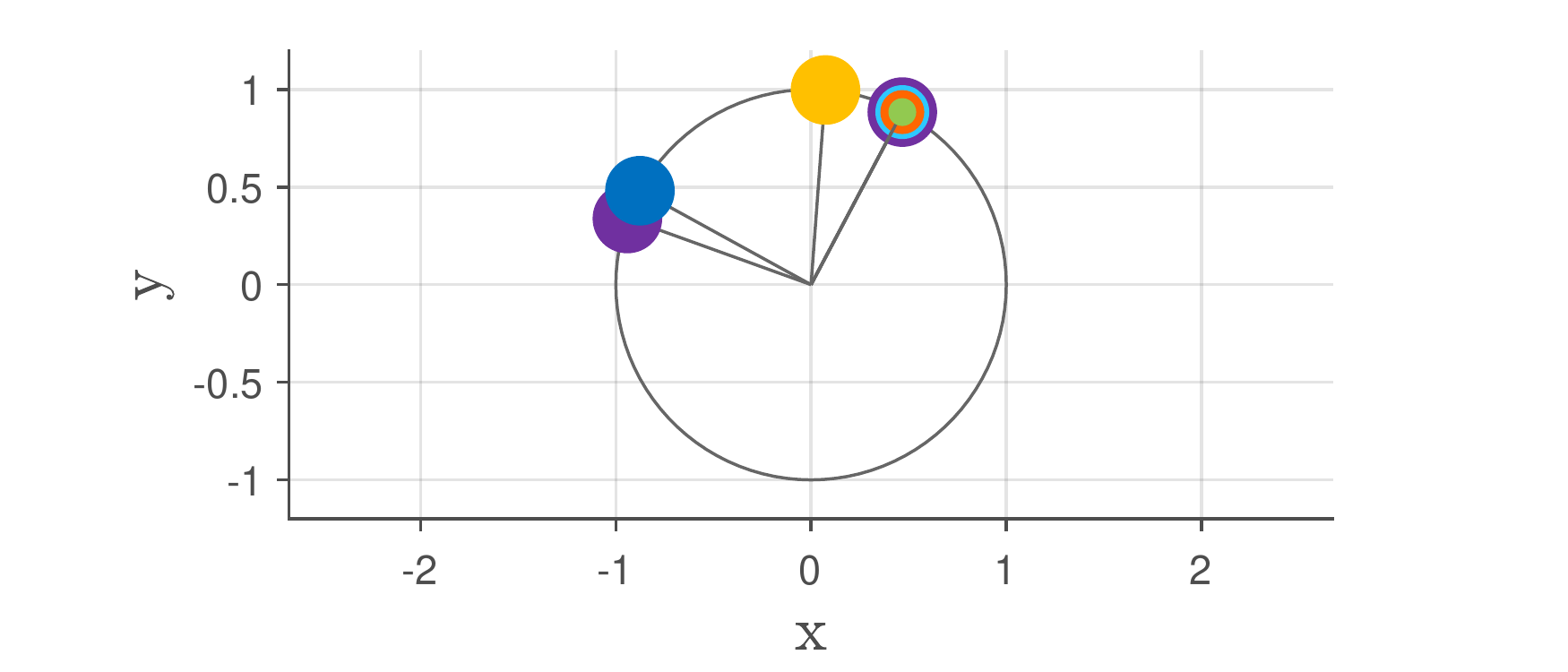}
	\caption{Embedding of rows of matrix $U$ in Example~\ref{ex:graph}.}
	\label{fig:points1}
\end{figure}

From $(F^i)_{jk} \eqdef \| u_j - u'_k \|^2$, where $u_j$ are rows of $U$ that correspond to view $i$ and $u'_k$ are pivot rows we obtain
%
%
\begin{gather*} 
F^1 = \begin{bsmallmatrix}
0     &  2.28 \\        
2.28  &     0 \\ 
\end{bsmallmatrix}, ~~
F^2 = \begin{bsmallmatrix} 
0.02~     &  1.97
\end{bsmallmatrix},  
~~ 
F^3 = \begin{bsmallmatrix} 
1.46~  &  0.17
\end{bsmallmatrix}, 
\\
F^4 = \begin{bsmallmatrix} 
2.28~  &  0
\end{bsmallmatrix},
~~
F^5 = \begin{bsmallmatrix} 
2.28~  &  0
\end{bsmallmatrix}, 
~~
F^6 = \begin{bsmallmatrix} 
2.28~  &  0
\end{bsmallmatrix}. 
\end{gather*}
%
By solving a linear assignment problem for each $F^i$ as the cost matrix (which aims to find the permutation matrix $P^i$ such that $\langle P^{i},  F^i \rangle$ is minimized) we obtain lifting permutations
%
\begin{gather*} 
P^1 = \begin{bsmallmatrix}
1  &  0 \\        
0  &  1 \\ 
\end{bsmallmatrix}, 
~~
P^2 = \begin{bsmallmatrix} 
1~  &  0
\end{bsmallmatrix},  
~~
P^3 = \begin{bsmallmatrix} 
0~  &  1
\end{bsmallmatrix}, 
\\
P^4 = \begin{bsmallmatrix} 
0 ~ &  1
\end{bsmallmatrix},  
~~
P^5 = \begin{bsmallmatrix} 
0 ~ &  1
\end{bsmallmatrix}, 
~~
P^6 = \begin{bsmallmatrix} 
0 ~ &  1
\end{bsmallmatrix}. 
\end{gather*}
%

\vspace{0.2cm}
\noindent $\bullet$ \textbf{Step 5}: Cycle-consistent pairwise associations are obtained by $P^i_j = P^i P^j$. Note that these associations correspond to the graph $\sg_3$ in Fig.~\ref{fig:GraphExample}.

\end{example}

\section{Theoretical Justifications} \label{sec:Theory}

In this section, we discuss the insights and theoretical justifications behind steps of the CLEAR algorithm.
To improve the readability, proof of all lemmas and propositions are presented in the Appendix. 

The discrete and combinatorial nature of the multi-way data association problem makes finding the optimal solution computationally prohibitive in practice. Hence, similar to the state-of-the-art methods, the CLEAR algorithm aims to find a suboptimal solution via a series of approximations of the original problem.

\subsection{Step 1: Reformulation} \label{sec:}

Before proceeding with obtaining an approximate solution, 
we reformulate \eqref{eq:optim} to obtain an equivalent problem. 
This equivalent problem, given in the following proposition, is amenable to a relaxation, which grants us an approximate solution in a computationally tractable manner.

\begin{prop} \label{prop:reform}
Problem~\ref{prob:Main} is equivalent to 
%
\begin{gather} \label{eq:p0}
\begin{array}{ll}
\underset{U \in \bu}{\text{\normalfont minimize}}& \tr \big(U^\top \Lnt \, U \big),  \\
\end{array}
\end{gather}
%
where $\bu \eqdef \{U \,:\, U = C^{-\frac{1}{2}} V,\, V\in \bv\}$,  $\bv$ is defined as the set of all matrices of form \eqref{eq:V}, and $U^\top U = I$.
\end{prop}

\begin{remark} \label{rem:symmetry}
The skew-symmetric part of $\Lnt$ does not affect the solution of \eqref{eq:p0} since for all $U$ and any skew-symmetric matrix $B$, $\tr(U^\top B \, U) =0$. 
This observation justifies using only the symmetric part of $\Lnt$ in step 1 of the CLEAR algorithm.		
\end{remark}

\subsection{Step 2: Estimating Size of Universe} \label{sec:}

From \eqref{eq:mEstimate} and \eqref{eq:mEstimate2}, CLEAR obtains an estimate for the size of universe based on the spectrum of $\Lnt$. By definition, the size of universe is the total number of unique items observed in all views (e.g., the size of universe in Fig.~\ref{fig:ExPj} is five), which essentially determines the number of columns of $U$ in \eqref{eq:p0} (or equivalently $V$ in \eqref{eq:optim}).
This approach is justified in the following analysis, which aims to show that, under certain bounded noise regimes, the estimated size $\hat{m}$ is guaranteed to be identical to its true value $m$.
Let us denote the ground truth association graph by $\sg$. Note that $\sg$ consists of $m$ clusters, each representing an item of the universe.

\begin{lemma} \label{lem:EigLhat}
If $\Ln$ is the normalized Laplacian matrix of the cluster graph $\sg$, then $\eig(\Ln)$ consists of zeros and ones. Moreover, the multiplicity of the zero eigenvalues is the number of clusters. 
\end{lemma}

Lemma~\ref{lem:EigLhat} implies that in the noiseless setting the number of zero eigenvalues of $\Ln$ is the size of universe, which is correctly recovered from \eqref{eq:mEstimate2} by counting the eigenvalues that are less than $0.5$.
We now consider the noisy association graph $\tilde{\sg}$ with normalized Laplacian $\Lnt = \Ln + N$, where $N$ is a symmetric matrix that represents the noise. Here, the symmetry assumption follows from using only the symmetric component of $\Lnt$ in the algorithm (see Remark~\ref{rem:symmetry}).

\begin{lemma} \label{lem:m}
Consider the estimate $\tilde{m}$ obtained by \eqref{eq:mEstimate2} from $\Lnt = \Ln + N$.
If $\|N\| < 0.5$, then $\tilde{m} = m$.
\end{lemma}

Lemma~\ref{lem:m} implies that, under a bounded noise regime, the estimated size of universe is equal to the true value.
In order to obtain a bound in terms of the number of wrong associations for which $\tilde{m} = m$ is guaranteed, let us consider a noise model where $\tilde{C} = C$. In this model, correct associations are potentially replaced with wrong ones, however, the degrees of vertices in  $\sg$ and $\tilde{\sg}$ remain the same.
Let $\tilde{A} = A + E$, where $A$ and $\tilde{A}$ are respectively the adjacency matrices of $\sg$ and $\tilde{\sg}$, and $E \in \{-1, 0,1 \}^{l \times l}$ represents the noise.  Further, let $e_{\max} \eqdef \max{\{e_1,\, e_2,\, \dots,\, e_l \} } $, where $e_i \eqdef \sum_{j=1}^{l}{|(E)_{ij}|}$ denotes the number of wrong associations at vertex $i$ of the graph $\tilde{\sg}$.
Let $c_{\min} > 0$ denote the smallest diagonal entry of the $C$ matrix.

\begin{prop} \label{prop:m}
Given $e_{\max}, \, c_{\min}$ defined above and $\tilde{m}$ obtained from \eqref{eq:mEstimate2}, if $e_{\max} < 0.5 \, c_{\min}$, then $\tilde{m} = m$.
\end{prop}

Proposition~\ref{prop:m} implies that when the noise magnitude (in terms of the number of mismatches) is sufficiently small, the estimated size of universe $\hat{m}$ is equal to the true value $m$. 
We point out that in practice the bound in Proposition~\ref{prop:m} is conservative and correct estimates may be obtained in larger noise regimes or for noise with a more realistic model. 
In higher noise regimes where the estimate can have a large error, taking the maximum in \eqref{eq:mEstimate} ensures the distinctness constraint (which implies that items in each view are unique), and therefore the estimated $m$ cannot be less than the maximum number of items observed at a view.

The estimated value of $m$ obtained from \eqref{eq:mEstimate} fixes the size of $U$ in \eqref{eq:p0} throughout the algorithm.
Since (as we will show) each iteration of the CLEAR algorithm has a small execution time, instead of using a fixed value an alternative approach is to consider multiple values for $\hat{m}$ (e.g., by looping over all feasible $\hat{m}$) and choosing the solution that maximizes the objective in \eqref{eq:optim}.
In our empirical evaluations we observed that this approach, which comes at the expense of a higher execution time, does not notably improve the accuracy of the results. This empirical observation hints that the estimated value of $\hat{m}$ is often close to its optimal value, advocating the proposed estimation approach.  
%

\begin{remark} \label{rem:eigengap}
In the spectral graph clustering methods, the ``eigengap'' heuristic is often used to estimate the number of clusters \cite{VonLuxburg2007}. 
In this approach, $\tilde{m}$ is chosen such that ${| \lambda_{\tilde{m}} -
\lambda_{\tilde{m}+1} |}$ is maximized, where $\lambda_k$'s are sorted eigenvalues
of ${ L_{\text{\normalfont sym}} \eqdef D^{-\frac{1}{2}} L D^{-\frac{1}{2}} }$. 
Unlike the normalized Laplacian $L_{\text{\normalfont nrm}}$ proposed in this
work (see Lemma~\ref{lem:EigLhat}), in the noiseless setting, the nonzero eigenvalues of $L_{\text{\normalfont sym}}$ depend on the size of clusters.
We believe that this can make the eigengap method more sensitive to noise. As we
will see in our empirical analysis in Section~\ref{sec:Simulations}, the
estimated universe size based on the eigengap heuristic can deviate considerably
from the true value in moderate noise regimes, while our approach
exhibits more robustness.	
\end{remark}

\subsection{Step 3: Lifting and Relaxation} \label{sec:}

In order to solve \eqref{eq:p0} in a computationally tractable manner, the second approximation used in the CLEAR algorithm is to
drop the discrete constraint $U \in \bu$ and allow $U$ to take values in $\br^{l \times m}$. This leads to the relaxed problem 
%
\begin{gather} \label{eq:p3}
\begin{array}{ll}
\underset{U \in \br^{l \times m}}{\text{minimize}}& \tr \big(U^\top \Lnt \, U \big)  \\
\text{subject to}   & ~ U^\top \, U = I,
\end{array}
\end{gather}
%
which is a generalized Rayleigh quotient problem. From the Rayleigh-Ritz theorem \cite[Sec 5.2.2]{Lutkepohl1996}, it follows that the solution of \eqref{eq:p3} is given by the eigenvectors corresponding to the $m$-smallest eigenvalues of $\Lnt$ (note that $m$ is estimated in the previous step).

We point out that the relaxation technique used here is similar to the relaxation used to solve the normalized minimum-cut problem in the spectral graph clustering literature \cite{VonLuxburg2007}.
This similarity is not surprising given the graph-theoretic interpretation of our problem discussed in Section~\ref{sec:GraphFrom}. 
Nevertheless, note that spectral graph clustering is based on $\tilde{L}_{\text{\normalfont sym}} \eqdef \tilde{D}^{-\frac{1}{2}} \tilde{L} \tilde{D}^{-\frac{1}{2}}$ (or other normalized Laplacians) instead of $\Lnt$.

\subsection{Step 4: Projection and Embedding} \label{sec:}

In order to obtain an approximate solution for the original problem \eqref{eq:p0}, the solution $U^* \in \br^{l\times m}$ obtained from solving \eqref{eq:p3} should be projected back to the discrete set $\bu$. 
This step is critical for ensuring the cycle consistency and distinctness constraints. In fact, as we will show in Section~\ref{sec:Simulations}, the solutions returned by some state-of-the-art methods could violate the cycle consistency or distinctness constraints in high-noise regimes due to bad projections.

To project $U^*$ onto $\bu$, several approaches can be considered.
Our approach is inspired by the spectral graph clustering literature \cite{VonLuxburg2007,Shi2000,Ng2002}, where rows of $U^*$ are normalized and embedded as points in $\br^{m}$.
These points are then grouped into $m$ disjoint sets based on their distance to cluster centers.
Despite the aforementioned similarity, a key difference in our setting is the
existence of the distinctness constraint (i.e., vertex coloring), which is not
present in spectral graph clustering \cite{Ng2002}. Hence, the $k$-means algorithm commonly used for grouping the embedded points in general violates the distinctness constraint. 
Furthermore, compared to other projection techniques that consider  this constraint (e.g., methods in \cite{Gallier2013,Bernard2018}), our approach has a lower complexity that leads to superior execution time.

Our approach is based on noting that rows of $V$ (defined in \eqref{eq:V})
consist of standard bases vectors which are orthogonal. 
Furthermore, as explained earlier, $V$ identifies graph clusters $\sa_1, \sa_2, \dots, \sa_m$, where vertices that belong to the same cluster have identical rows in $V$.
Since $U \eqdef C^{-\frac{1}{2}} \, V$ and $C$ is a diagonal matrix, it follows that in the \textit{noiseless setting} the set of \textit{normalized} rows of $U$ consists exclusively of $m$ orthogonal vectors.
Additionally, similar to $V$, normalized rows of $U$ that are identical correspond to vertices that belong to the same cluster.

In the noisy setting, from the Davis-Kahan theorem \cite{Davis1970} the eigenspace of the ground truth Laplacian matrix and its noisy version are ``close'' to each other (where ``closeness'' can be quantified by the noise magnitude, cf. the discussion in \cite[Sec. 7]{VonLuxburg2007}). 
Hence, in modest noise regimes, the rows of $U^*$ that belong to the same clusters are expected to remain close (in terms of the Euclidean distance) and almost perpendicular to other rows. 
This observation is leveraged by choosing $m$ rows of $U^*$ that are most orthogonal to each other (called pivots) to represent the clusters. The remaining rows are then associated to pivots
(while preserving distinctness)
based on distance in order to identify which cluster they belong to.

If $u_i$ denotes the $i$-th row of $U^*$, the problem of finding the $m$ most orthogonal rows can be formulated as finding a subset $\ss$ of rows that solves
%
\begin{gather} \label{eq:pivotRows}
\begin{array}{ll}
\underset{\ss \subset \bn_l}{\text{minimize}} & \sum_{i,j \in \ss}{ | \langle u_i, u_j \rangle |}
\\
\text{subject to} & |\ss| =m.
\end{array}
\end{gather}
%
The greedy strategy explained in step 3 of the CLEAR algorithm can be leveraged to efficiently find an approximate solution for \eqref{eq:pivotRows}.

After choosing the pivot rows, which are denoted by $u'_k$ and represent clusters, the remaining rows of $U^*$ are assigned to pivot rows through minimizing the within-cluster distances. This is formally stated as
%
\begin{gather} \label{eq:p4}
\begin{array}{ll}
\underset{\sa_1,\dots, \sa_m}{\text{minimize}} & \sum_{k = 1}^{m}  \sum_{v_j \in \sa_k}{ \| u_j - u'_k \|^2 }
\\
\text{subject to} & g(v_i) \neq g(v_j), \quad \forall v_i, v_j\in \sa_s.
\end{array}
\end{gather}
%
The constraint in \eqref{eq:p4} enforces the distinctness constraint (i.e., items in a view should not be in the same cluster). 
Let us define $F \in \br^{l \times m}$ such that $(F)_{jk} \eqdef \| u_j - u'_k \|^2$,
and denote its row blocks by 
\begin{equation}
  F = \begin{bmatrix}
	F^{1\, \top} & F^{2\, \top} & \cdots & F^{n\, \top} 
  \end{bmatrix}^\top,
\end{equation}
where the number of rows of block $F^i$ is equal to the number of items at view $i$.
Using this notation, and since $V$ encapsulates both the distinctness constraint and the cluster structure,\footnote{If the $j$-th entry in column $k$ of $V$ is nonzero, then $v_j \in \sa_k$.} \eqref{eq:p4} can be represented in matrix form as $\underset{V\in \bv}{\mathrm{min}} ~ \langle V, \, F \rangle$.
Noting that
%
\begin{subequations}
\begin{align} 
\underset{V\in \bv}{\mathrm{min}} ~ \langle V,\, F \rangle & = 
\underset{P^i \in \bp^L}{\mathrm{min}} ~ \textstyle{\sum}_{i=1}^{n}{\langle P^{i},\, F^i \rangle} \\
& =  \textstyle{\sum}_{i=1}^n  ~ \underset{P^i \in \bp^L}{\mathrm{min}} ~{\langle P^{i},\, F^i \rangle},
\label{eq:p5}
\end{align}
\end{subequations}
%
and since each subproblem in \eqref{eq:p5} 
is a linear assignment problem \cite{Burkard2012},
the optimal solution can be obtained by, e.g., applying the Hungarian (Kuhn-Munkres) algorithm on each block $F^i$.

From the implementation point of view, as long as the lifting permutation structure of $P^i$ is preserved, faster suboptimal methods can be used instead to solve \eqref{eq:p5}.
To improve the runtime, instead of the Hungarian algorithm CLEAR uses a suboptimal greedy strategy based on sorting, where the index of the smallest entries of $F^i$ are used to determine the index of one entries in $P^i$. These indices are chosen with care to ensure that $P^i$ is a lifting permutation (i.e., each row has a single one entry and each column has at most a single one entry). 
In our empirical evaluations we observed that this suboptimal strategy performs as well as the optimal Hungarian algorithm most of the time in term of accuracy, but has a considerable speed advantage.

Lastly, we emphasize that the proposed projection technique is based on the orthogonality property of the embedded rows. Hence, the results are not affected by any transformation that preserves the orthogonality. 
This is particularly important since solutions of \eqref{eq:p3} are only recovered up to an orthogonal transformation (i.e., if $U^*$ is a solution so is $U^* R$ for any orthogonal matrix $R$).

\subsection{Computational Complexity} \label{sec:}

The computational complexity of CLEAR is determined by the eigendecomposition algorithm (used for estimating the universe size and computing the eigenvectors of $\Lnt$) and the linear assignment problem (used for the projection step). 
The time complexity of the eigendecomposition and optimal linear assignment (e.g., Hungarian algorithm) are respectively $\mathrm{O}(l^3)$ and $\mathrm{O}(n\, m^3)$, where $l$ is the number of vertices in the assignment graph, $n$ is the number of views, and $m$ is the size of universe.

In order to improve the speed and scalability of CLEAR, a breadth-first search (BFS), which has the worst computational complexity of $\mathrm{O}(l^2)$, can be used to first find the connected components of $\tilde{\sg}$. 
The spectrum (i.e., eigenvalues of normalized Laplacian) of $\tilde{\sg}$ is then obtained by taking the disjoint union of components' spectra (similarly eigenvectors are given by zero padding the components' eigenvectors) \cite{Chung1997}. 
Through this approach, the complexity of the eigendecomposition is reduced to $\mathrm{O}(l_1^3)$, where $l_1$ is the number of vertices in the largest connected component of $\tilde{\sg}$. 
In practice, often the association graph consists of many disjoint components (e.g., see examples in Section~\ref{sec:Experiments}), and the aforementioned procedure considerably improves the runtime and scalability.

The second improvement in speed is achieved by replacing the Hungarian algorithm with the suboptimal sorting strategy. This approach reduces the computational complexity of the projection step to  $\mathrm{O}(n \, m^2 \log(m))$.

\section{Applications: Edge-Centric vs. Clique-Centric} \label{sec:apps}

In this section, we divide the applications that benefit from solving the
multi-view matching problem into two categories, namely 
\emph{edge-centric} and
\emph{clique-centric}. It will become clear shortly that making this subtle distinction is
crucial for choosing the appropriate evaluation metric for each category.

In edge-centric applications, one ultimately
seeks to establish associations between \emph{pairs of views} (and not \emph{all} views).
In graph terms, this corresponds to seeking individual edges of the
association graph (hence the name). 
For example, using multi-view matching algorithms to associate features
between multiple images for estimating relative transformation
between the corresponding pair of camera poses \cite{Zhou2015} belongs to this category.
The purpose of using multi-view matching
techniques in
such applications is to \emph{refine} the initial noisy associations by
incorporating information from multiple views and enforcing cycle consistency.
Based on this definition, even a cycle
\emph{inconsistent} set of
associations is still a \emph{feasible} (although erroneous) solution in
edge-centric applications. 
As a result, computing standard metrics such as precision/recall based on
\emph{individual}
edges of the association graph is appropriate for evaluating
the performance of multi-view association algorithms in such applications.

By contrast, in what we refer to as clique-centric applications, one 
ultimately seeks to fuse information \emph{globally} (i.e., across \emph{all} views) as prescribed by the cliques of the association graph. For example, consider the map fusion problem that arises in
single/multi-robot SLAM \cite{aragues2011consistent}. After
identifying every sighting of each unique landmark in all maps (i.e., encoded in
		the cliques of a cycle-consistent association graph) via multi-view
matching techniques, the corresponding measurements (across
\emph{all} maps) must be fused together in the SLAM \mbox{back-end} to generate the global
map.
Note that such notion of \emph{global} fusion is well-defined only if
association, as a binary relation on the set of observations, is an
equivalence relation.\footnote{This mainly refers to transitivity since for all
		practical purposes in robotics, associations are always reflexive and
symmetric.} Therefore, unlike edge-centric applications, cycle consistency
 of associations is a must in clique-centric applications
where the observations in each equivalence class are fused together.
Cycle-inconsistent solutions must therefore be made cycle consistent before being used in
such applications. An implicit and natural way of doing this is via the so-called
transitive closure of associations which gives the smallest equivalence relation
containing the original associations. In graph terms, this is equivalent to
completing each connected component of the association graph into a clique.
Thus evaluating such cycle-inconsistent solutions
by computing precision/recall based on individual edges
can be highly misleading in the case of clique-centric applications. In such
cases, precision/recall must be computed \emph{after} completing the connected
components of the association graph (i.e., for the transitive closure).

\begin{figure}[t]
\centering
\includegraphics[trim = 0mm 0mm 0mm 0mm, clip, width=0.4\textwidth] {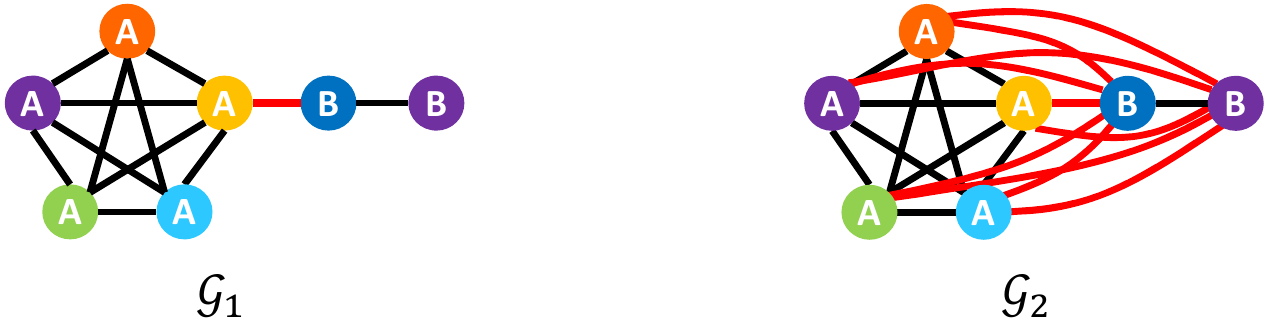}	
\caption{Evaluating the performance of cycle-inconsistent solutions (e.g.,
  $\sg_1$) for
  clique-centric applications must be done \emph{after} completing the connected
components of the association graphs (i.e., for the transitive closure
$\sg_2$).
Even a single incorrect edge (drawn in red) may have catastrophic consequences in
clique-centric applications.}
\label{fig:GraphExample2}
\end{figure}

Note that a single incorrect association only affects local
(pairwise) fusions in edge-centric
applications, while it may have a catastrophic global impact in clique-centric
domains.
This is illustrated in Fig.~\ref{fig:GraphExample2} using a simple example.
Here the association graph $\sg_1$ contains only a single incorrect edge drawn in
red. Although $\sg_1$ has a high precision and a high recall for edge-centric applications,
it is not cycle consistent and thus does not immediately prescribe a valid
solution to clique-centric applications. As mentioned above, for such applications one must first compute
the transitive closure of $\sg_1$. The transitive closure of $\sg_1$ is given by
$\sg_2$ which performs poorly in terms of precision. Note that each red edge in $\sg_2$
indicates an incorrect fusion of two observations.

Although CLEAR, by construction, always returns cycle-consistent solutions, as
we will see in the following sections several existing algorithms
may violate cycle consistency in noisy regimes. It is thus crucial to be aware
of the distinction between local (pairwise) and global fusion in order to use the
appropriate performance metric in a particular application.

\section{Simulation Results} \label{sec:Simulations}

\begin{figure*}[t!]
	\centering
	\begin{subfigure}[b]{0.142\textwidth}	\includegraphics[trim = 10mm 0mm 18mm 0mm, clip, width=1.0\textwidth] {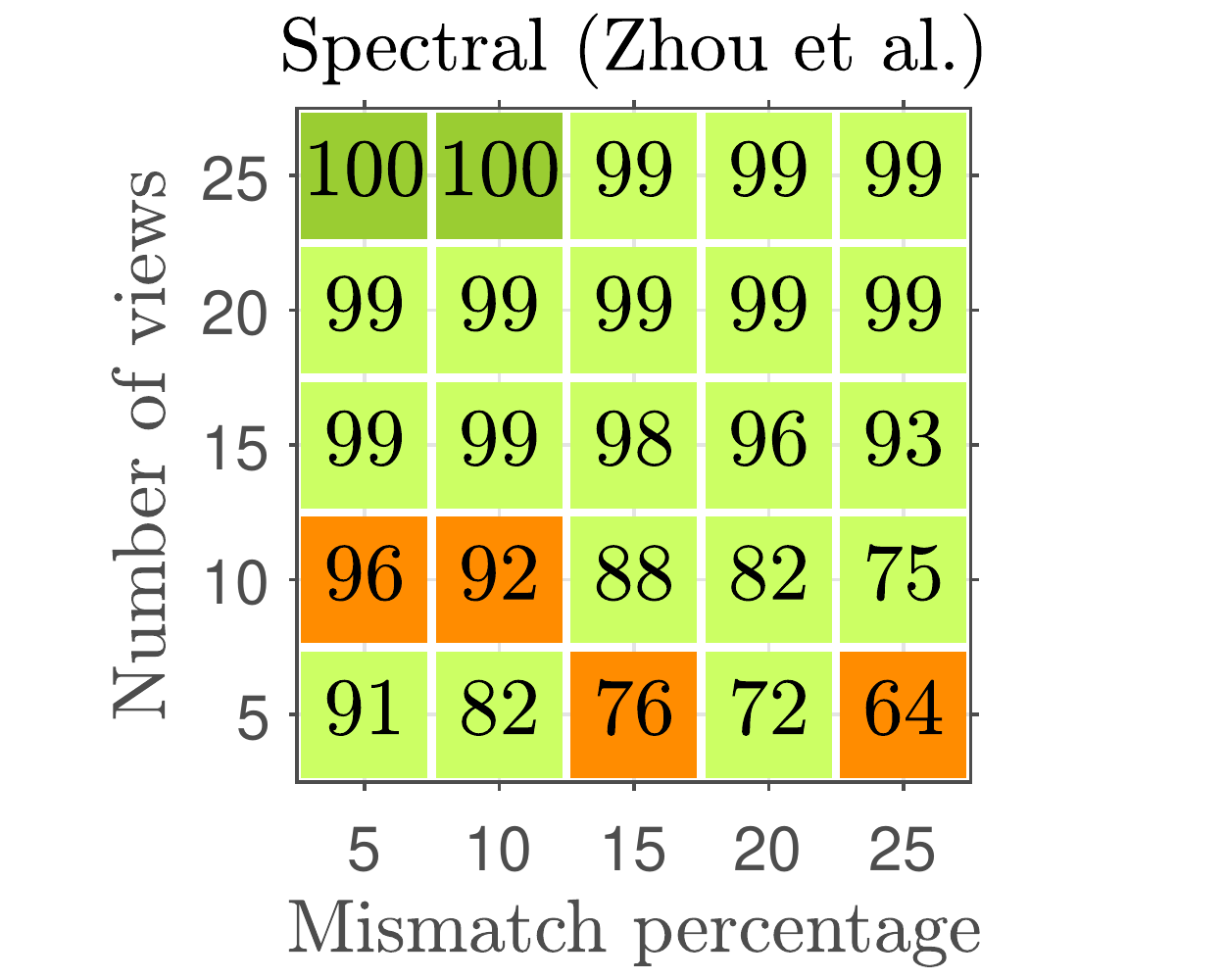}
	\end{subfigure}%
	\begin{subfigure}[b]{0.142\textwidth}
		\includegraphics[trim = 10mm 0mm 18mm 0mm, clip, width=1.0\textwidth] {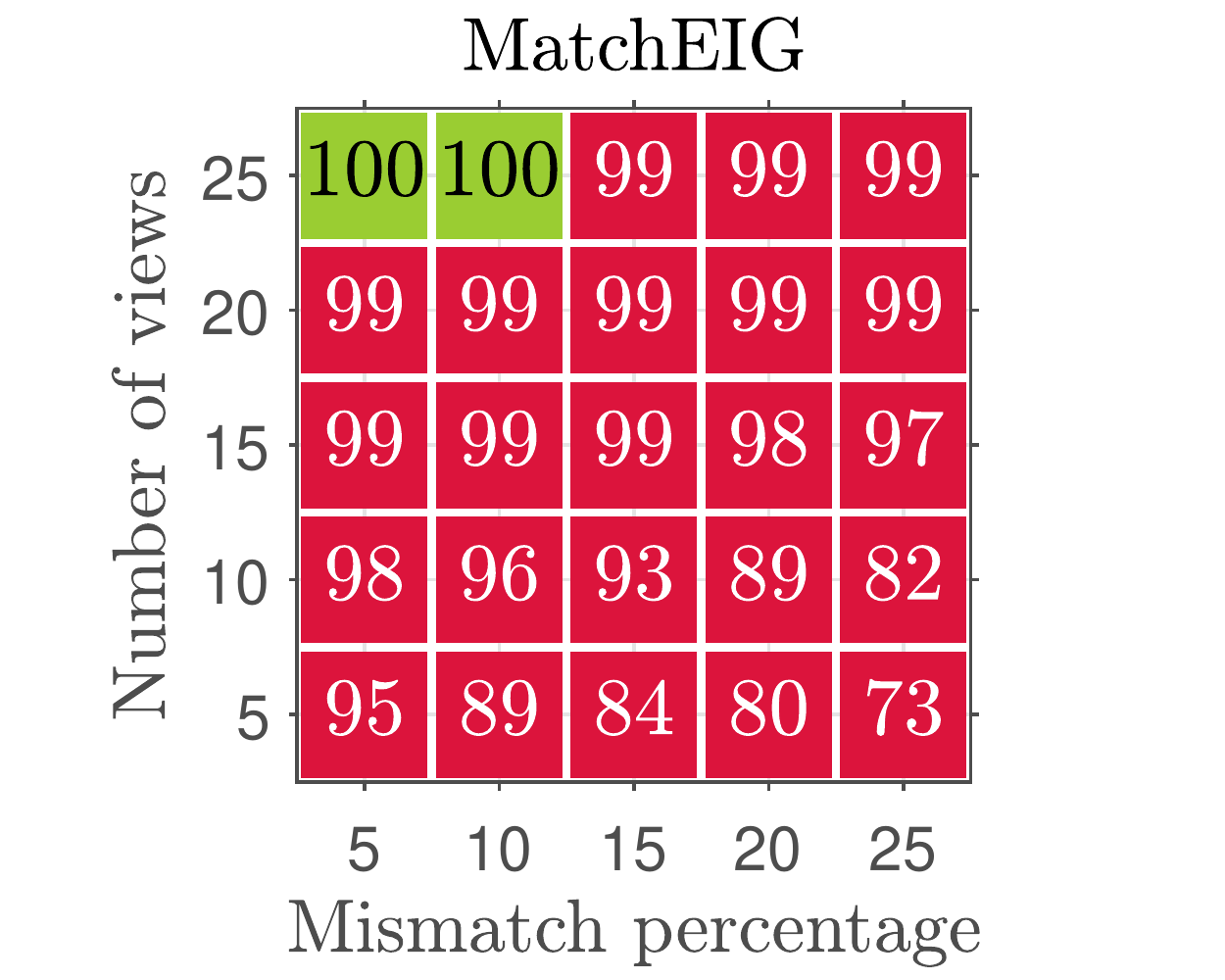}
	\end{subfigure}%
	\begin{subfigure}[b]{0.142\textwidth}
		\includegraphics[trim = 10mm 0mm 18mm 0mm, clip, width=1.0\textwidth] {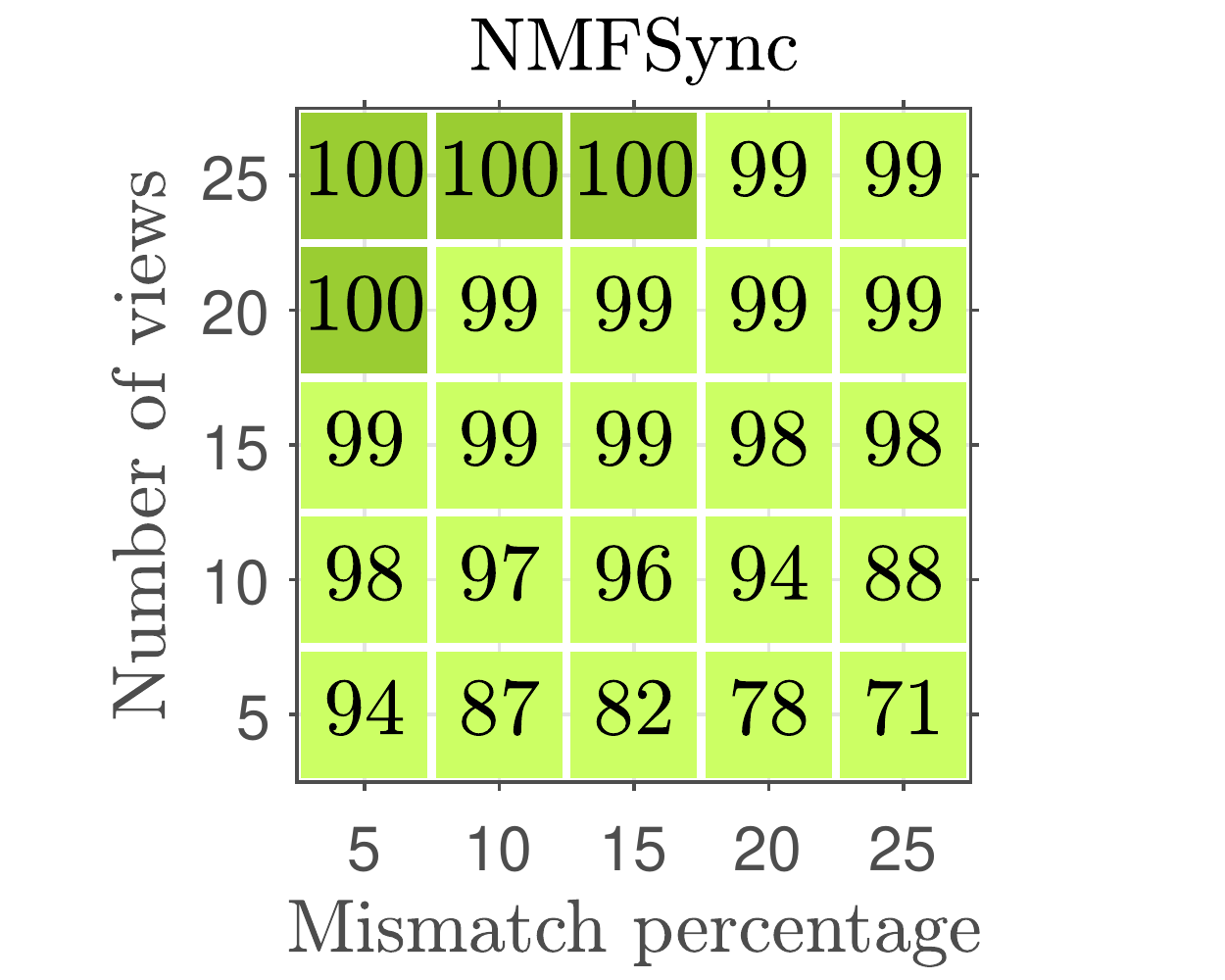}
	\end{subfigure}%
	\begin{subfigure}[b]{0.142\textwidth}
		\includegraphics[trim = 10mm 0mm 18mm 0mm, clip, width=1.0\textwidth] {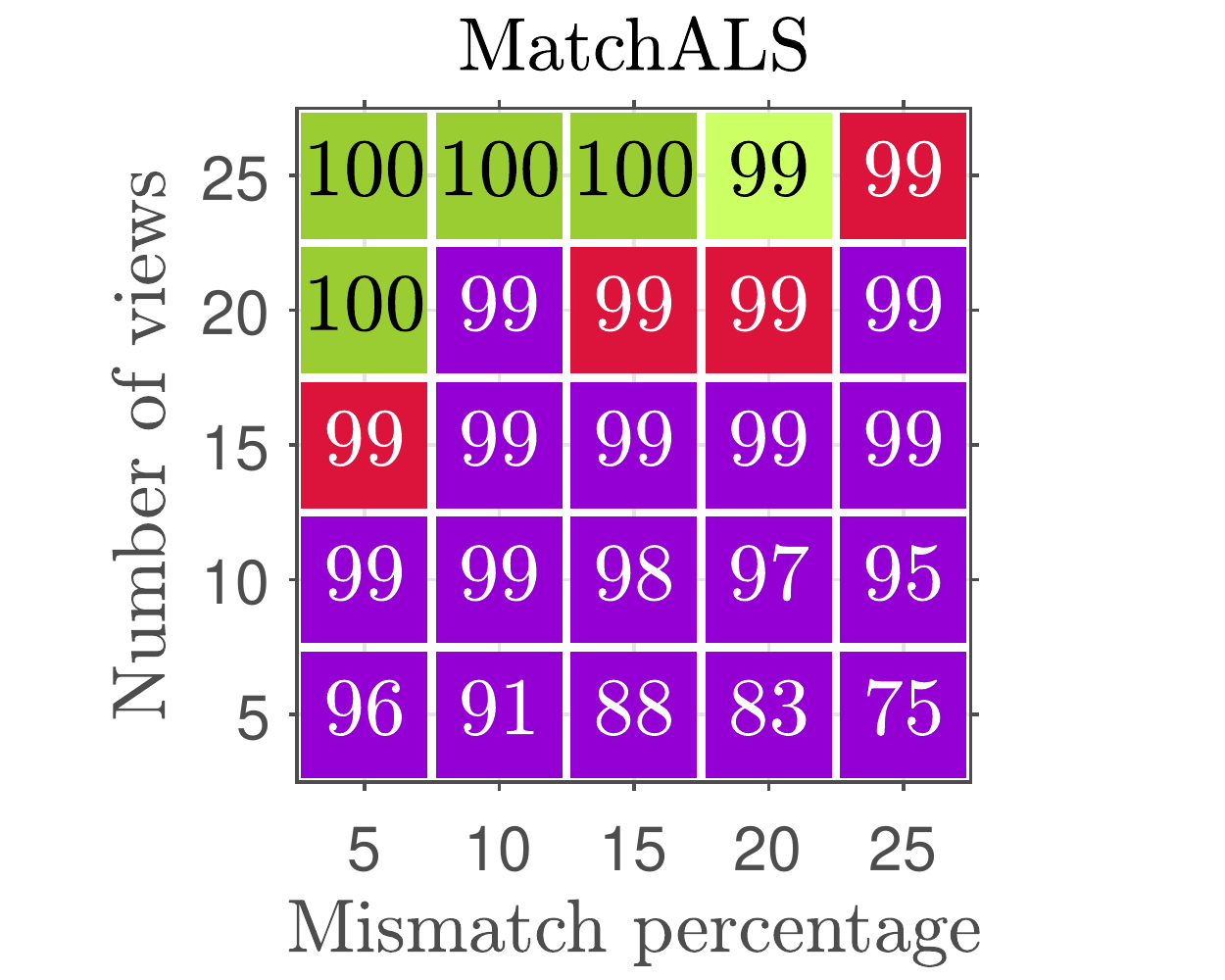}
	\end{subfigure}%
	\begin{subfigure}[b]{0.142\textwidth}
		\includegraphics[trim = 10mm 0mm 18mm 0mm, clip, width=1.0\textwidth] {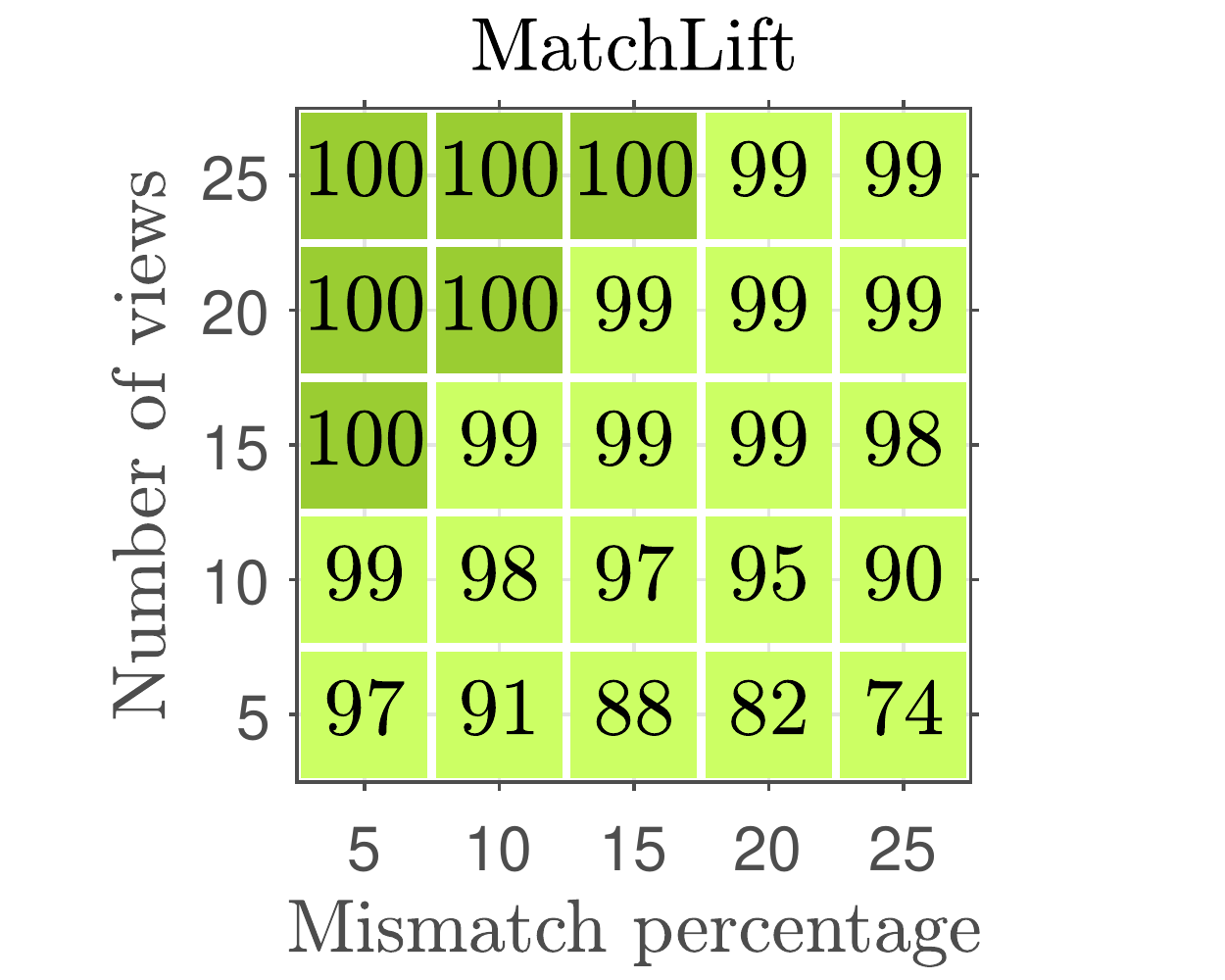}
	\end{subfigure}%
	\begin{subfigure}[b]{0.142\textwidth}
		\includegraphics[trim = 10mm 0mm 18mm 0mm, clip, width=1.0\textwidth] {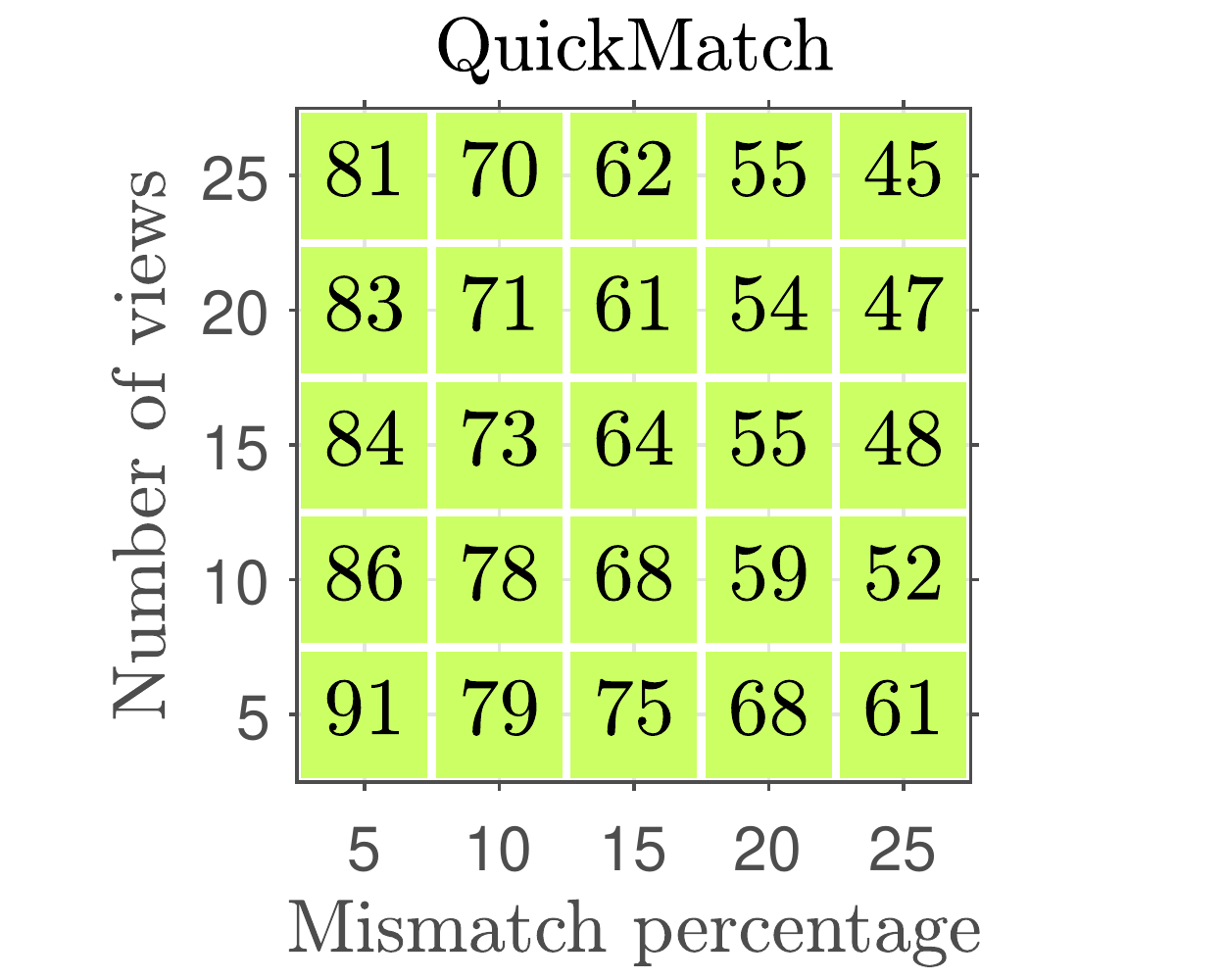}
	\end{subfigure}%
	\begin{subfigure}[b]{0.142\textwidth}
		\includegraphics[trim = 10mm 0mm 18mm 0mm, clip, width=1.0\textwidth] {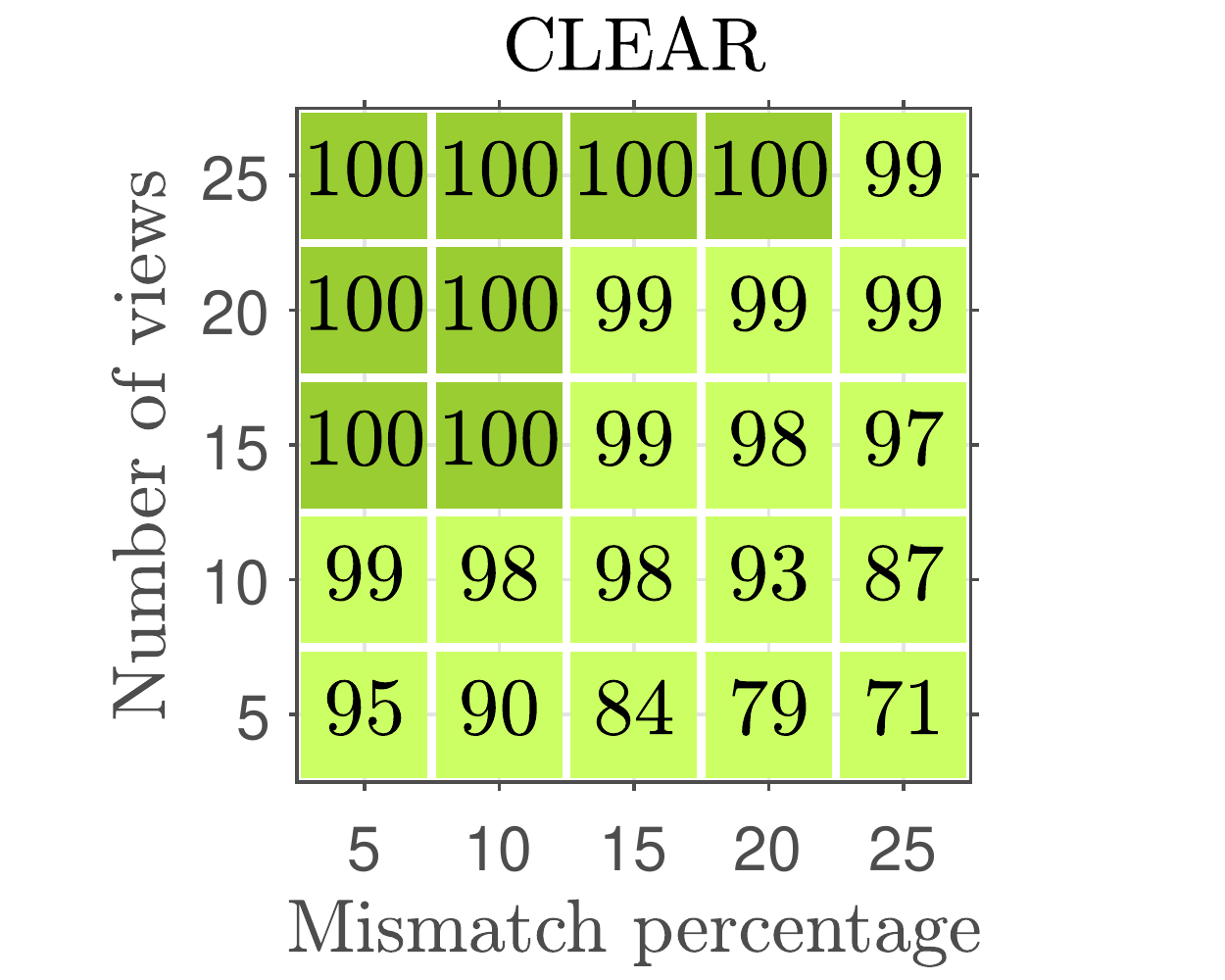}
	\end{subfigure}%
	%
	%
	\\ \vspace{0.05in}
	\begin{subfigure}[b]{0.142\textwidth}
		\includegraphics[trim = 10mm 0mm 18mm 0mm, clip, width=1.0\textwidth] {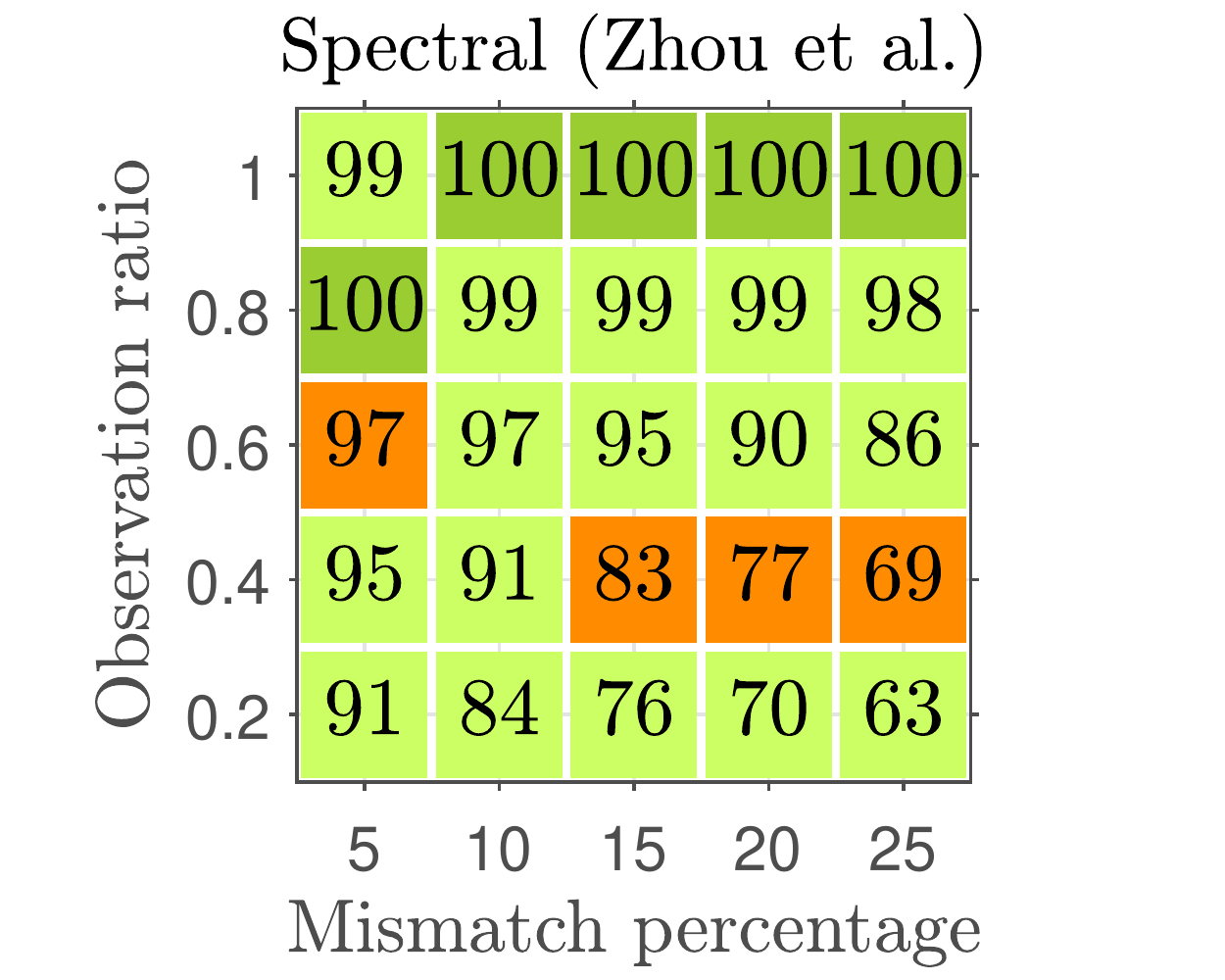}
	\end{subfigure}%
	\begin{subfigure}[b]{0.142\textwidth}
		\includegraphics[trim = 10mm 0mm 18mm 0mm, clip, width=1.0\textwidth] {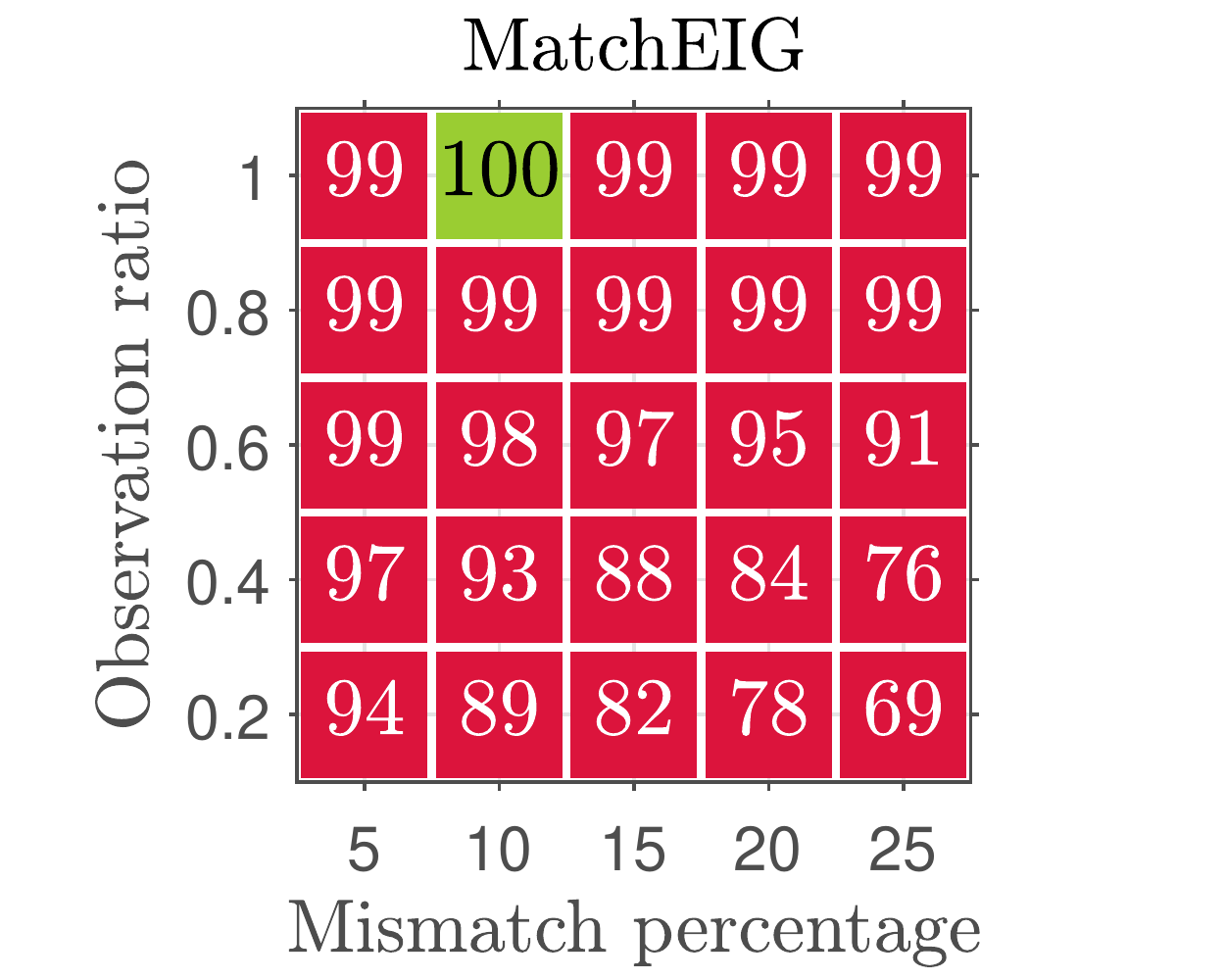}
	\end{subfigure}%
	\begin{subfigure}[b]{0.142\textwidth}
		\includegraphics[trim = 10mm 0mm 18mm 0mm, clip, width=1.0\textwidth] {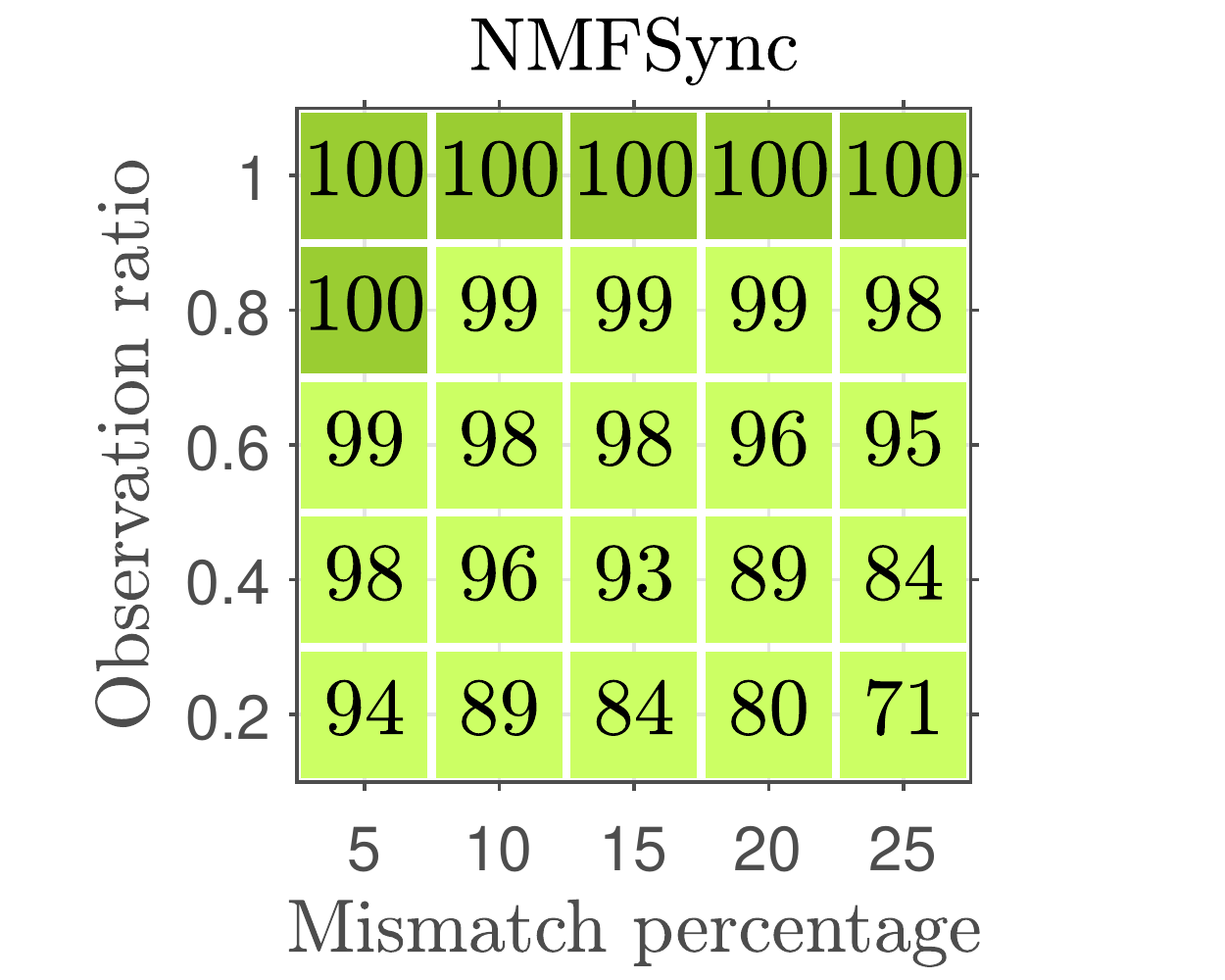}
	\end{subfigure}%
	\begin{subfigure}[b]{0.142\textwidth}
		\includegraphics[trim = 10mm 0mm 18mm 0mm, clip, width=1.0\textwidth] {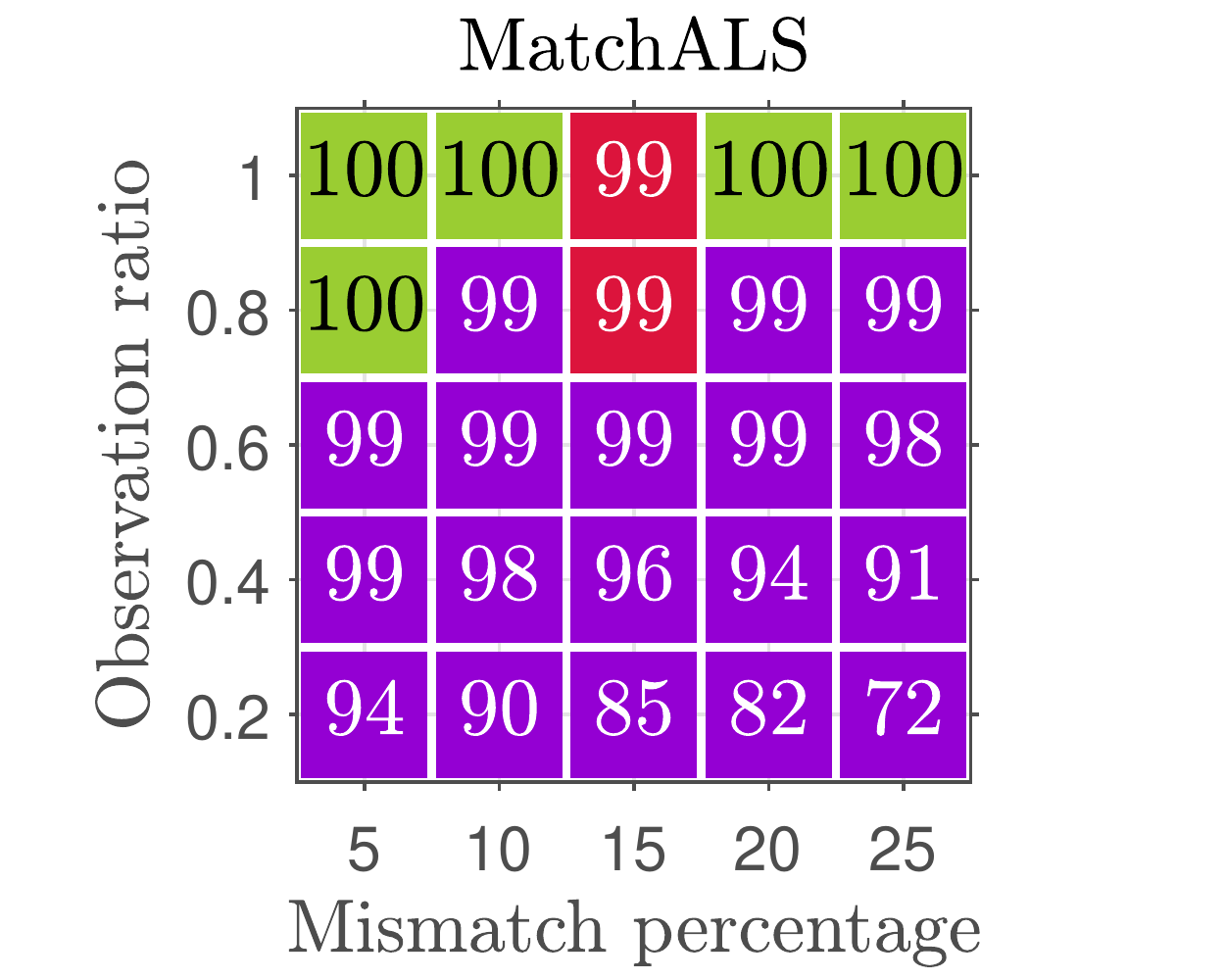}
	\end{subfigure}%
	\begin{subfigure}[b]{0.142\textwidth}
		\includegraphics[trim = 10mm 0mm 18mm 0mm, clip, width=1.0\textwidth] {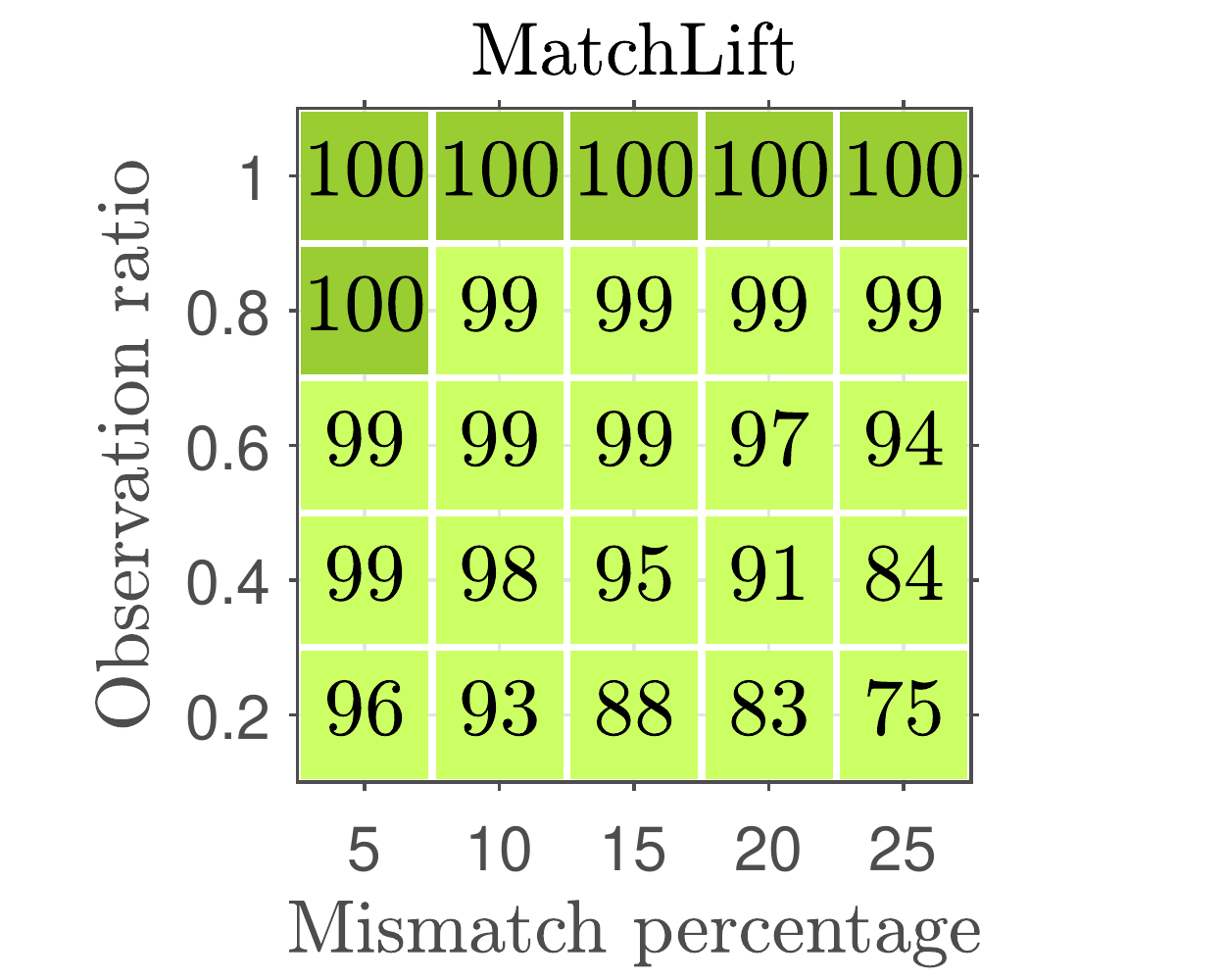}
	\end{subfigure}%
	\begin{subfigure}[b]{0.142\textwidth}
		\includegraphics[trim = 10mm 0mm 18mm 0mm, clip, width=1.0\textwidth] {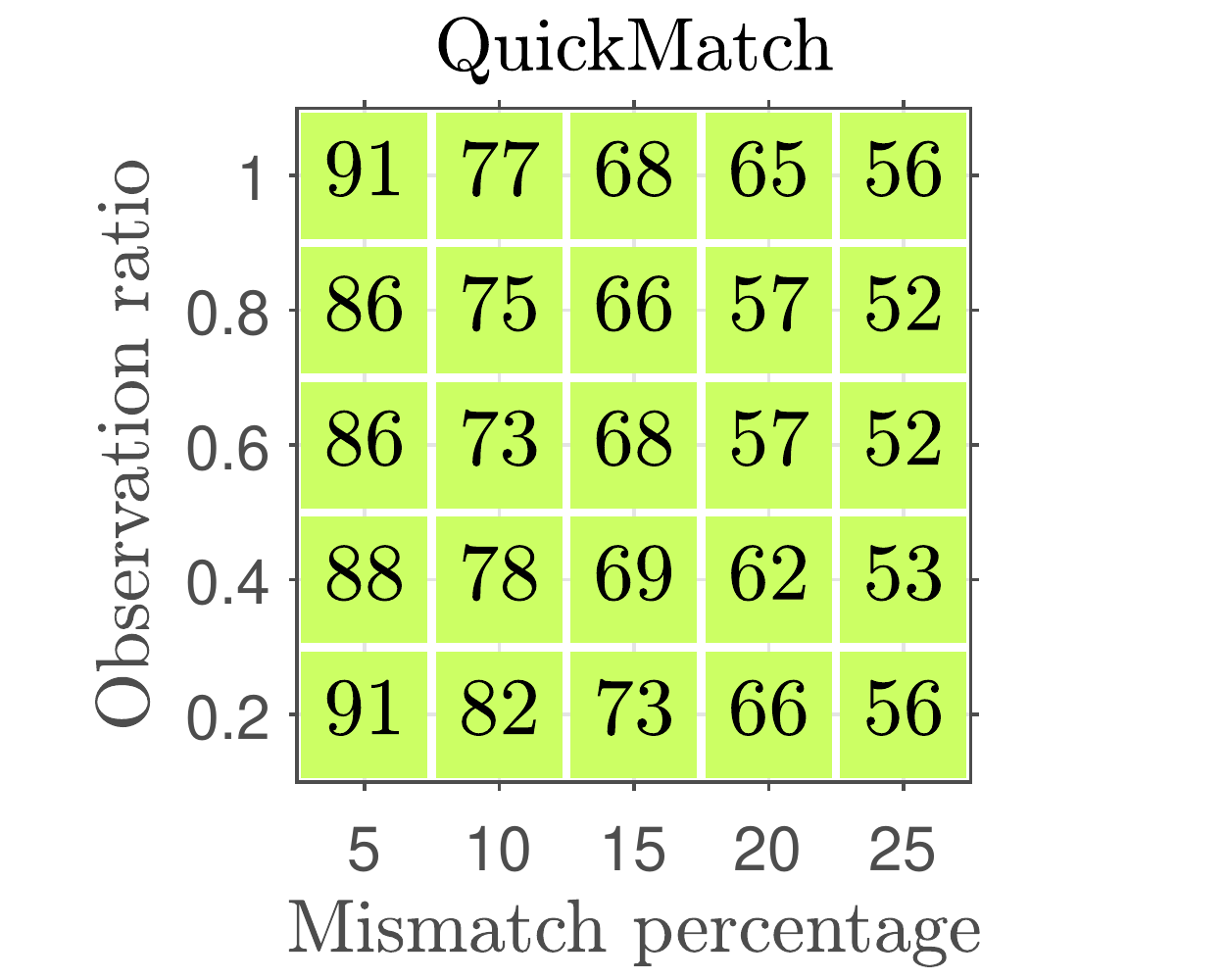}
	\end{subfigure}%
	\begin{subfigure}[b]{0.142\textwidth}
		\includegraphics[trim = 10mm 0mm 18mm 0mm, clip, width=1.0\textwidth] {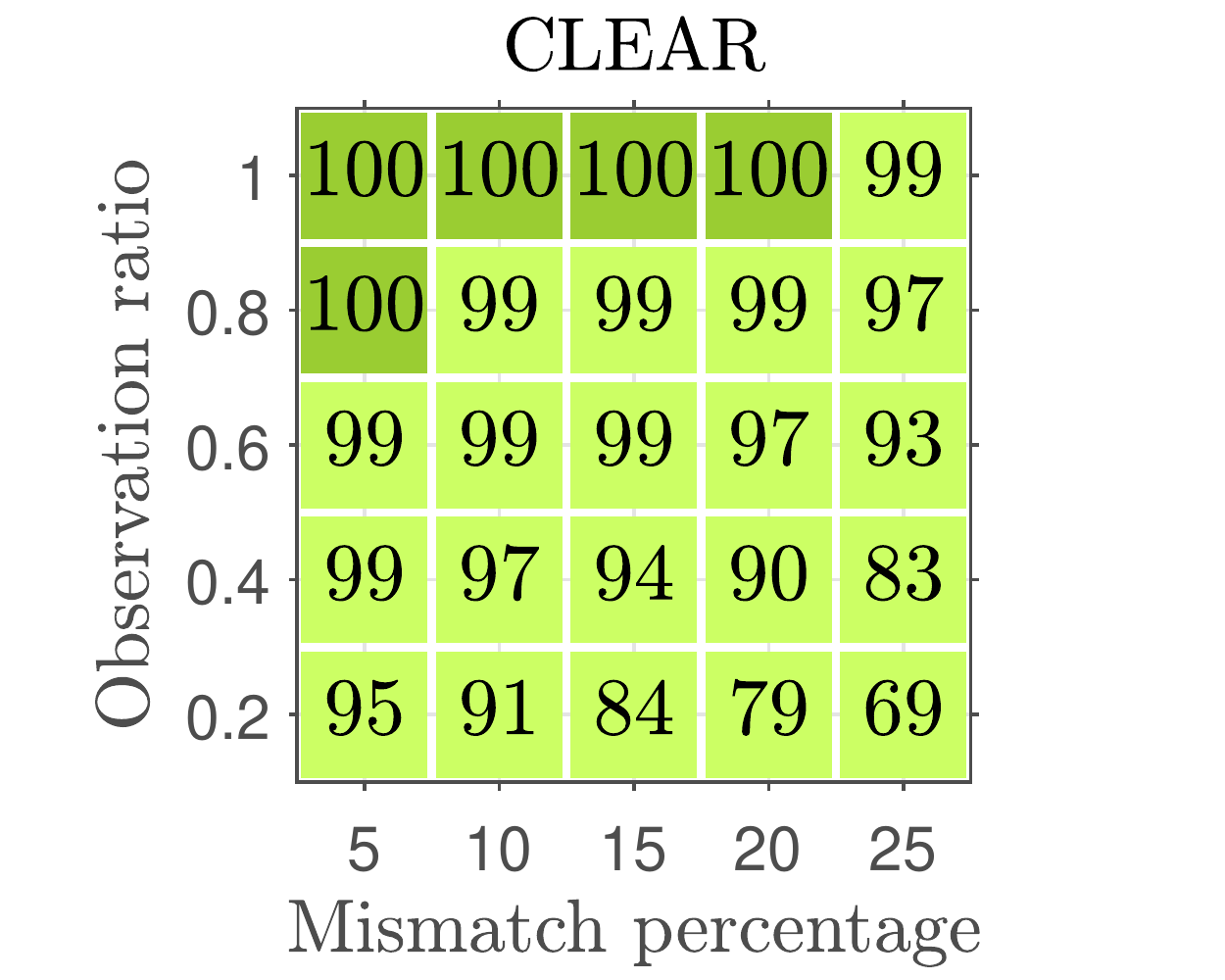}
	\end{subfigure}%
	\\ \vspace{0.05in}
	\begin{subfigure}[b]{1.0\textwidth}
		\includegraphics[trim = 13mm 108mm 13mm 100mm, clip, width=1.0\textwidth] {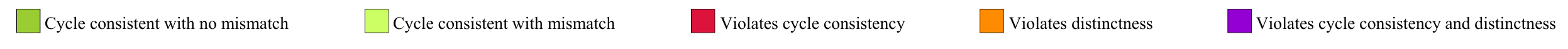}
	\end{subfigure}%
	\caption{(Best viewed in color) Comparison of the state-of-the-art
	  algorithms with CLEAR for uniformly sampled observations. Various number
	  of views and observation ratios versus the percentage of mismatch in the
	  input are considered. The $\text{F}_1$ score is reported in percentage in
	  each grid (the higher the better). These values are computed based on
	  individual edges in the association graph (i.e., for edge-centric applications); see
	  Section~\ref{sec:apps}.
		%
	}
	\label{fig:AgentCompare}	
\end{figure*}

In this section, we use Monte Carlo analysis with synthetic data to compare  CLEAR with several state-of-the-art algorithms across different noise regimes. The aim of these comparisons is to 1) analyze the accuracy of the returned solutions; 2) identify algorithms that violate the cycle consistency or distinctness constraints in high-noise regimes;
and 3) evaluate the accuracy of the proposed technique for estimating the universe
size.

Algorithms used in our comparisons, which span across three aforementioned domains, include: 1) MatchLift \cite{Chen2014} and MatchALS \cite{Zhou2015} that are based on a convex relaxation; 2) Spectral algorithm \cite{Pachauri2013} extended for partial permutations by Zhou et al. \cite{Zhou2015}, MatchEig \cite{Maset2017}, and NMFSync \cite{Bernard2018} that are based on a spectral relaxation; 3) and QuickMatch \cite{Tron2017} that is a graph clustering approach.


We consider scenarios with various number of views and observations across different mismatch percentage in the pairwise correspondences.
The mismatch in correspondences is introduced by randomly reassigning correct matches to wrong ones according to a uniform distribution.
In all comparisons, the universe is set to contain $100$ items, where this value is assumed to be unknown to algorithms and should be estimated.
For algorithms that require the knowledge of universe size (all except QuickMatch), the same estimated value obtained for CLEAR from \eqref{eq:mEstimate} is used. 

We report the $\text{F}_1$ score, which is commonly used in the literature and is defined as
$f \eqdef \frac{2\, p \, r}{p + r} \in [0, 1]$, to evaluate the performance of
the algorithms.
Here, precision $p \in [0, 1]$ is defined as the number of correct associations
divided by the total number of associations in the output, and recall $r \in [0,
1]$  is the number of correct associations in the output divided by the total
number of associations in the ground truth. The best performance is achieved
when $f = 1$ (when $p = q = 1$) and the worst when $f = 0$ (zero precision and/or zero recall).

In the first comparison, the algorithms are evaluated for different number of views and percentage of mismatch in the input. 
The observation ratio is fixed at $0.5$; i.e., in each view, $50$ (out of
$100$) items of the
universe are observed. These items are sampled uniformly at random.
For each number of views and mismatch percentage, $10$ Monte Carlo simulations are generated 
and the average F$_1$ score of the outputs across these simulations is reported
in the first row of Fig.~\ref{fig:AgentCompare} (in percentage).
In the second comparison (second row in Fig.~\ref{fig:AgentCompare}), the number
of views in all Monte Carlo simulations is fixed at the value of $n = 10$, and
results for various observation ratios of universe items and input mismatch percentage are reported. 
Similar to the first comparison (first row in Fig.~\ref{fig:AgentCompare}), each observation ratio indicates the number of
items that were observed (i.e., uniformly sampled at random) in a
view. For example, observation ratio of $0.2$ indicates that each agent observed $20$
(out of $100$) items of the universe.
%

Fig.~\ref{fig:AgentCompare} shows that for a fixed observation ratio,
as the number of views increases, the F$_1$ score also increases. This
indicates that the algorithms are able to leverage the redundancy in
observations with the help of the cycle consistency constraint.
For the same reason, for a fixed number of views, the F$_1$ score improves as
the observation ratio increases.

We also tested the returned solutions for cycle
consistency (transitive associations) and distinctness (two observations in a view cannot be associated to
each other). The results are displayed using colors in Fig.~\ref{fig:AgentCompare}. 
In particular, here dark green indicates that the (cycle consistent) ground
truth was recovered in all Monte Carlo iterations. Light green 
indicates that the returned solutions satisfied cycle consistency and distinctness, but
contained wrong associations in at least one of the simulations. Furthermore, red indicates that, in at least one
simulation, the output was not cycle consistent, orange indicates violation of
the distinctness constraint, and finally purple indicates violation of both cycle
consistency and distinctness constraints.

\begin{figure}
\centering
\begin{subfigure}[b]{0.142\textwidth}	\includegraphics[trim = 10mm 0mm 18mm 0mm, clip, width=1.0\textwidth] {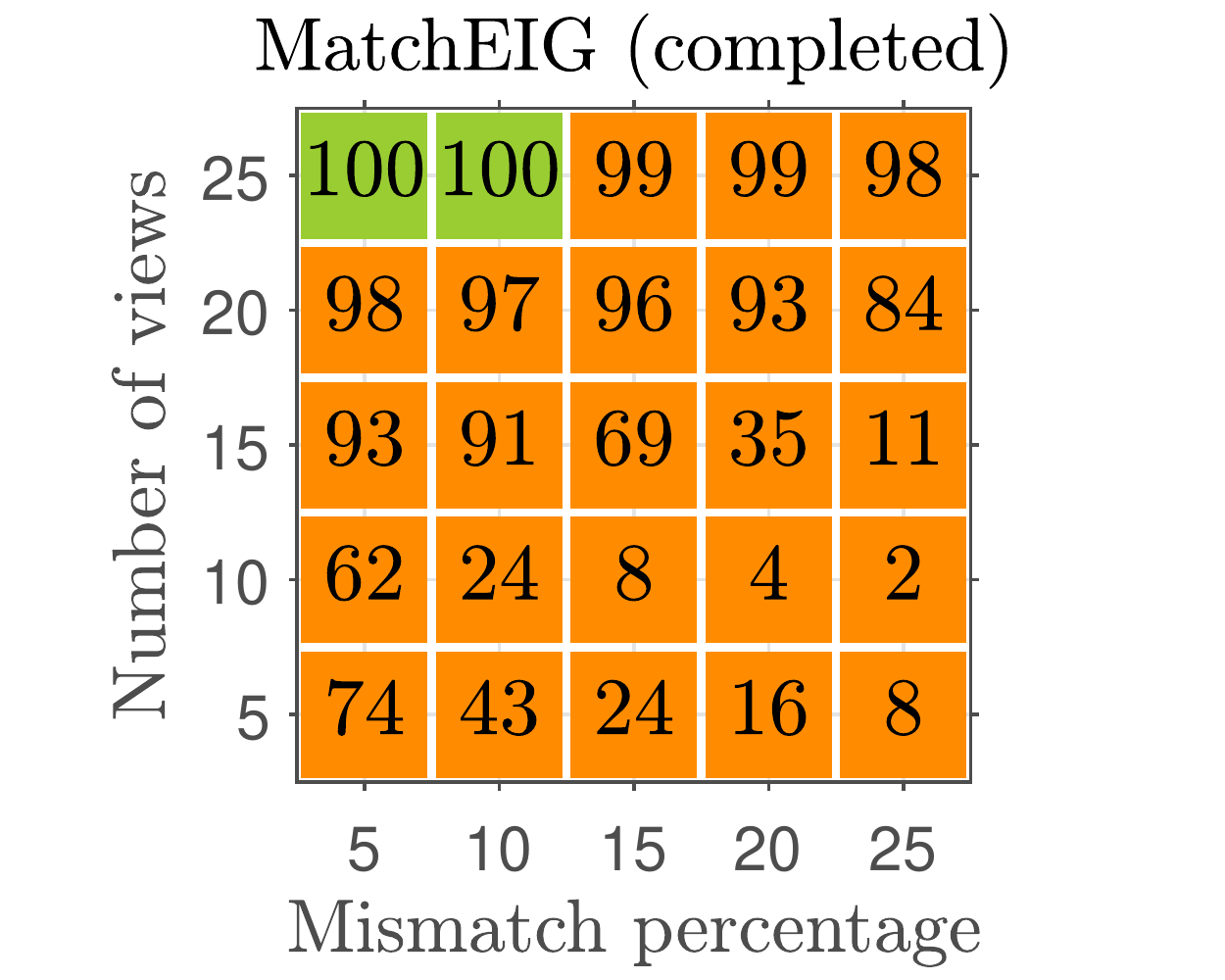}
\end{subfigure}%
~~~~~~~
\begin{subfigure}[b]{0.142\textwidth}
	\includegraphics[trim = 10mm 0mm 18mm 0mm, clip, width=1.0\textwidth] {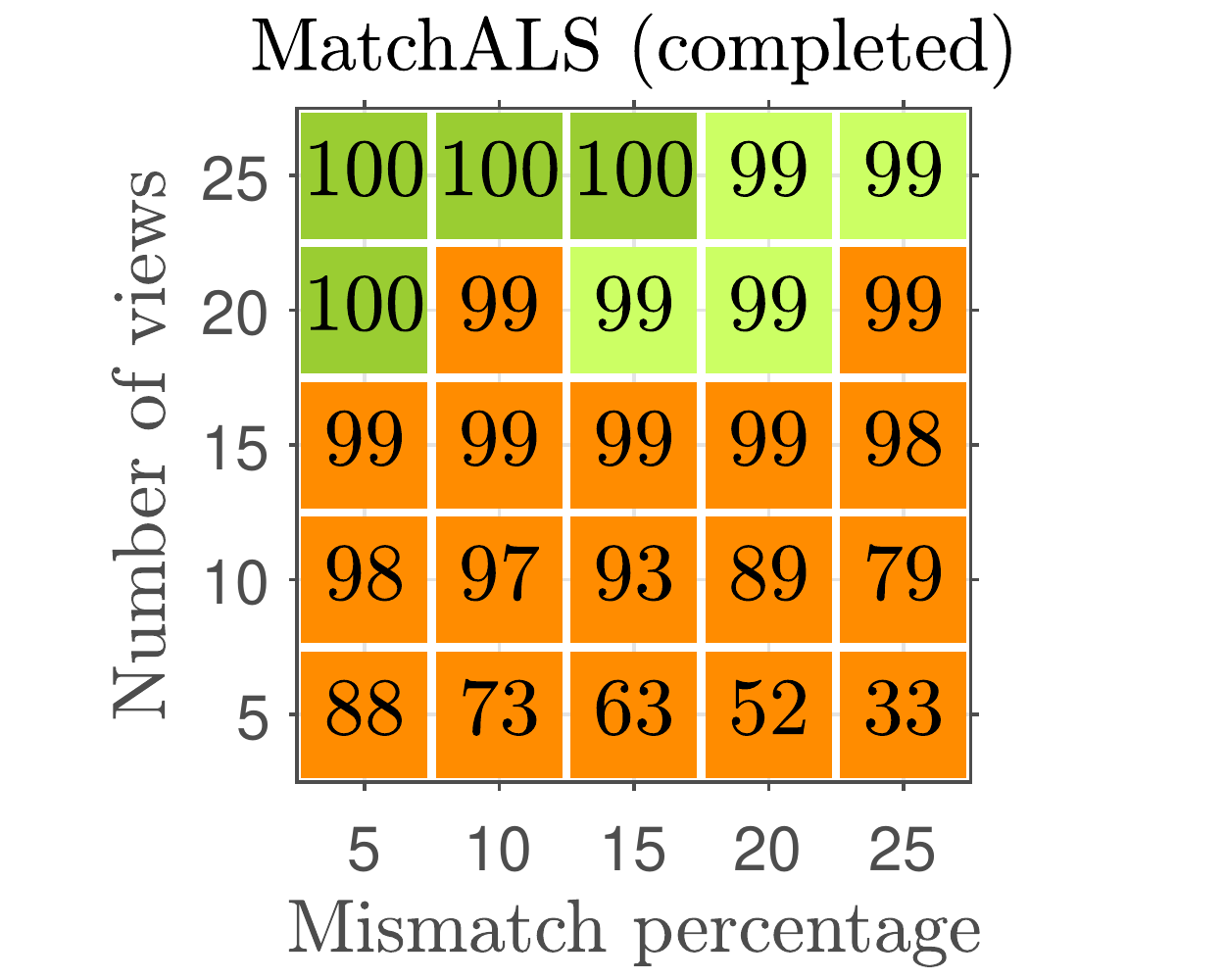}
\end{subfigure}%
\caption{(Best viewed in color) The average F$_1$ score of the inconsistent
  algorithms after making them cycle consistent by completing the graph's connected
  components for clique-centric applications (see Section~\ref{sec:apps}). 
	%
}
\label{fig:completed}	
\end{figure}

In addition, Fig.~\ref{fig:AgentCompare} demonstrates that the extended spectral algorithm, MatchEig, and MatchALS may return results that
violate the cycle consistency and/or distinctness constraints in moderate to high
noise regimes. 
Recall from Section~\ref{sec:apps} that although a cycle-inconsistent solution may
exhibit a high F$_1$ score in terms of individual associations, in
clique-centric applications its F$_1$ score can
dramatically decrease after completing the connected components of the
association graph (i.e., transitive closure).
This is demonstrated in Fig.~\ref{fig:completed} for MatchEIG and
MatchALS algorithms (compare Fig.~\ref{fig:completed} with Fig.~\ref{fig:AgentCompare}).
For example, the average F$_1$ score of MatchEIG with $10$ views and under
$15\%$ mismatch drops from $0.93$
(Fig.~\ref{fig:AgentCompare}) to  
$0.08$ (Fig.~\ref{fig:completed}). As discussed in Section~\ref{sec:apps}, here
the F$_1$ score of $0.93$ can be very misleading if the solution obtained by the algorithm is going
to be used for fusion in the context of clique-centric applications.

Among the algorithms that do not violate the consistency and distinctness
constraints, on average, MatchLift, NMFSync, and CLEAR have the highest F$_1$
scores.
The poor performance of QuickMatch is mainly due to the fact that this algorithm was
originally designed and tuned for matching image features based on
\emph{weighted} associations, whereas in our setting the associations are
binary.\footnote{Nonetheless, it is straightforward to generalize CLEAR and
other algorithms to the weighted case.}
In conclusion, synthetic comparisons demonstrate that CLEAR returns cycle
consistent solutions with high F$_1$ scores. 
In the next section, we evaluate the runtime and scalability of the algorithms
in real-world examples, where the total number of observations can reach several
thousands.

\begin{figure}
	\centering
	\begin{subfigure}[b]{0.142\textwidth}	\includegraphics[trim = 10mm 0mm 18mm 0mm, clip, width=1.0\textwidth] {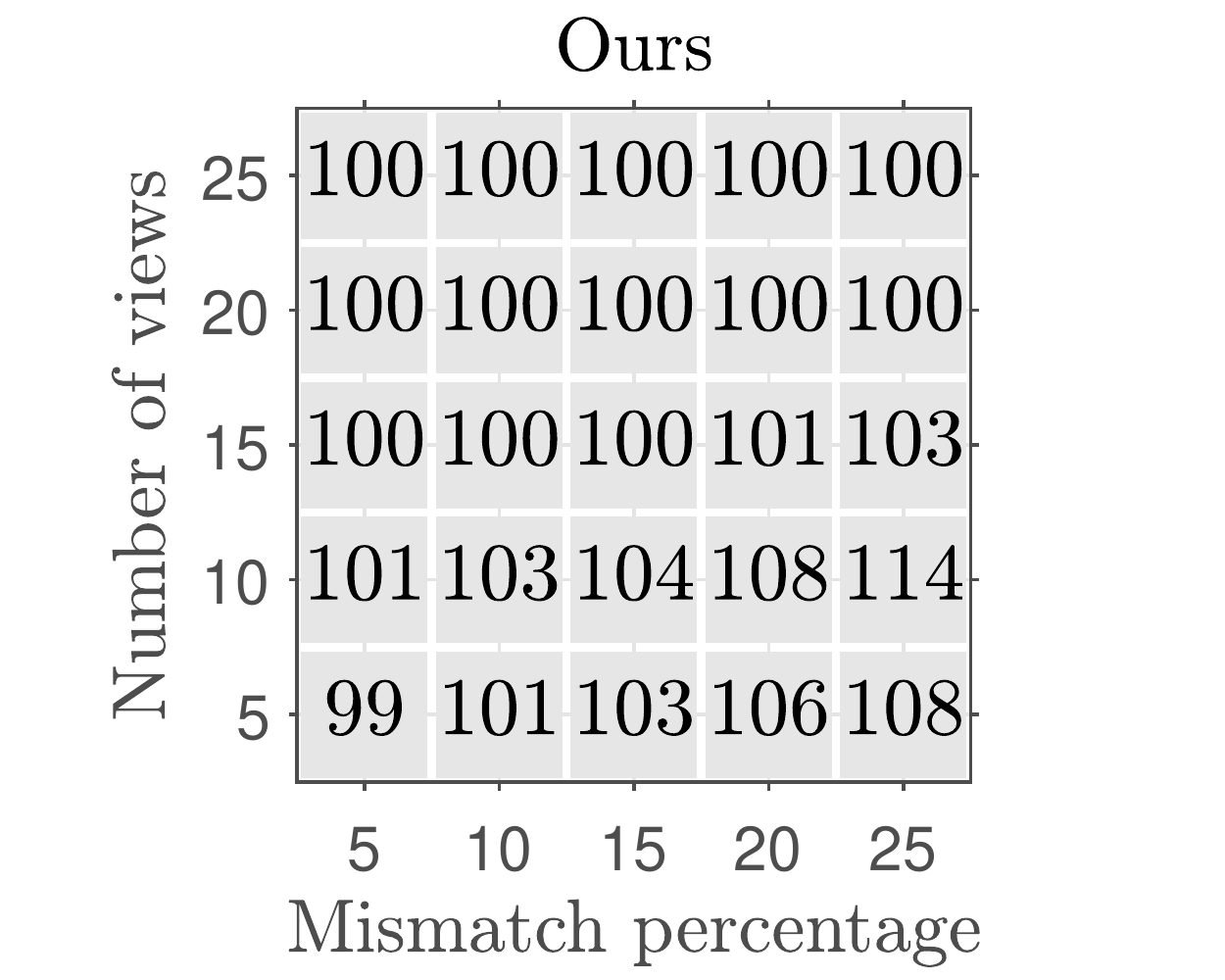}
	\end{subfigure}%
	~~~~~~~
	\begin{subfigure}[b]{0.142\textwidth}
		\includegraphics[trim = 10mm 0mm 18mm 0mm, clip, width=1.0\textwidth] {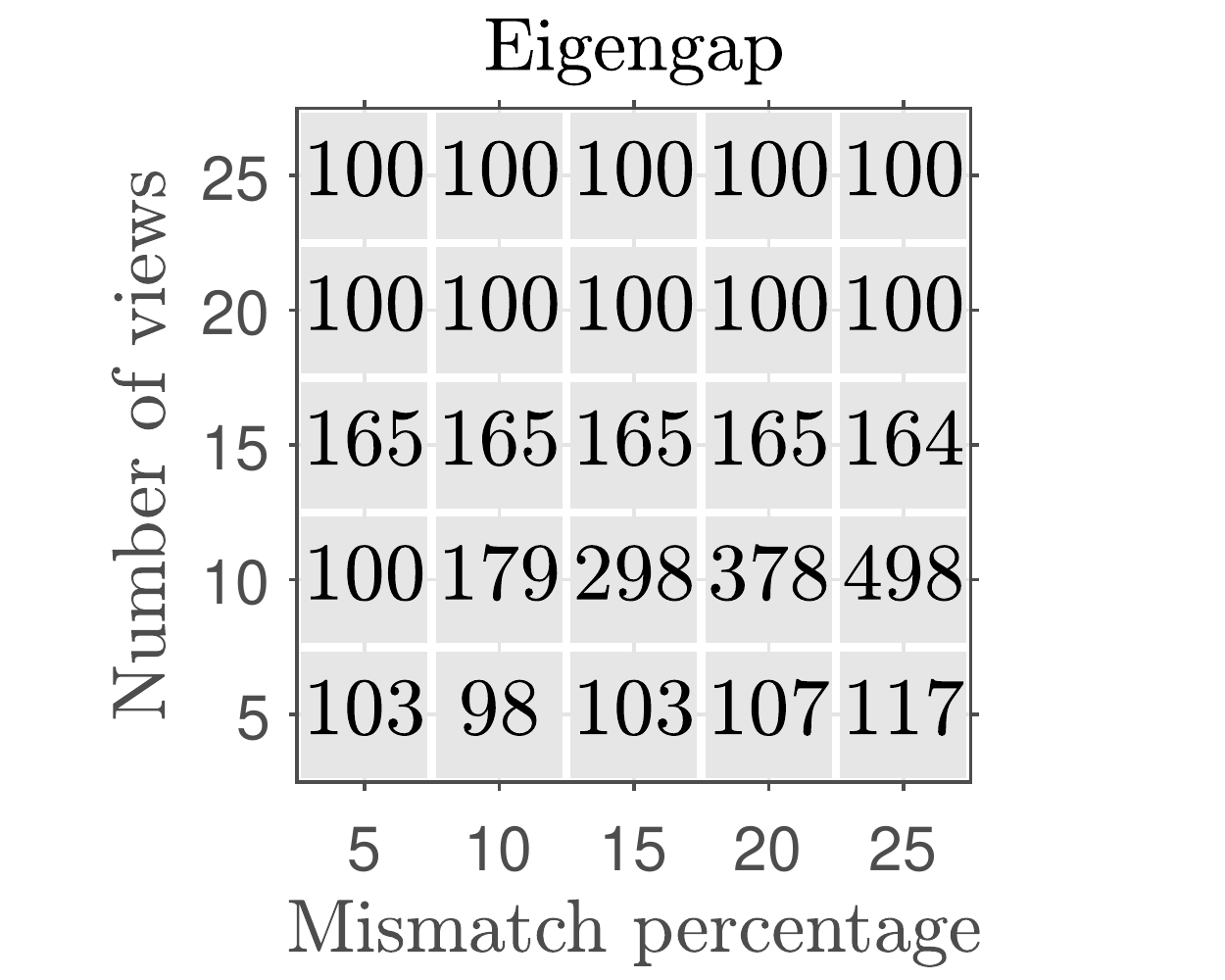}
	\end{subfigure}%
	\\  \vspace{0.05in}
	\begin{subfigure}[b]{0.142\textwidth}	\includegraphics[trim = 10mm 0mm 18mm 0mm, clip, width=1.0\textwidth] {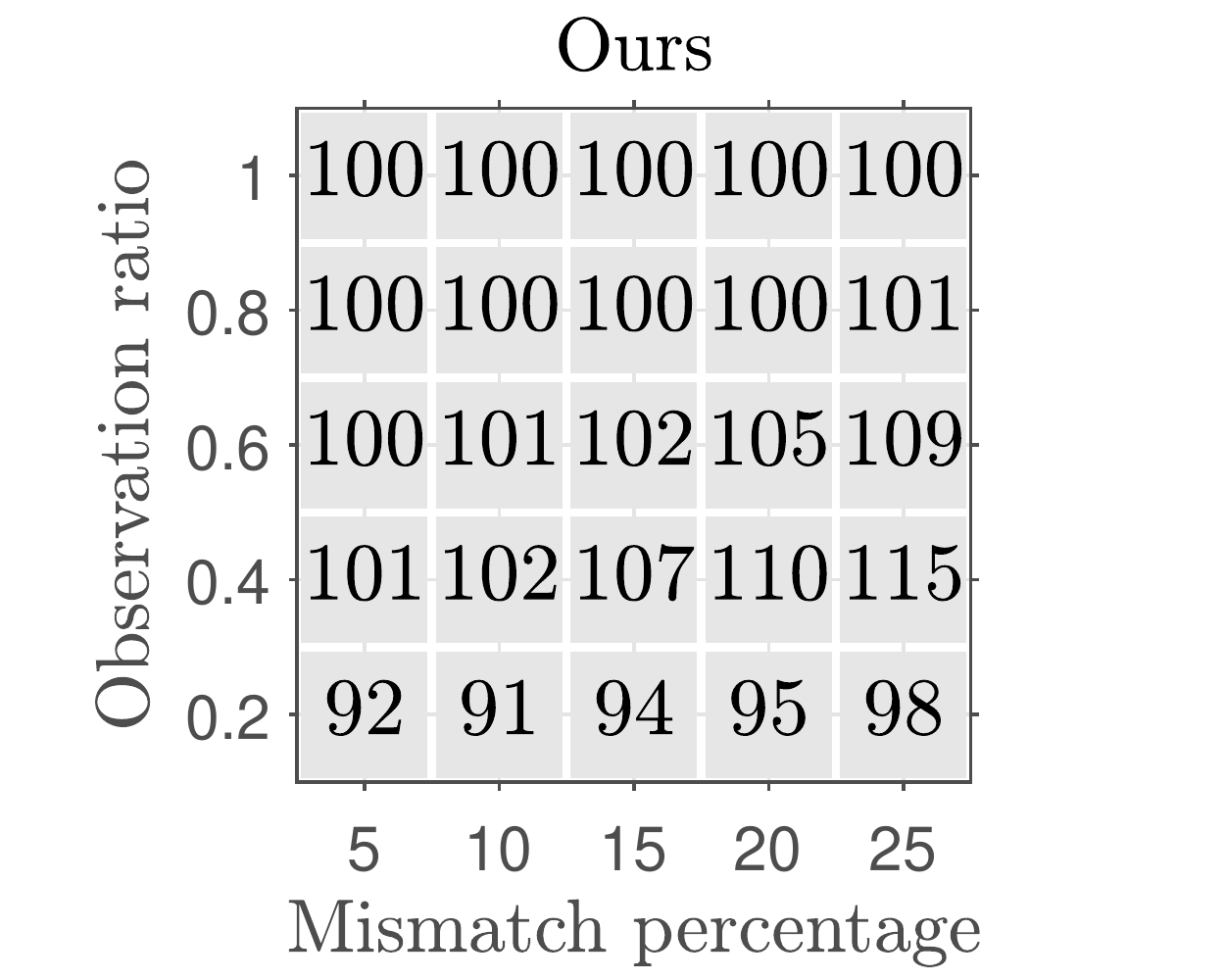}
	\end{subfigure}%
	~~~~~~~
	\begin{subfigure}[b]{0.142\textwidth}
		\includegraphics[trim = 10mm 0mm 18mm 0mm, clip, width=1.0\textwidth] {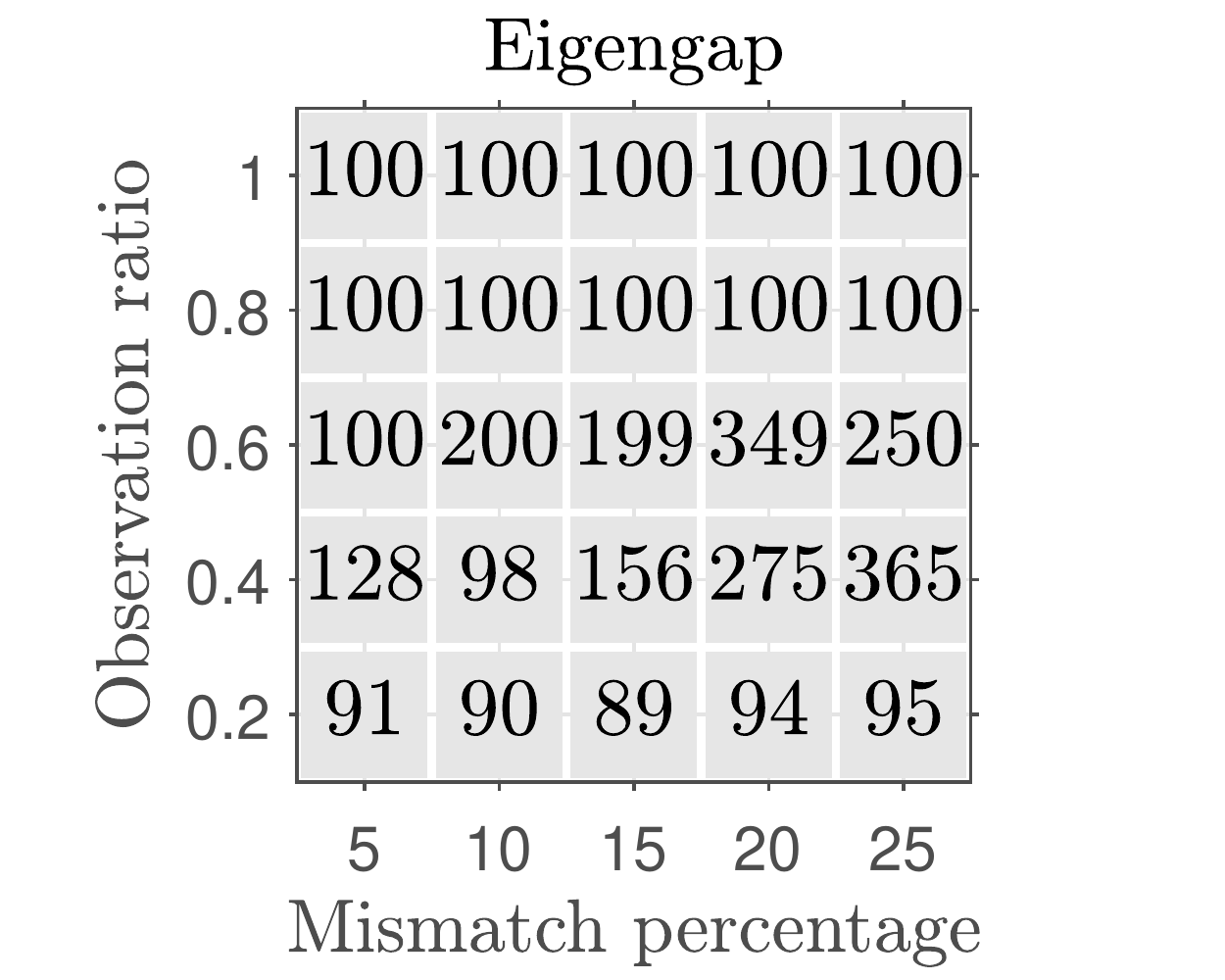}
	\end{subfigure}%
	\caption{The average of estimated universe sizes in the Monte
	  Carlo runs of Fig.~\ref{fig:AgentCompare} by CLEAR and the 
	  eigengap method based on the symmetric Laplacian. The closer to $100$, the better.}
	\label{fig:mEstimate}	
\end{figure}

Finally, we compare the estimated size of universe, obtained from  \eqref{eq:mEstimate},
with the eigengap method commonly used in the spectral graph clustering
literature (see Remark~\ref{rem:eigengap}).
The
results are reported in Fig.~\ref{fig:mEstimate}. The number written inside each
square is the average of estimated universe sizes (rounded) in the Monte Carlo runs of
Fig.~\ref{fig:AgentCompare}. The correct universe size is $100$. According to
the results depicted in Fig.~\ref{fig:mEstimate}, although both techniques
perform equally well under a high
signal-to-noise ratio (top two rows in each figure), the proposed approach is
more robust to noise and significantly outperforms the standard eigengap
heuristic (bottom three
rows in each figure).

\section{Experimental Results} \label{sec:Experiments}

\begin{figure*}[t!]
	\centering
	\begin{subfigure}[b]{0.2\textwidth}
		\includegraphics[trim = 10mm 0mm 18mm 0mm, clip, width=1.0\textwidth] {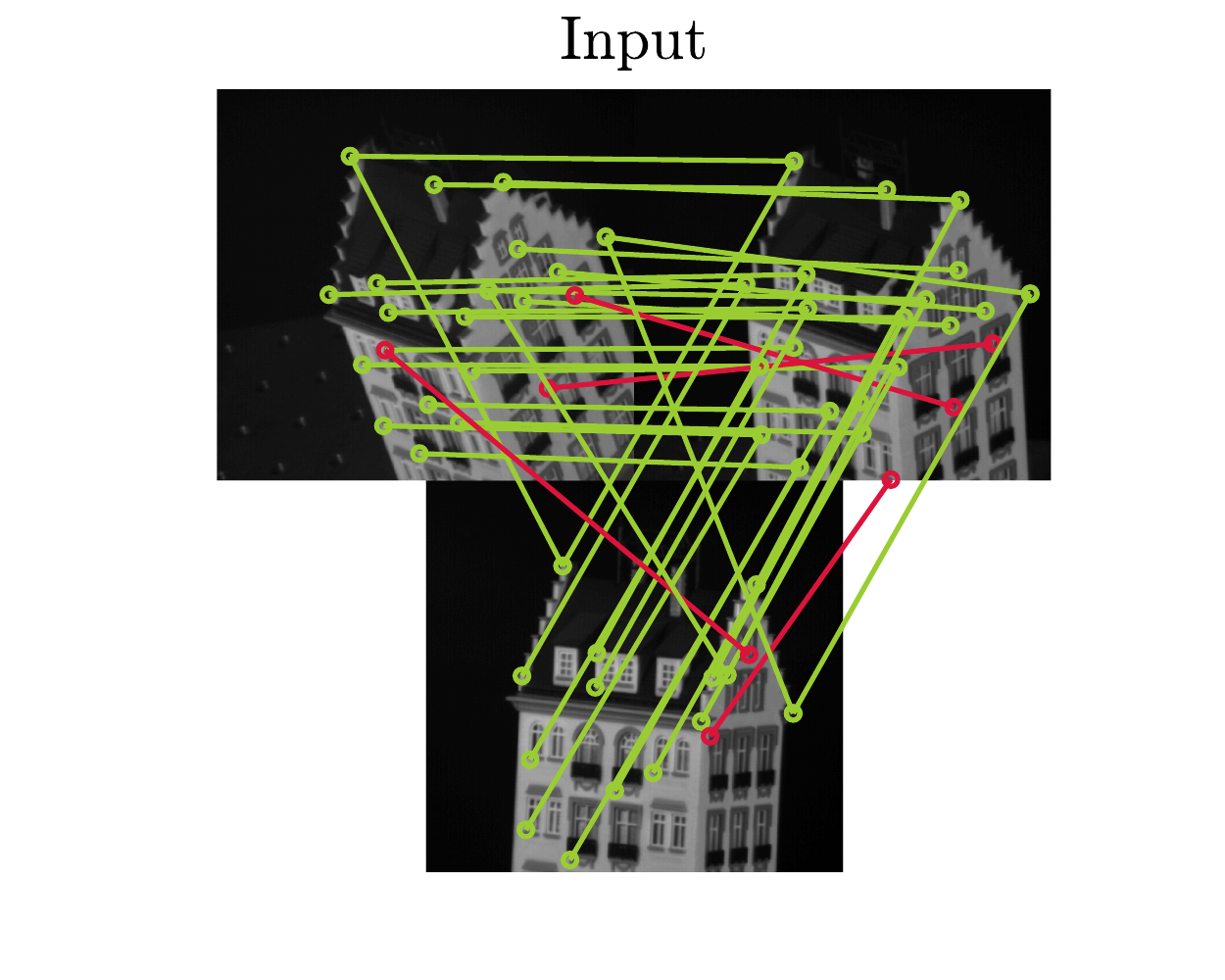}
	\end{subfigure}%
	\begin{subfigure}[b]{0.2\textwidth}
		\includegraphics[trim = 10mm 0mm 18mm 0mm, clip, width=1.0\textwidth] {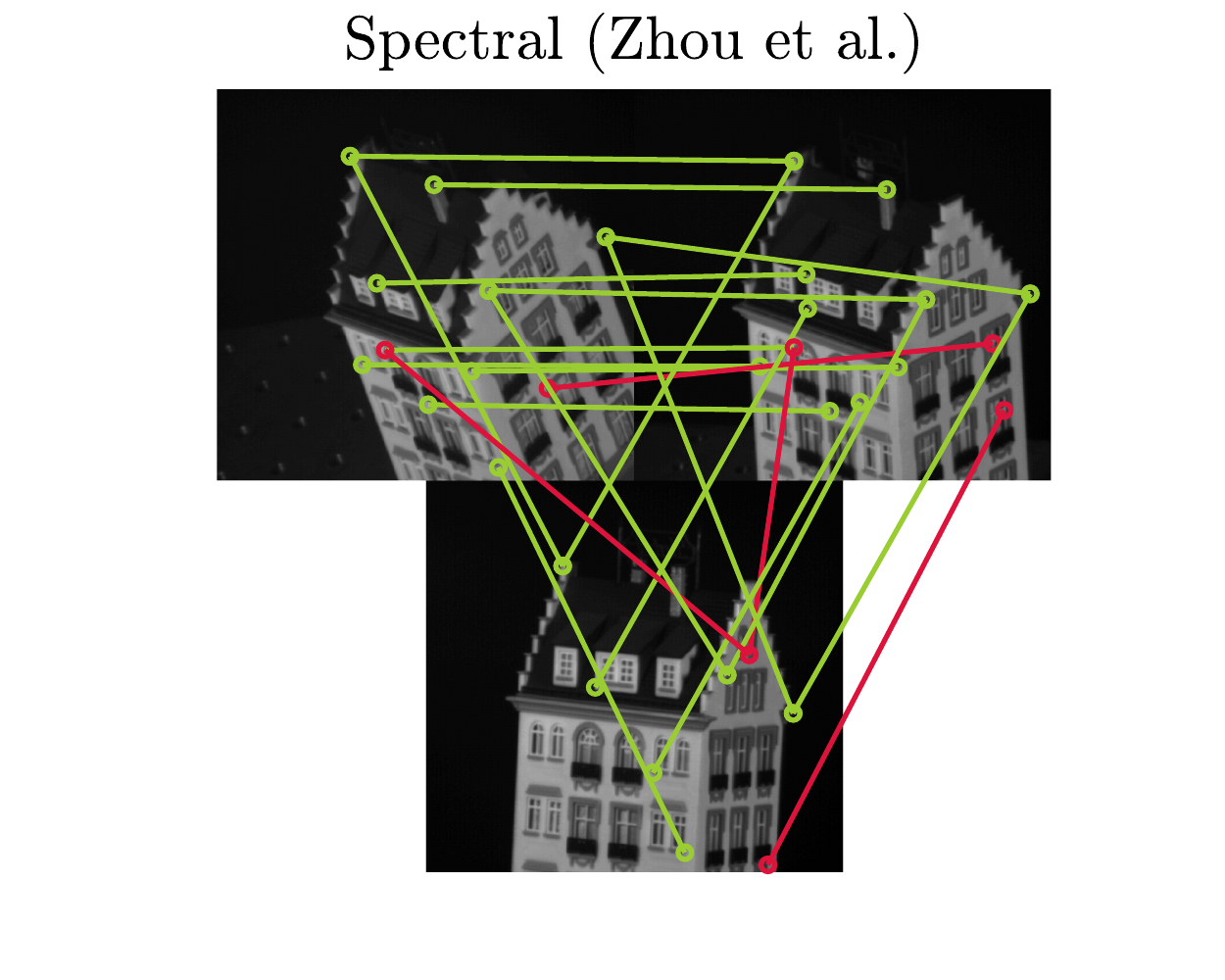}
	\end{subfigure}%
	\begin{subfigure}[b]{0.2\textwidth}
		\includegraphics[trim = 10mm 0mm 18mm 0mm, clip, width=1.0\textwidth] {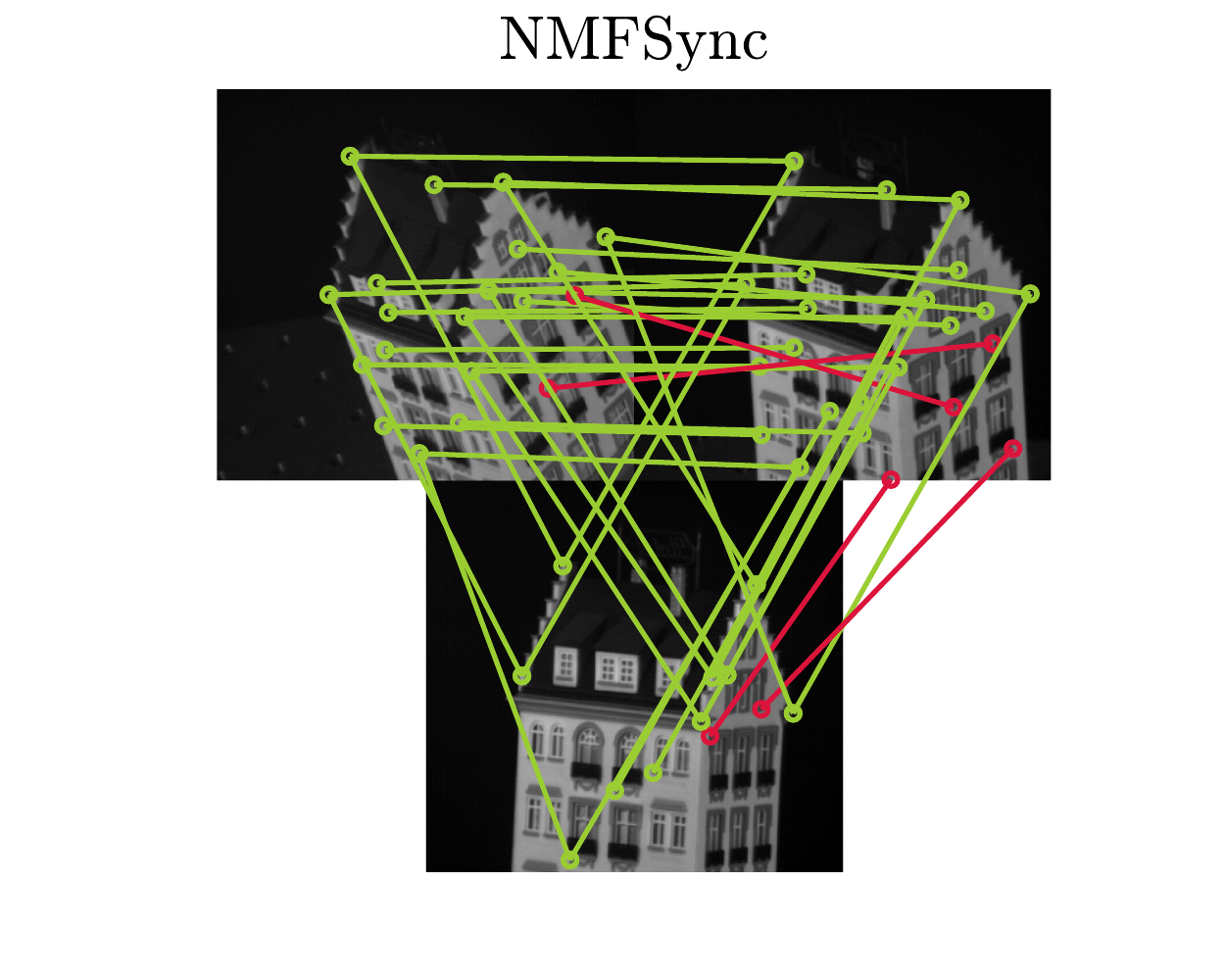}
	\end{subfigure}%
	\begin{subfigure}[b]{0.2\textwidth}
		\includegraphics[trim = 10mm 0mm 18mm 0mm, clip, width=1.0\textwidth] {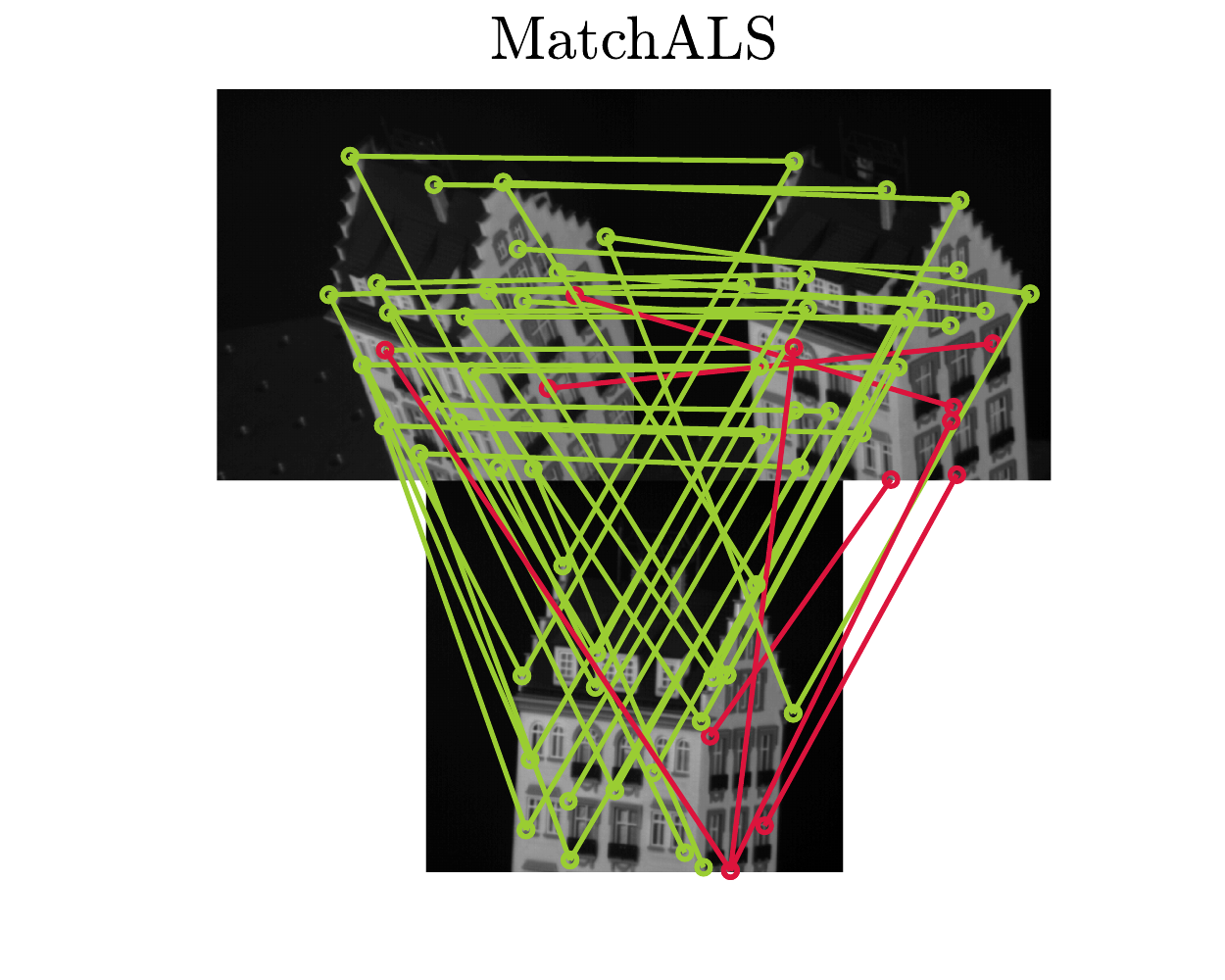}
	\end{subfigure}%
	\begin{subfigure}[b]{0.2\textwidth}
		\includegraphics[trim = 10mm 0mm 18mm 0mm, clip, width=1.0\textwidth] {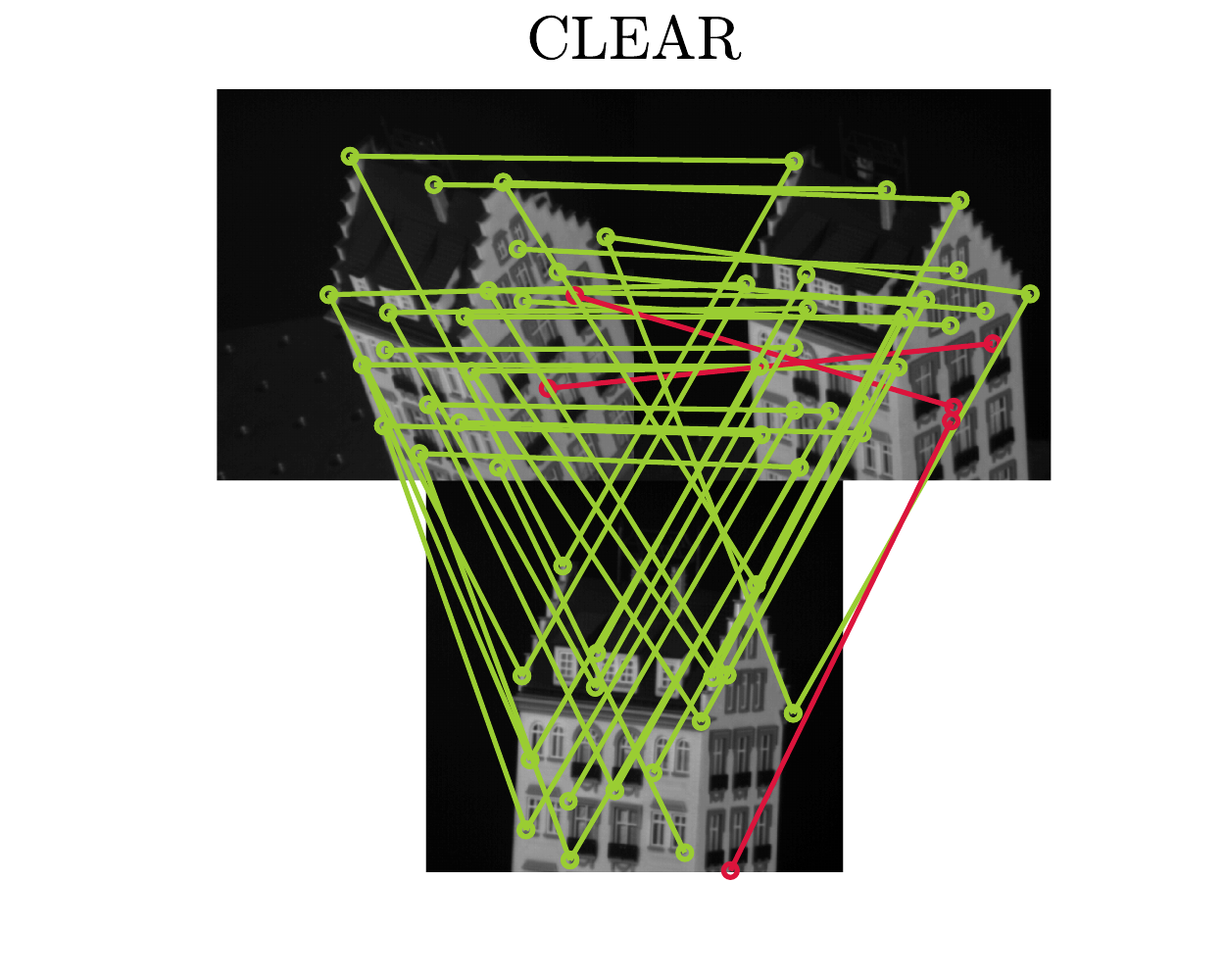}
	\end{subfigure}%
	\\ \vspace{-0.09in}
	\begin{subfigure}[b]{0.9\textwidth}
		\includegraphics[trim = 13mm 110mm 13mm 102mm, clip, width=1.0\textwidth] {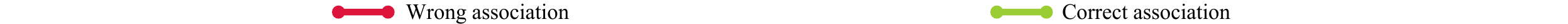}
	\end{subfigure}%
	\caption{(Best viewed in color) An example of matched feature points across
	three images of the CMU Hotel dataset. Input, obtained by matching features
	across image pairs independently, contains error and is inconsistent. CLEAR
	returns cycle-consistent results and improves the precision of the input.}
	\label{fig:HotelPics}	
\end{figure*}

To further evaluate the accuracy and speed of CLEAR in real-world robotics
applications, we
consider two scenarios, namely multi-image feature matching and map
fusion in landmark-based SLAM.
Feature matching datasets have become standard benchmarks for comparing the
performance of multi-way data association algorithms. Hence, we report the
results on two publicly available standard benchmark datasets, namely Graffiti\footnote{
  \href{http://www.robots.ox.ac.uk/~vgg/data/data-aff.html}{http://www.robots.ox.ac.uk/\textasciitilde
  vgg/data/data-aff.html}} and CMU Hotel.\footnote{ \href{http://pages.cs.wisc.edu/~pachauri/perm-sync/}{http://pages.cs.wisc.edu/\textasciitilde pachauri/perm-sync/}}
The aim of our experimental comparisons is to 1) compare the runtime of
algorithms; 2) evaluate the precision/recall for the returned solutions.

\begin{figure}[t!]
	\centering
	\includegraphics[trim = 0mm 0mm 0mm 5mm, clip, width=0.5\textwidth] {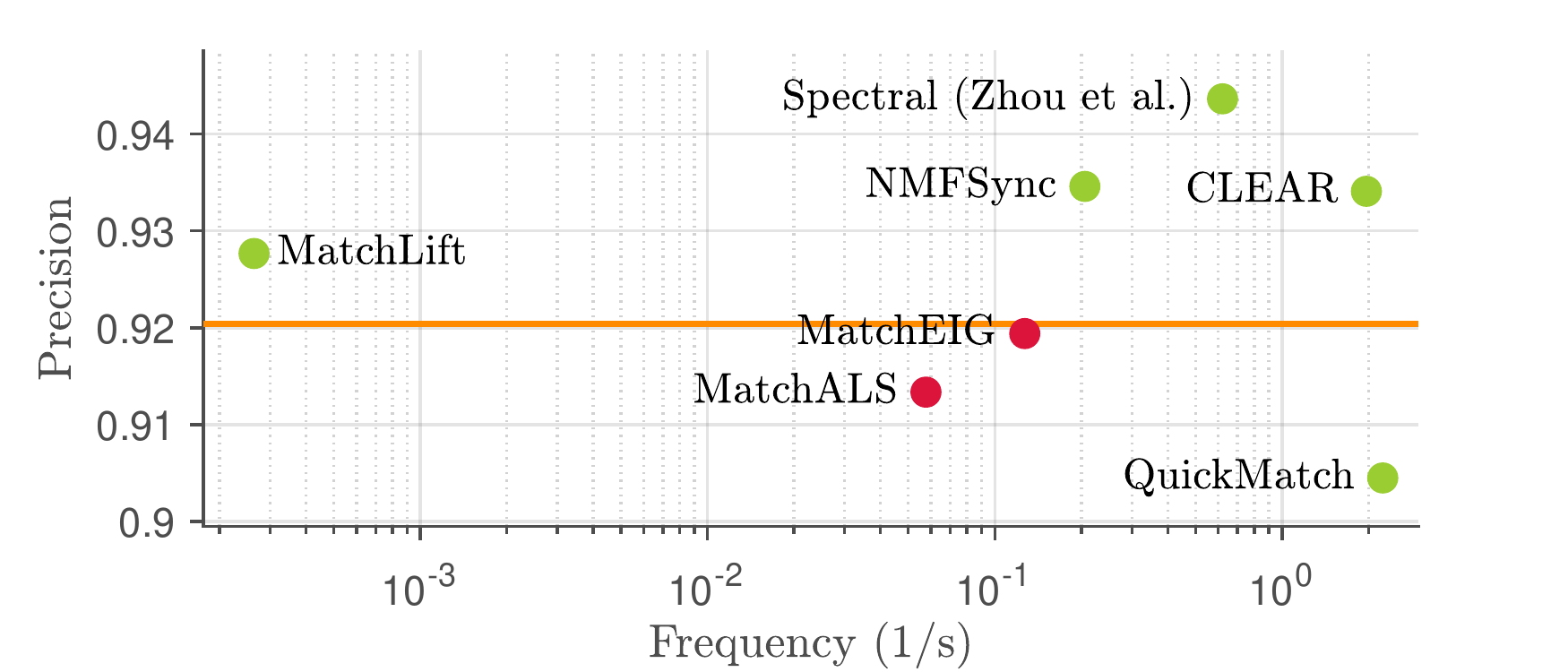}
	\caption{Precision vs. frequency (the inverse of execution time) in CMU
		Hotel dataset. The frequency axis has a logarithmic scale (CLEAR is about
		3x faster than Spectral, 10x faster than NMFSync,
		7500x faster than MatchLift). The precision of the input (based on
		individual edges and for edge-centric applications) is denoted
		by the orange line; see Section~\ref{sec:apps}. Cycle-consistent/inconsistent outputs are respectively denoted by \tc[white, fill=YellowGreen]{3pt} and \tc[white, fill=red]{3pt}. The closer to the top-right corner, the better.}
	\label{fig:Hotel}
\end{figure}

\subsection{CMU Dataset}

The CMU hotel dataset consists of $101$ images. The ground truth provided by this dataset consists of $30$ feature points per image and their correct associations. These feature points are visible across all images, leading to a total of $3030$ features across all images. 
Due to the large ratio of the number of images ($101$ images) to the number of feature points per image ($30$ features), this dataset represents scenarios where observations have high redundancy.
To obtain the input for algorithms, we compute the SIFT descriptor \cite{Lowe2004} of each feature point using the VLFeat library\footnote{\href{http://www.vlfeat.org/}{http://www.vlfeat.org/} } \cite{Vedaldi2010}. 
The standard \verb+vl_ubcmatch+ routine in VLFeat is used to match feature points across image pairs based on the Euclidean distance between their descriptor vectors.
By taking this input (as a $3030 \times 3030$ aggregate association matrix), each algorithm returns an output which is then compared with the ground truth to evaluate its accuracy. 
We further record the execution time of each algorithm.
All results are based on Matlab implementation of algorithms on a machine with an Intel Core i7-7700K CPU @ 4.20GHz and 16GB RAM.

Fig.~\ref{fig:HotelPics} shows an example of three images in the CMU hotel sequence, where feature points and their associations across images are shown for the input and the output of four algorithms.
Note that the input associations, which are obtained by matching features on
image pairs, are cycle inconsistent and contain errors. The output of the algorithms should ideally identify and remove these errors based on the cycle consistency principle.

Fig.~\ref{fig:Hotel} reports the precision (i.e., number of correct matches
divided by the total number of returned matches) versus the frequency (rate) of the solutions returned by algorithms. The frequency (i.e., the inverse of execution
time) indicates the number of times an algorithm can run in one second. Due to
the large difference between the runtimes of the algorithms, the frequency axis is scaled logarithmically. 
The precision of the input is indicated by the orange line on the plot and
approximately has the value of $0.92$. Note that this value is calculated based
on individual edges and thus is only meaningful for edge-centric applications; see
Section~\ref{sec:apps}.
Solutions that were not cycle consistent are colored in red. 
An ideal algorithm should have a high frequency (i.e., small runtime) and a high
precision output (i.e., based on individual edges and for edge-centric applications). 
Among the cycle-consistent algorithms, QuickMatch is the fastest, however, the returned solution does not improve the precision of the input.
CLEAR, Spectral, NMFSync, and MatchLift algorithms improve the precision, while
CLEAR has a higher frequency: CLEAR is about 3x faster than Spectral, 10x faster than NMFSync, 7500x faster than MatchLift.   

The faster runtime of CLEAR is due to 1) the structure of the input association
graph, which consists of several disjoint connected components (this graph
consists of $81$ connected components, where the largest component has $297$
vertices). This structure is exploited by the proposed eigendecomposition
approach, which uses the BFS algorithm to find the spectrum of the graph as the
union of its connected components' spectra. 2) The projection technique, which
uses a suboptimal sorting strategy (instead of, e.g., the Hungarian algorithm)
to improve the speed while ensuring consistency and distinctness. More
specifically, running CLEAR with the Hungarian algorithm results in the same output (i.e., the same value for precision and recall), however, the execution time increases from $0.5$s to $0.7$s.

\begin{figure} [t!]
	\centering
	\includegraphics[trim = 0mm 0mm 0mm 5mm, clip, width=0.5\textwidth] {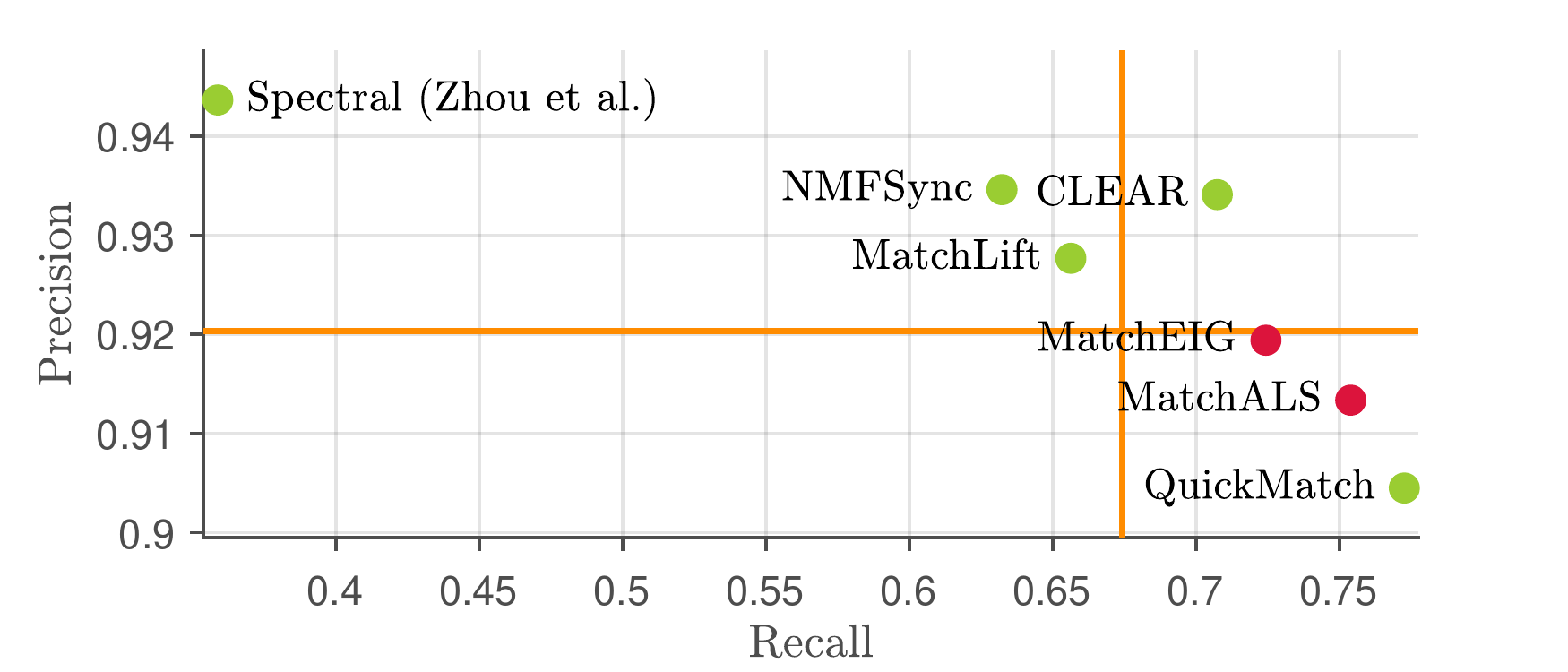}
	\caption{Precision vs.\ recall in CMU Hotel dataset. Precision and recall of
		input (based on individual edges and for edge-centric applications) are
		denoted by the orange lines; see Section~\ref{sec:apps}.
		The closer to the top-right corner, the better.}
	\label{fig:Hotel2}
\end{figure}

Fig.~\ref{fig:Hotel2} reports the precision and recall of returned solutions. An ideal solution simultaneously has high precision and recall. 
The output of the Spectral algorithm has the highest precision and lowest recall. On the other hand, the output of QuickMatch has the highest recall and lowest precision. In comparison, the output of CLEAR shows a balanced precision versus recall.

We note that the precision and recall of MatchEig after making its solution cycle
consistent by completing the association graph's connected components
(Section~\ref{sec:apps}) become $0.67$ and $0.8$, respectively. Similarly,
MatchALS's output after completion takes the precision and recall of $0.73$ and $0.76$,
respectively. 
This sharp drop in precision underlines the importance of taking cycle
consistency into account in evaluating multi-view matching algorithms for clique-centric applications.

\subsection{Graffiti Dataset}

The Graffiti dataset consists of six images, each taken from a different viewpoint of a textured planar wall. Due to the large difference between the viewpoints, this dataset is particularly challenging for feature point detection/matching algorithms (thus, pairwise associations have a lower precision compared to the CMU hotel dataset).
The dataset provides ground truth homography transformations between the
viewpoints. We use the VLFeat library to extract the SIFT feature points for
each image. To obtain the ground truth associations, the provided homography matrices are used to match the extracted features (correct matches must satisfy the planar homography mapping \cite[see (5.35)]{Ma2012}). 
To make sure that ground truth associations are error-free, we only take feature points and associations that are cycle consistent across all images and discard the rest.
These associations are further visually inspected to ascertain that they do not contain mismatches. 
The number of feature points retained after this process ranges from $313$ to $657$ per image. The total number of feature points across all images is $3176$.
Unlike the CMU hotel dataset, the Graffiti dataset has a small ratio of the number of images to the number of feature points per image. Thus, it represents scenarios where observations have little to no redundancy.

\begin{figure}[t!]
	\centering
	\includegraphics[trim = 0mm 0mm 0mm 5mm, clip, width=0.5\textwidth] {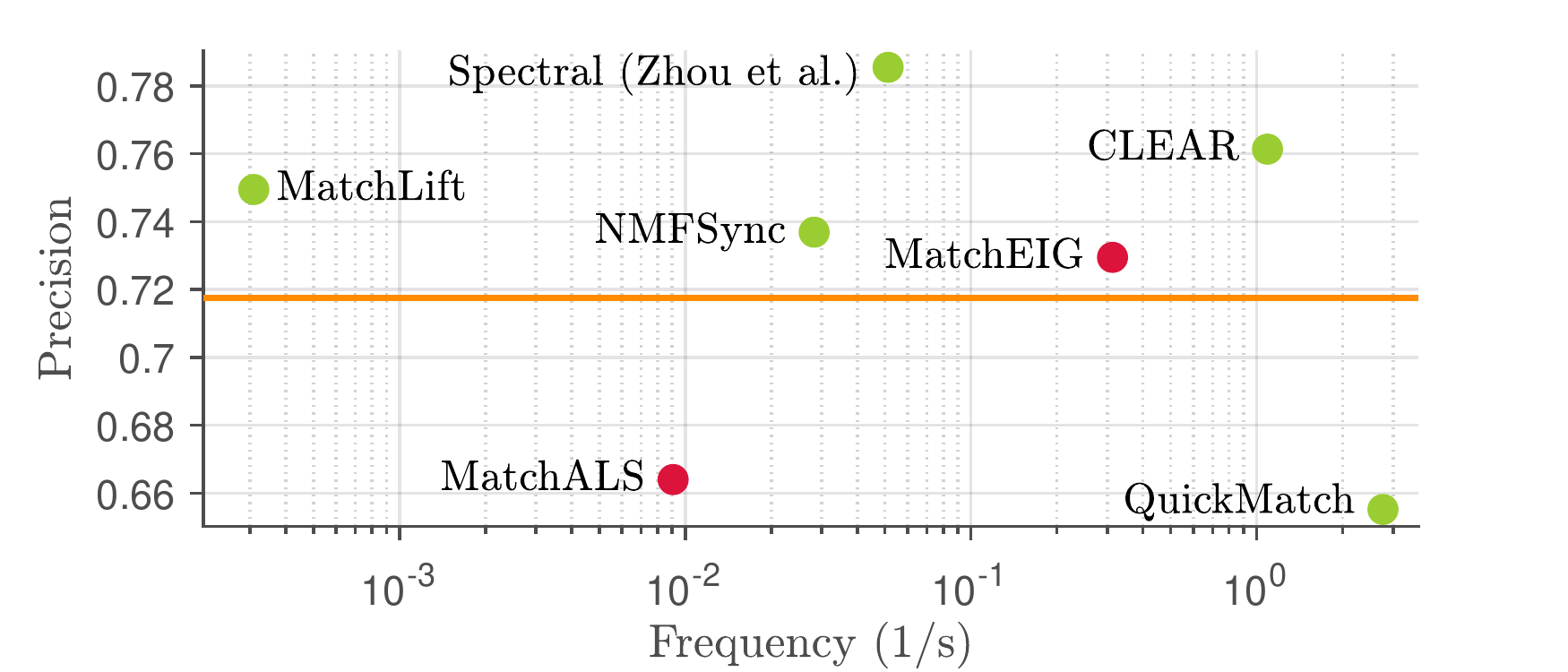}
	\caption{Precision vs. frequency (the inverse of execution time) in Graffiti dataset. The frequency axis has a logarithmic scale (CLEAR is about 21x faster than Spectral, 39x faster than NMFSync, 3800x faster than MatchLift). Precision of the input is denoted by the orange line. Cycle consistent and inconsistent outputs are respectively denoted by \tc[white, fill=YellowGreen]{3pt} and \tc[white, fill=red]{3pt}. The closer to the top-right corner, the better.}
	\label{fig:Graffiti}
\end{figure}

The precision and frequency of algorithms is reported in Fig.~\ref{fig:Graffiti}.
Among the cycle-consistent algorithms, QuickMatch is the fastest, however, it
does not improve the precision of the input computed based on individual edges
and for edge-centric applications (Section~\ref{sec:apps}). CLEAR improves the
precision and is considerably faster compared to the other algorithms that improve the input's precision: about 21x faster than Spectral, 39x faster than NMFSync, 3800x faster than MatchLift.

In the Graffiti dataset, the input association graph consists of $1506$
connected components, where the largest component has $22$ vertices. Running
CLEAR with the Hungarian algorithm results in an output with the same value for precision and recall (up to three decimals), however, the execution time of the algorithm increases considerably from $0.92$s to $49.5$s.

\begin{figure}[t!]
	\centering
	\includegraphics[trim = 0mm 0mm 0mm 5mm, clip, width=0.5\textwidth] {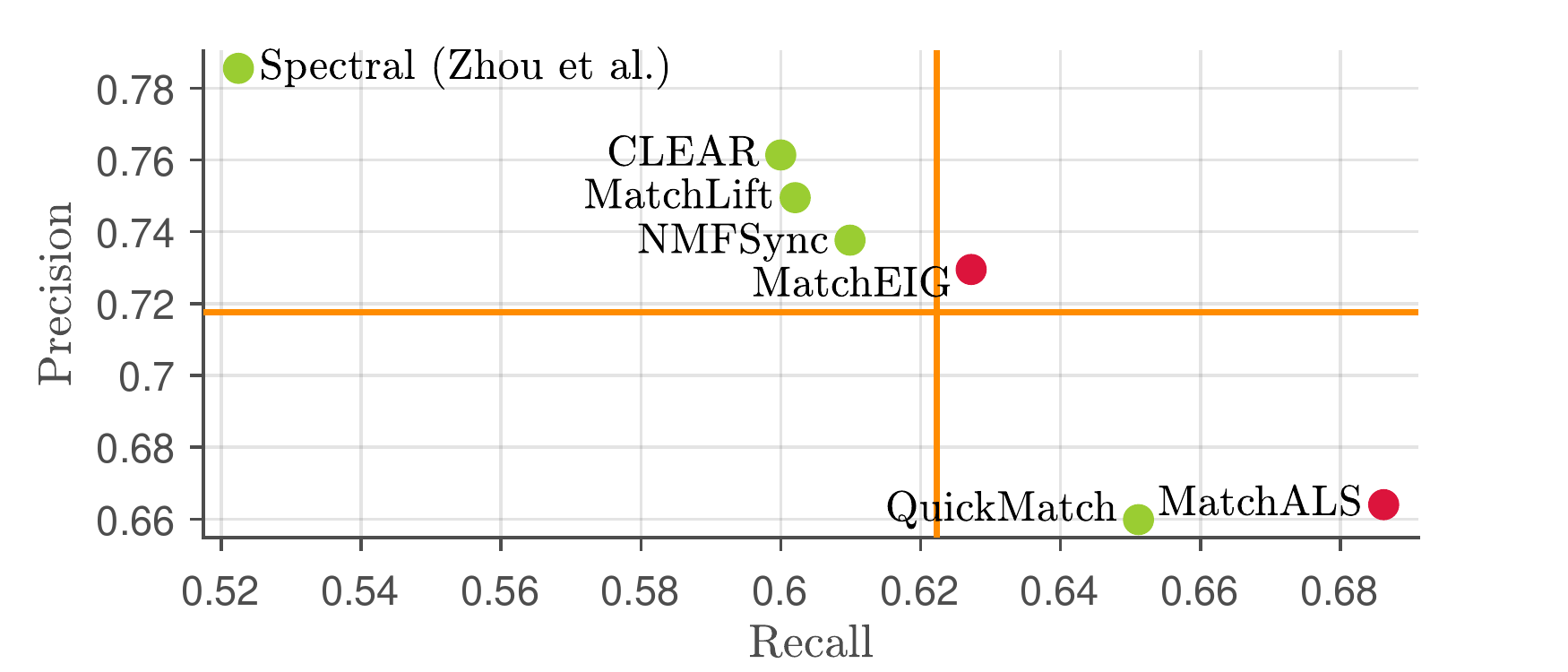}
	\caption{Precision vs. recall. in Graffiti dataset. Precision and recall of input are denoted by the orange lines.
		The closer to the top-right corner, the better.}
	\label{fig:Graffiti2}
\end{figure}

The precision and recall of returned solutions are reported in Fig.~\ref{fig:Graffiti2}. 
Among cycle-consistent algorithms, the Spectral algorithm has the highest
precision and lowest recall, while QuickMatch has the highest recall and lowest precision. In comparison, CLEAR, MatchLift, and NMFSynch have a balanced precision versus recall.
The precision and recall of MatchEig after making its solution cycle
consistent (for clique-centric applications) become
$0.53$ and $0.69$, respectively. Similarly, MatchALS's output after completion takes the precision and recall of $0.54$ and $0.69$, respectively.  
Once again, the difference between these values and those reported in
Fig.~\ref{fig:Graffiti2} highlights the importance of taking cycle
consistency into account in evaluating multi-view matching algorithms for clique-centric applications.

\subsection{Forest Landmark-based SLAM Dataset}

\begin{figure} [t]
	\centering
	\includegraphics[trim = 0mm 20mm 0mm 50mm, clip, width=\columnwidth] {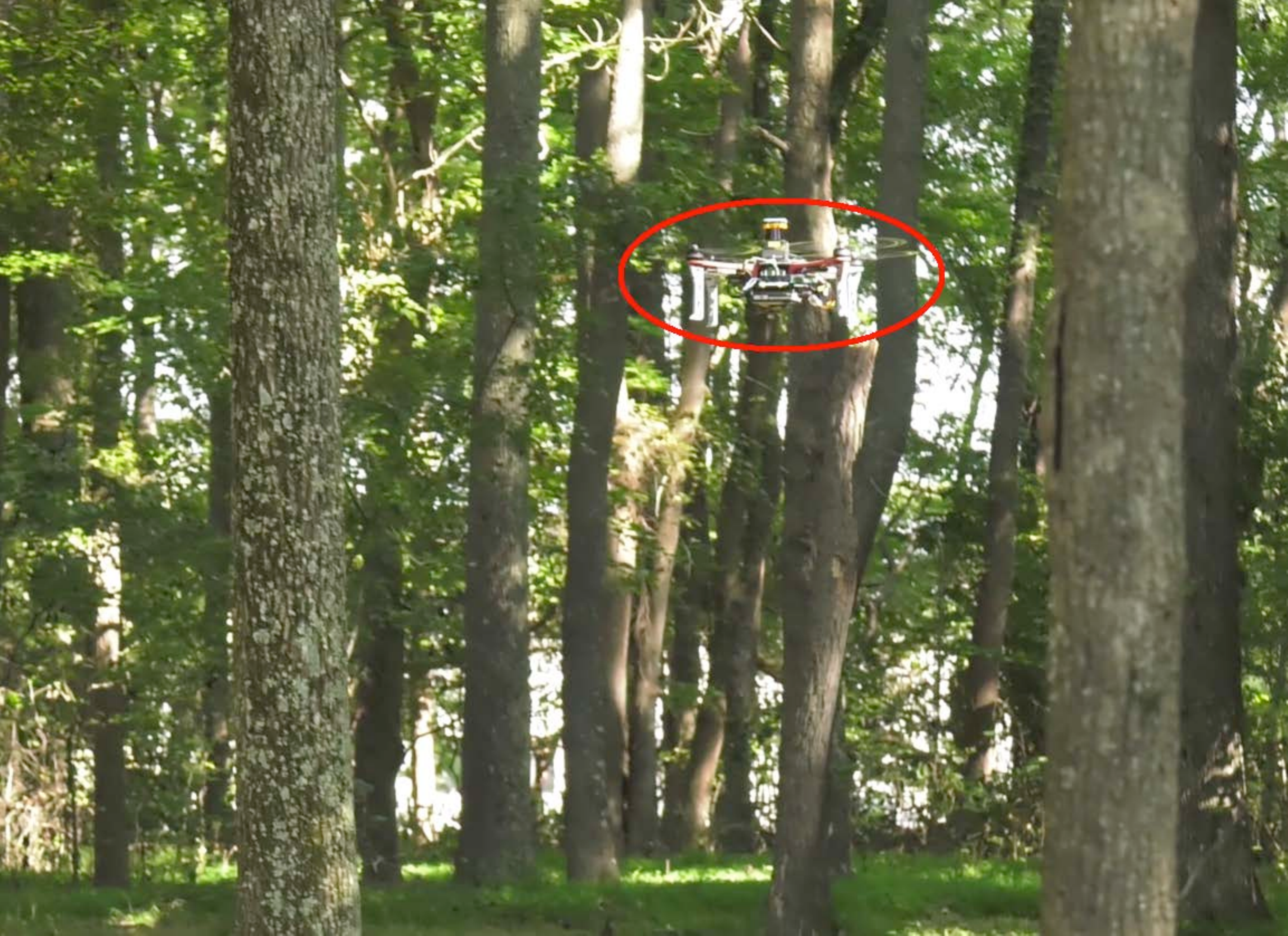}
	\caption{Single UAV autonomous exploration at NASA LaRC. The vehicle (highlighted in red) performs landmark-based SLAM based on detected trees in order to estimate its position within the forest. }
	\label{fig:slam_forest}
\end{figure}

Map fusion is an important clique-centric application of the multi-view matching problem in
single/multi-robot SLAM \cite{aragues2011consistent}. The goal in this problem is to
identify unique landmarks across a given set of local maps (created by one or
multiple robots) in order to fuse the corresponding measurements
in the landmark-based SLAM back-end.\footnote{Here, each local map represents a ``view'' in the
  multi-view matching problem. In practice, local maps may
represent one or multiple frames, and may be built by one or multiple
robots.}
In this section, we report the performance of CLEAR in the context of map fusion
based on a SLAM dataset collected in the forest
at the NASA Langley Research Center (LaRC) \cite{TianISER2018}. In this dataset,
a single unmanned aerial vehicle equipped with an inertial measurement unit (IMU) and a 2D LIDAR is tasked with autonomously exploring an area
under the tree canopy (Fig.~\ref{fig:slam_forest}). 
The exploration mission lasts $120$ seconds.
As the vehicle traverses the forest, it performs LIDAR-inertial odometry by fusing IMU measurements with incremental motion estimates from the iterative closest point algorithm at $40$~Hz.
In addition, the vehicle also uses a customized detector to identify trees from the LIDAR scans at a rate of $1$~Hz. 
The objective is to correctly match and fuse identical tree landmarks detected during the exploration, and subsequently optimize the landmark positions and vehicle trajectory inside a landmark-based SLAM framework.

To obtain the initial pairwise data association, we apply the correspondence graph matching algorithm \cite{BaileyICRA2000} that associates two sets of landmarks based on their local configurations.
Crucially, we note that this process does not use any global pose estimates, and thus is not affected by drift in the LIDAR-inertial odometry. 
Due to the presence of spurious detections and the lack of informative descriptors (e.g., SIFT), the initial data association matrix (of dimension $1091 \times 1091$) contains many mismatches and is not cycle consistent. 
We thus call CLEAR and other multi-view matching algorithms to achieve cycle consistency.
Recall from our discussion in Section~\ref{sec:apps} that map fusion is inherently a clique-centric application.
Therefore, we make any inconsistent data associations cycle consistent by completing the connected components in the association graph (Section~\ref{sec:apps}). In addition, we also introduce a baseline algorithm that directly completes the connected components in the input associations.

Since ground truth data association is not available, we adopt the following alternative performance metrics. 
A pair of associated trees is classified as either a \emph{definite negative} or a \emph{potential positive}, based on whether their
distance as estimated by the LIDAR-inertial odometry is higher than a threshold of $2$~m.
We note that these definitions are precise assuming that the threshold value of $2$~m accounts for the drift in the LIDAR-inertial odometry.\footnote{Since the vehicle is flying at a low speed ($2$~m/s) for a relative short amount of time $120$~s, we expect the estimation drift at any time is reasonably bounded.}
Since the number of definite negatives (denoted by DN) is an underestimate of the true number of mismatches, and the number of potential positives (denoted by PP) is an overestimate of the true number of correct matches,
we can further calculate an upper bound on the true precision as follows,
%
\begin{equation}
\bar{\text{P}}  \eqdef \frac{\text{PP}}{\text{DN} + \text{PP}} .
\label{eq:Pbar_def}
\end{equation}
%
We note that for landmark-based SLAM, the number of definite negatives (DN) is particularly important, since it is well known that any false data association could inflict catastrophic impact on the final solution. 
Therefore, an ideal data association should contain no definite negatives, or equivalently achieve a value of $100\%$ for $\bar{\text{P}}$ (upper bound on precision).

\begin{figure*}[t]
	\centering
	\begin{subfigure}{0.23\textwidth}
		\includegraphics[width=1.0\textwidth]{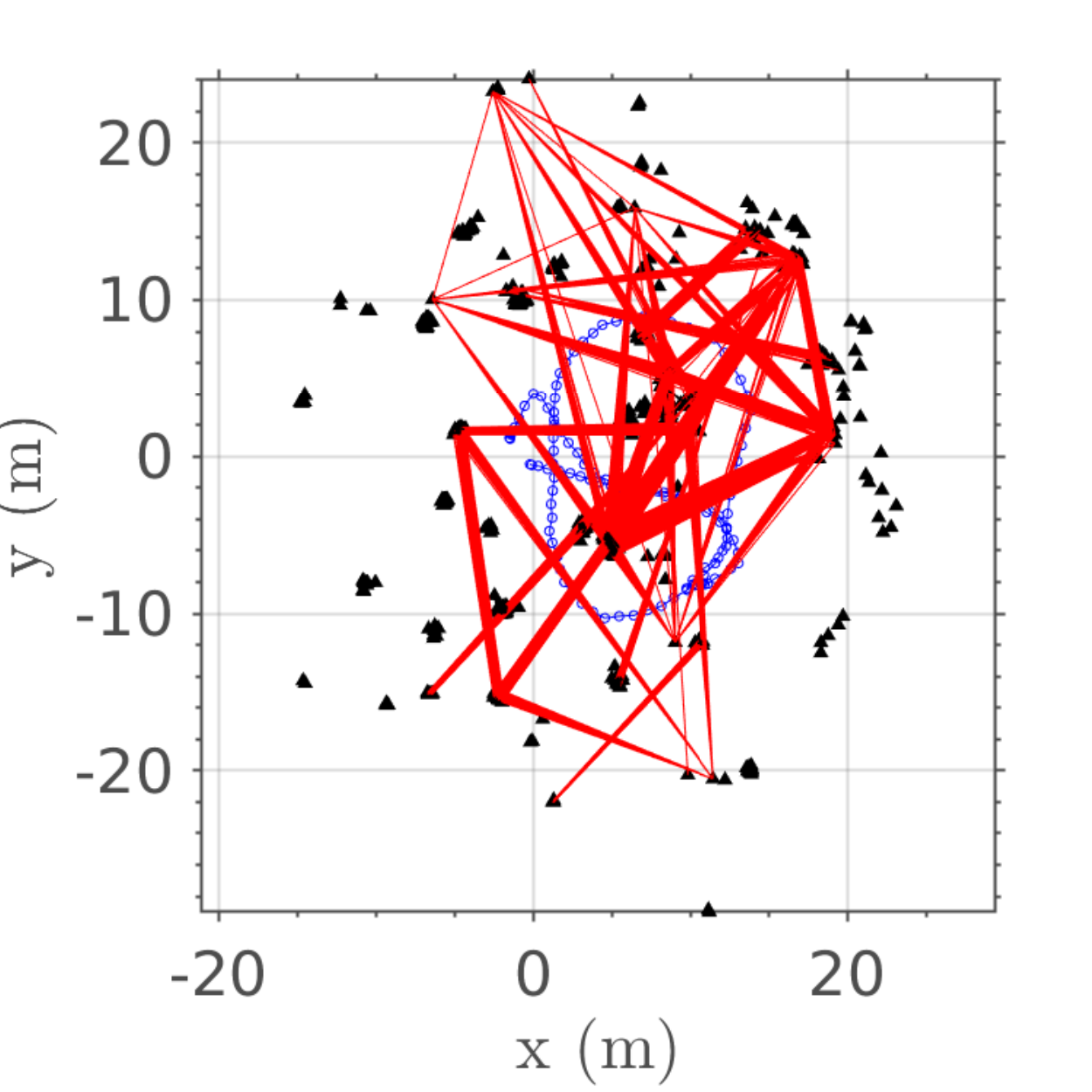}
		\caption{Baseline}
		\label{fig:SLAM_input}
	\end{subfigure}
	~
	\begin{subfigure}{0.23\textwidth}
		\includegraphics[width=1.0\textwidth]{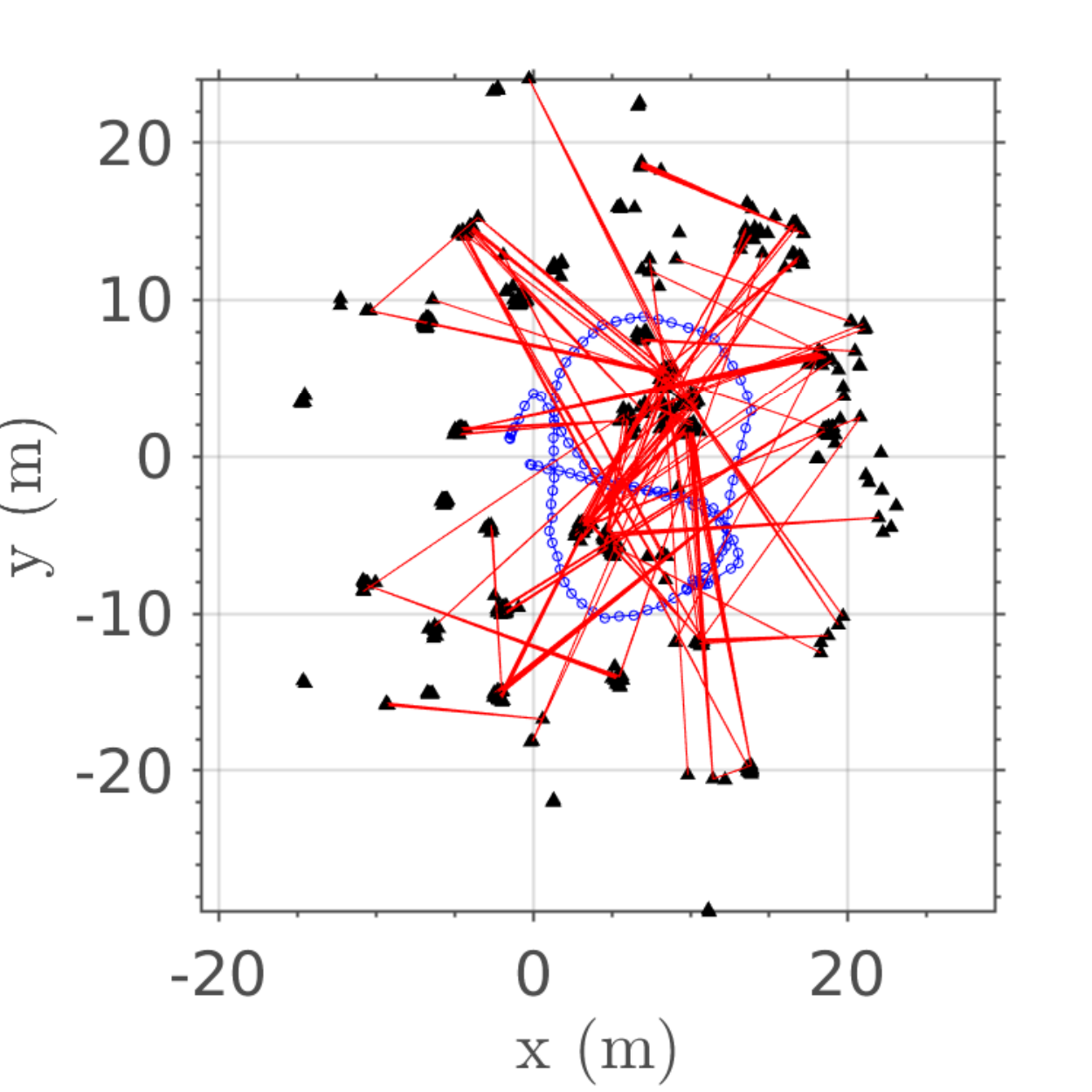}
		\caption{NMFSync}
		\label{fig:SLAM_NMF}
	\end{subfigure}
	~
	\begin{subfigure}{0.23\textwidth}
		\includegraphics[width=1.0\textwidth]{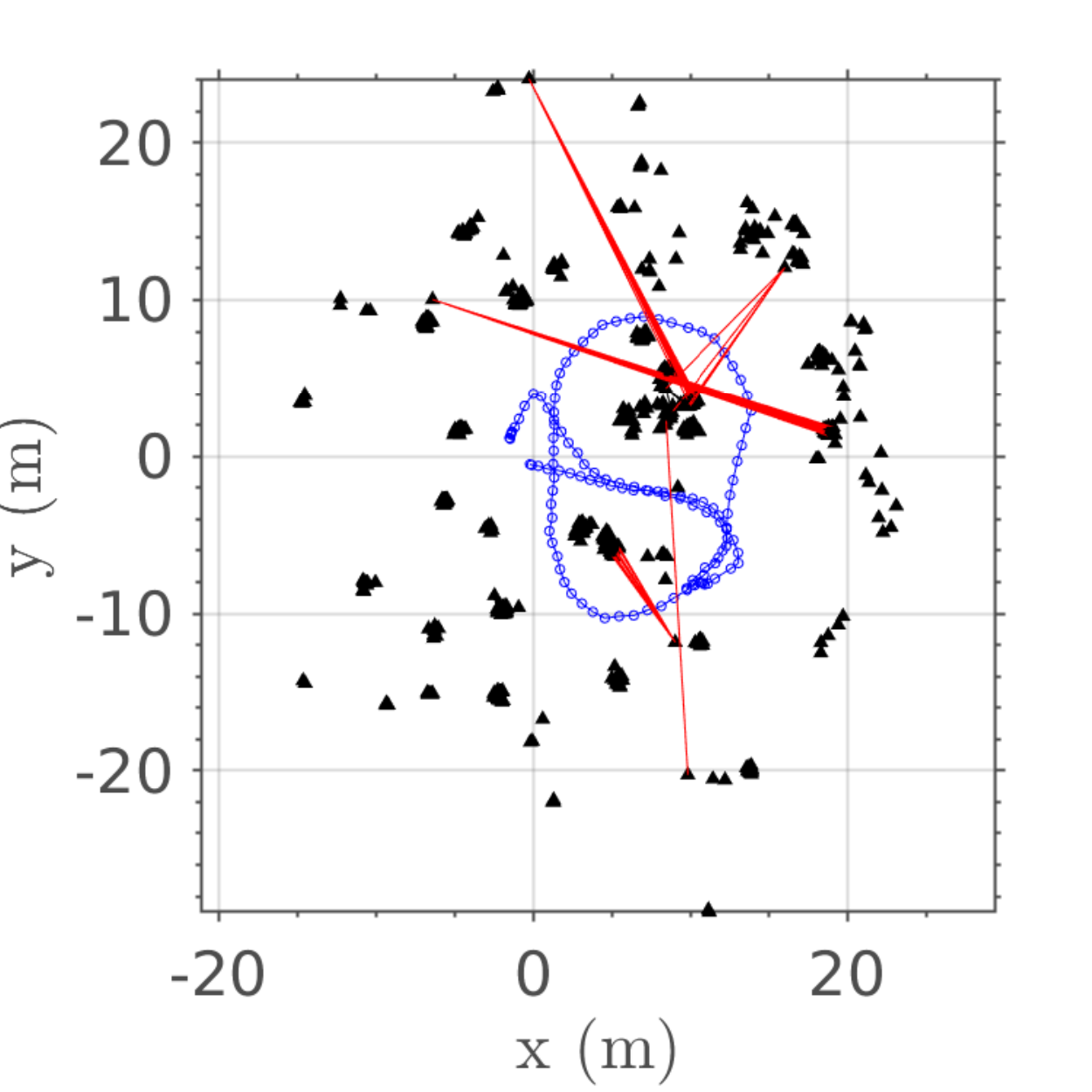}
		\caption{MatchALS (completed)}
	\end{subfigure}
	~
	\begin{subfigure}{0.23\textwidth}
		\includegraphics[width=1.0\textwidth]{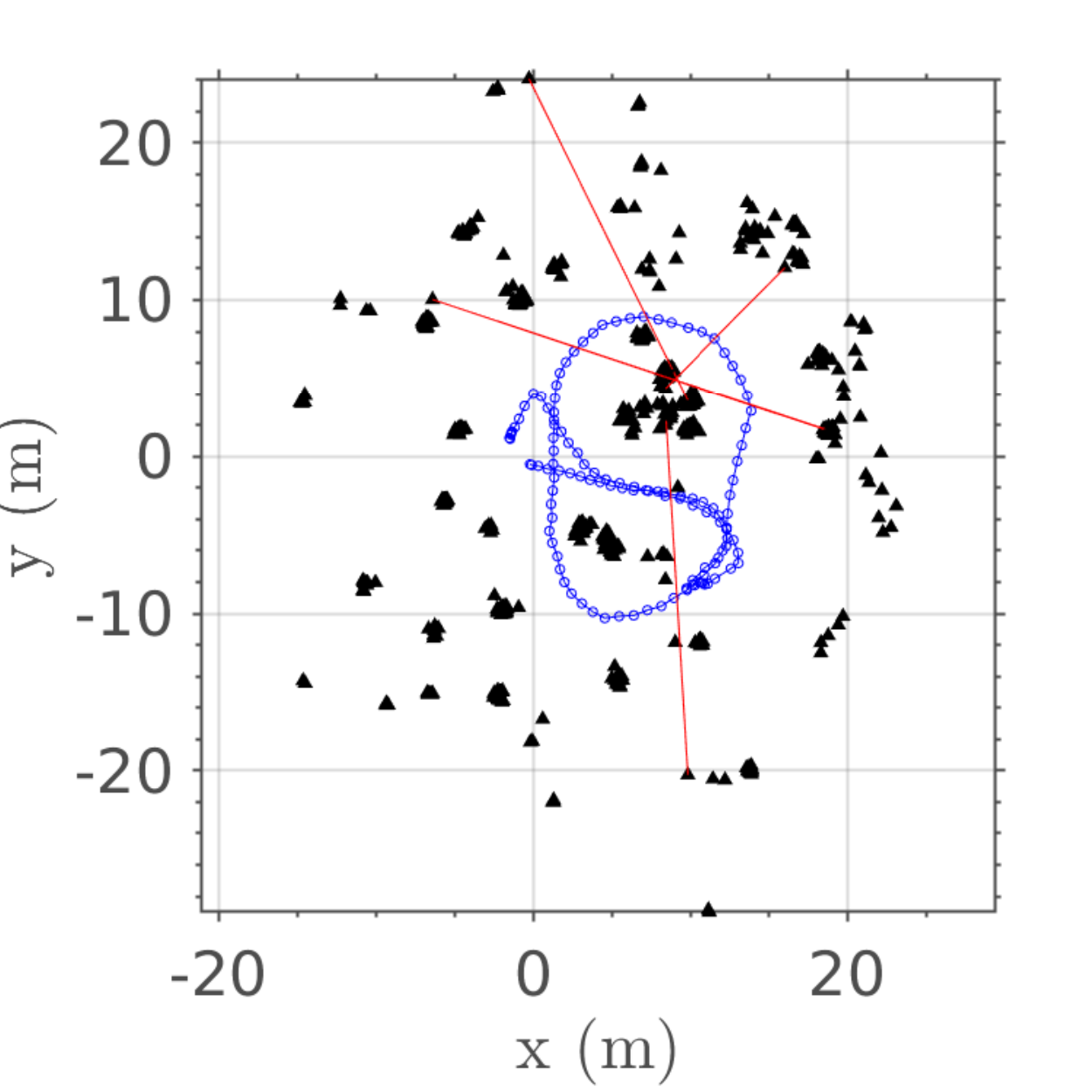}
		\caption{MatchLift }
	\end{subfigure}
	\\
	\begin{subfigure}{0.23\textwidth}
		\includegraphics[width=1.0\textwidth]{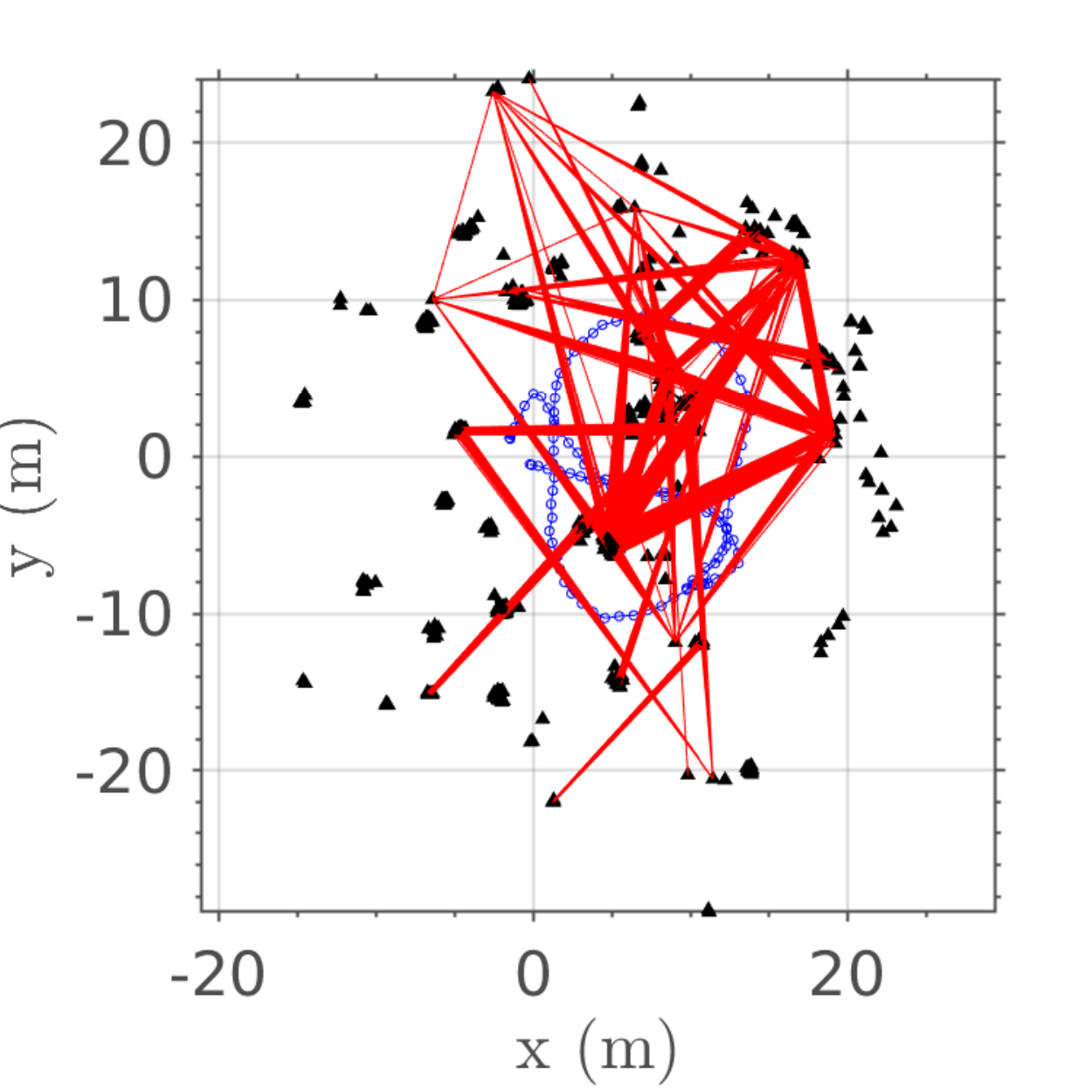}
		\caption{MatchEIG (completed)}
	\end{subfigure}
	~
	\begin{subfigure}{0.23\textwidth}
		\includegraphics[width=1.0\textwidth]{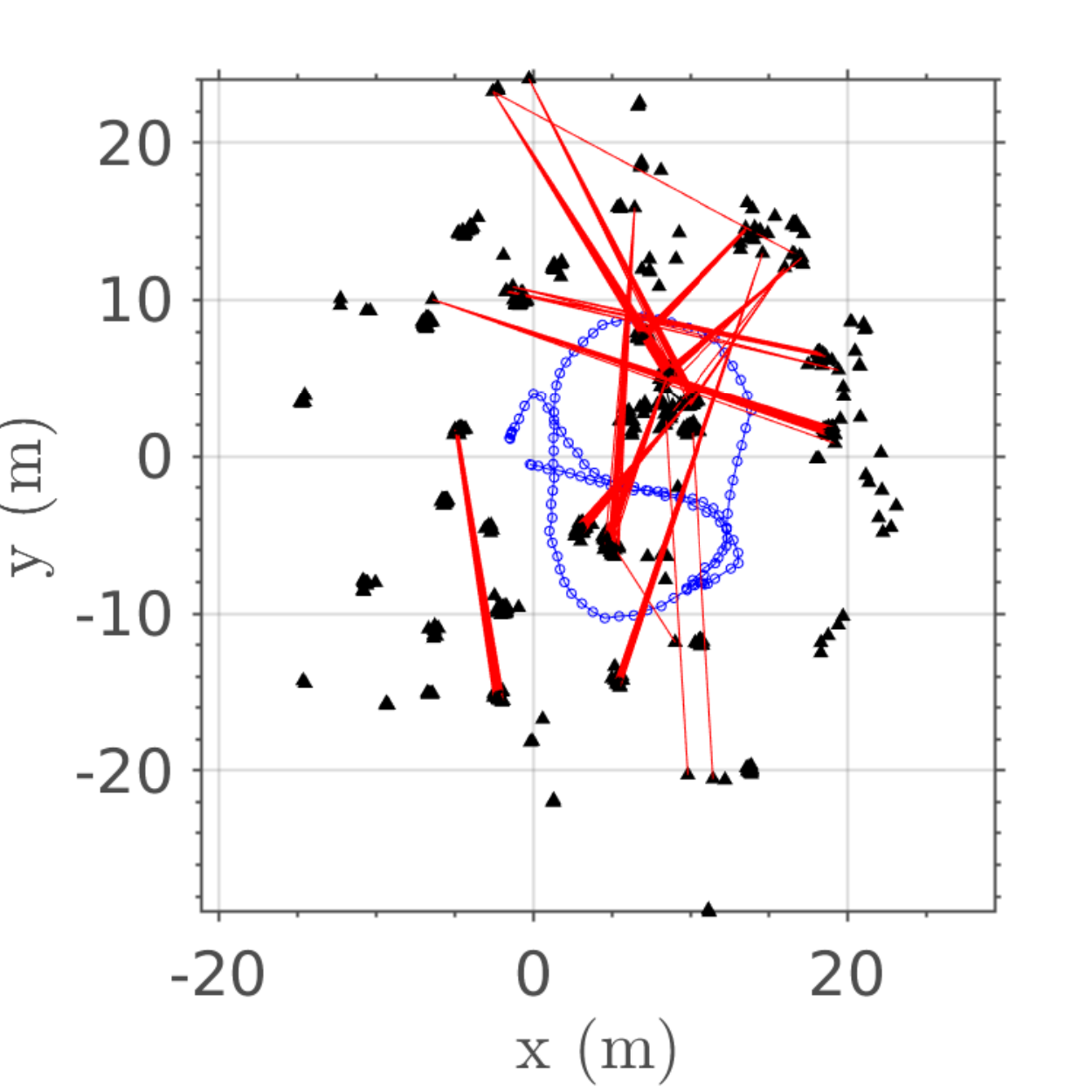}
		\caption{QuickMatch }
	\end{subfigure}
	~
	\begin{subfigure}{0.23\textwidth}
		\includegraphics[width=1.0\textwidth]{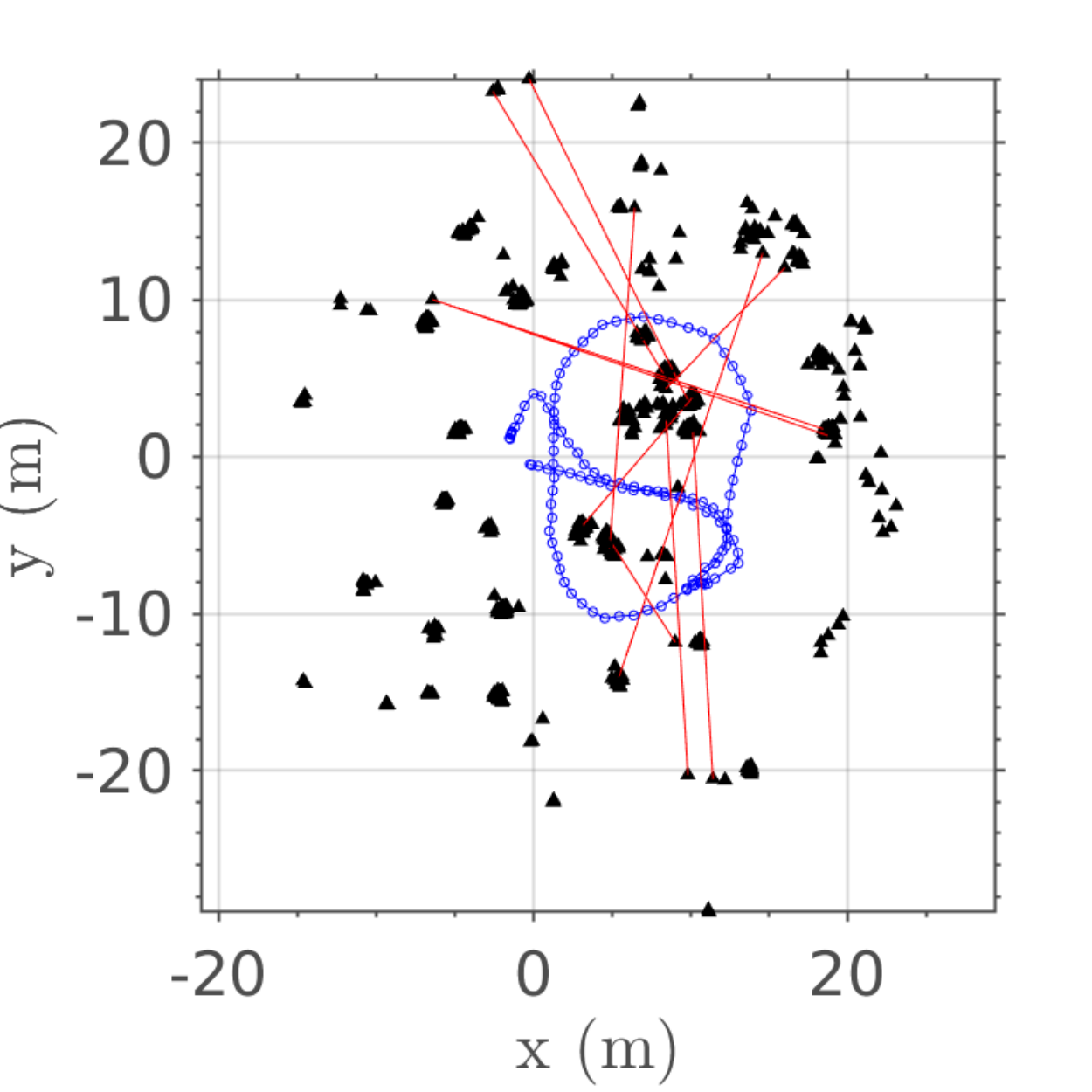}
		\caption{CLEAR }
		\label{fig:SLAM_CLEAR}
	\end{subfigure}
	~
	\begin{subfigure}{0.23\textwidth}
		\includegraphics[width=1.0\textwidth]{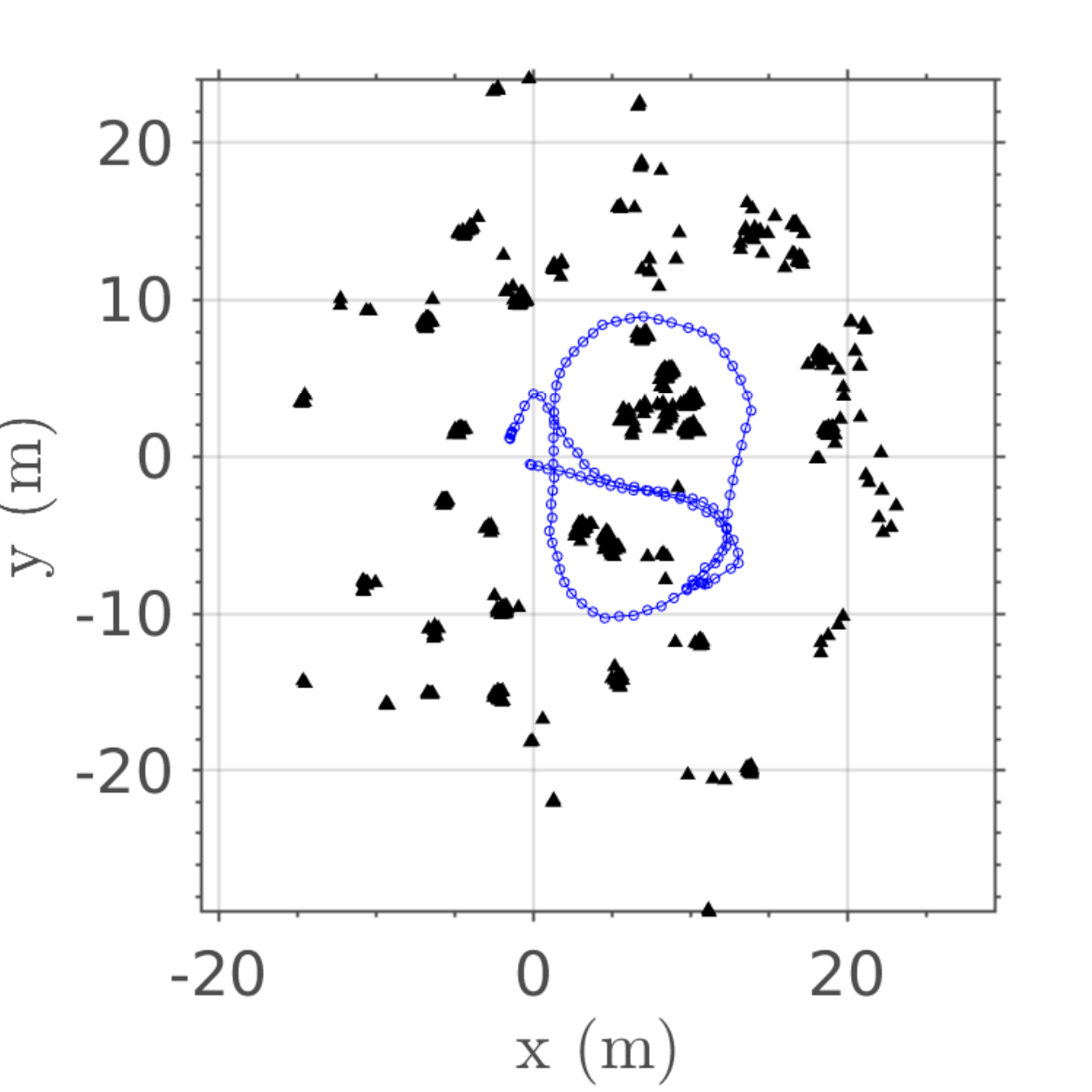}
		\caption{CLEAR (post-processed)}
		\label{fig:SLAM_CLEAR_clean}
	\end{subfigure}
	\caption{Output association of baseline (a) and each algorithm (b)-(g) in
	the landmark-based SLAM dataset. Cycle-inconsistent solutions are completed due
	the clique-centric nature of the problem (Section~\ref{sec:apps}). Each black triangle represents a single tree observation. \emph{The LIDAR-inertial odometry is shown in blue}. Definite negatives identified using the odometry estimates are highlighted as red edges.  We note that the output of CLEAR (g) still contains a few mismatches, but in practice, these can be filtered out by removing small clusters from the returned association, as shown in (h). }
	\label{fig:slam}
\end{figure*}

\begin{table}[h]
\renewcommand{\arraystretch}{1.5}
\center
\caption{Cross comparison of algorithms in terms of the number of definite nagatives (DN), potential positives (PP), upper bound on precision ($\bar{\text{P}}$), and runtime.  The  upper bound on precision is computed from \eqref{eq:Pbar_def}.}
\begin{tabular}{l c c c c}
\Xhline{1.0pt} 
Algorithm &  DN & PP & $\bar{\text{P}}$ (\%) & Runtime (s)  \\
\Xhline{1.0pt} 
CLEAR & \textbf{11} & 3393 & \textbf{99.677} & \textbf{0.084}  \\ \hline
MatchLift \cite{Chen2014} & \textbf{5} & 2394 & \textbf{99.792} & 124.7  \\ \hline
MatchALS \cite{Zhou2015} (completed) & 89 & 15230 & 99.419 & 4.580  \\ \hline
QuickMatch \cite{Tron2017} & 897 & 15757 & 94.614 & 0.118  \\ \hline
NMFSync \cite{Bernard2018} & 290 & 3233 & 91.768 & 4.272 \\ \hline
MatchEIG \cite{Maset2017} (completed) & 21415 & 20381 & 48.763 & 1.808  \\ \hline
Baseline & 26249 & \textbf{20487} & 43.836 & N/A  \\ 
\Xhline{1.0pt} 
\end{tabular}
\label{tab:slam_table}
\end{table}

\begin{figure} [t!]
	\centering
	\includegraphics[trim = 0mm 0mm 0mm 0mm, clip, width=\columnwidth] {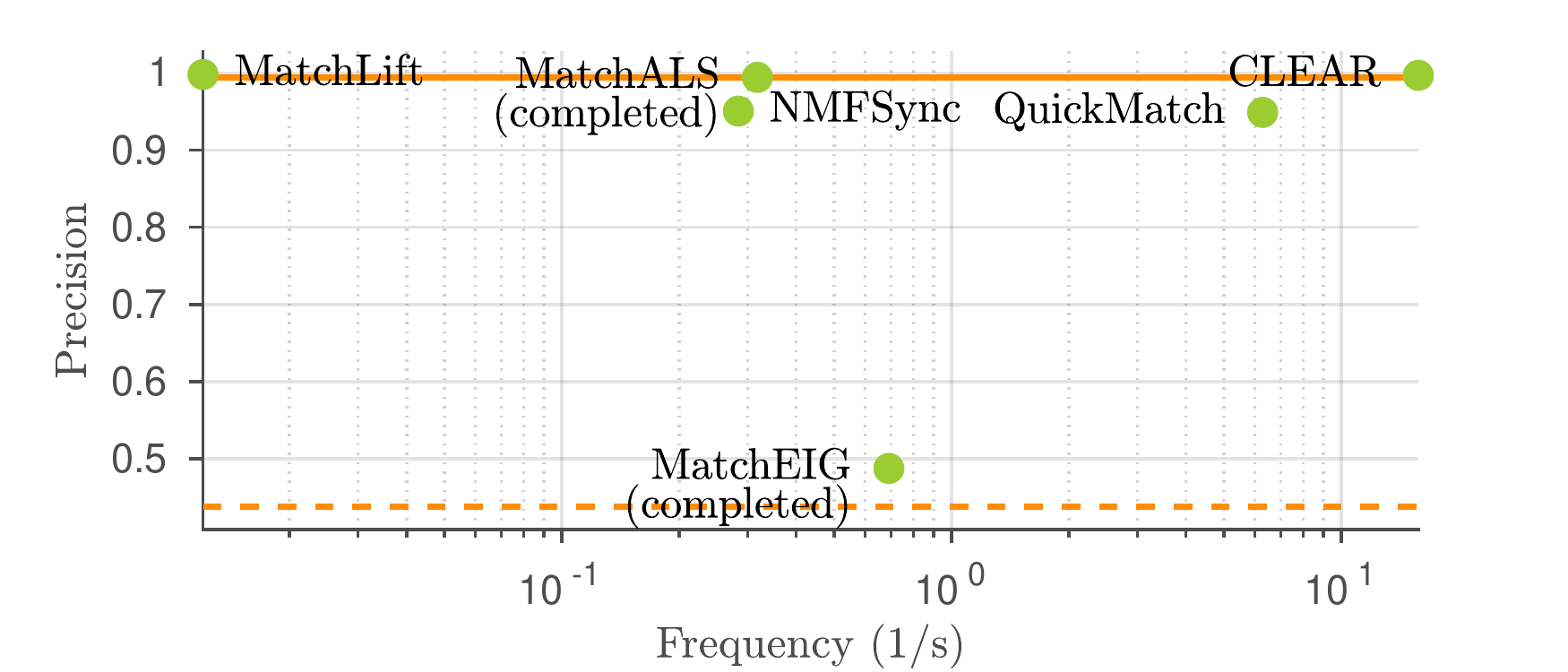}
	\caption{Precision upper bound $\bar{\text{P}}$ vs. frequency in the landmark-based SLAM dataset. The frequency axis has logarithmic scale.
	The closer to the top-right corner, the better.
	The solid orange line corresponds to the precision of input data associations.
	The dashed orange line corresponds to the precision of the baseline, obtained by completing the connected components in the input association graph.
	Further quantitative results are reported in Table~\ref{tab:slam_table}.}
	\label{fig:slam_precision_speed}
\end{figure}

Fig.~\ref{fig:slam} visualizes the data associations returned by each algorithm in the world frame.
All definite negatives are highlighted in red.
The solutions of MatchALS and MatchEIG are not cycle consistent initially, and are made cycle consistent by completing the connected components.
The solution of the spectral algorithm is omitted, as it contains significantly more mismatches due to its sensitivity in estimating the universe size.
Table~\ref{tab:slam_table} shows the complete set of quantitative results
and Fig.~\ref{fig:slam_precision_speed} shows the precision and frequency of each evaluated algorithm.

Due to the existence of mismatches in the input associations, the baseline algorithm which directly completes the connected components yields more than $25000$ definite negatives; see
Fig.~\ref{fig:slam}(a).
In contrast, most other algorithms are able to significantly reduce the number of definite negatives.
Among these algorithms, CLEAR and MatchLift nearly eliminate all definite negatives; see Table~\ref{tab:slam_table}.
However, MatchLift requires $124.7$\;s to converge while CLEAR only takes $0.084$\;s.
The superior speed of CLEAR thus makes the algorithm favorable for real-time SLAM applications.
On the other hand, we note that the output of CLEAR still contains a few definite negatives; see Fig.~\ref{fig:slam}(g). 
This is undesirable for landmark-based SLAM, as any incorrect fusion of landmarks could inflict catastrophic impact on the final SLAM solution.
In practice, these mismatches can be filtered out by removing clusters of small size from the returned solution. 
For example, Fig.~\ref{fig:slam}(h) shows the resulting association after removing clusters of size smaller than four from the output of CLEAR. After this post-processing step, the final association is accurate and can be used by any SLAM back-end to solve for the vehicle trajectory and landmark positions.

\begin{figure*}[t]
	\centering
	\begin{subfigure}{0.23\textwidth}
		\includegraphics[width=1.0\textwidth]{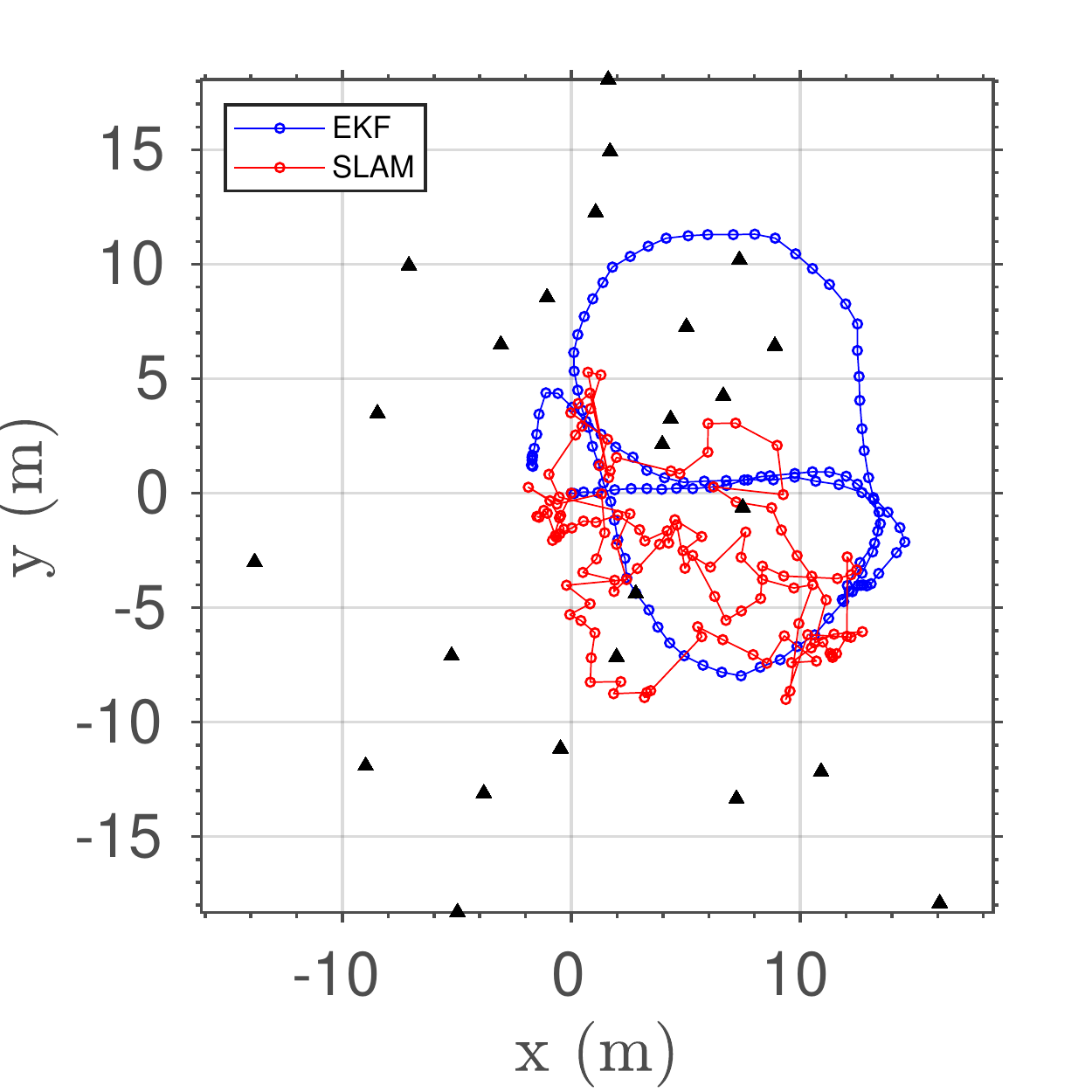}
		\caption{Baseline (post-processed)}
		\label{fig:fused_input}
	\end{subfigure}
	~
	\begin{subfigure}{0.23\textwidth}
		\includegraphics[width=1.0\textwidth]{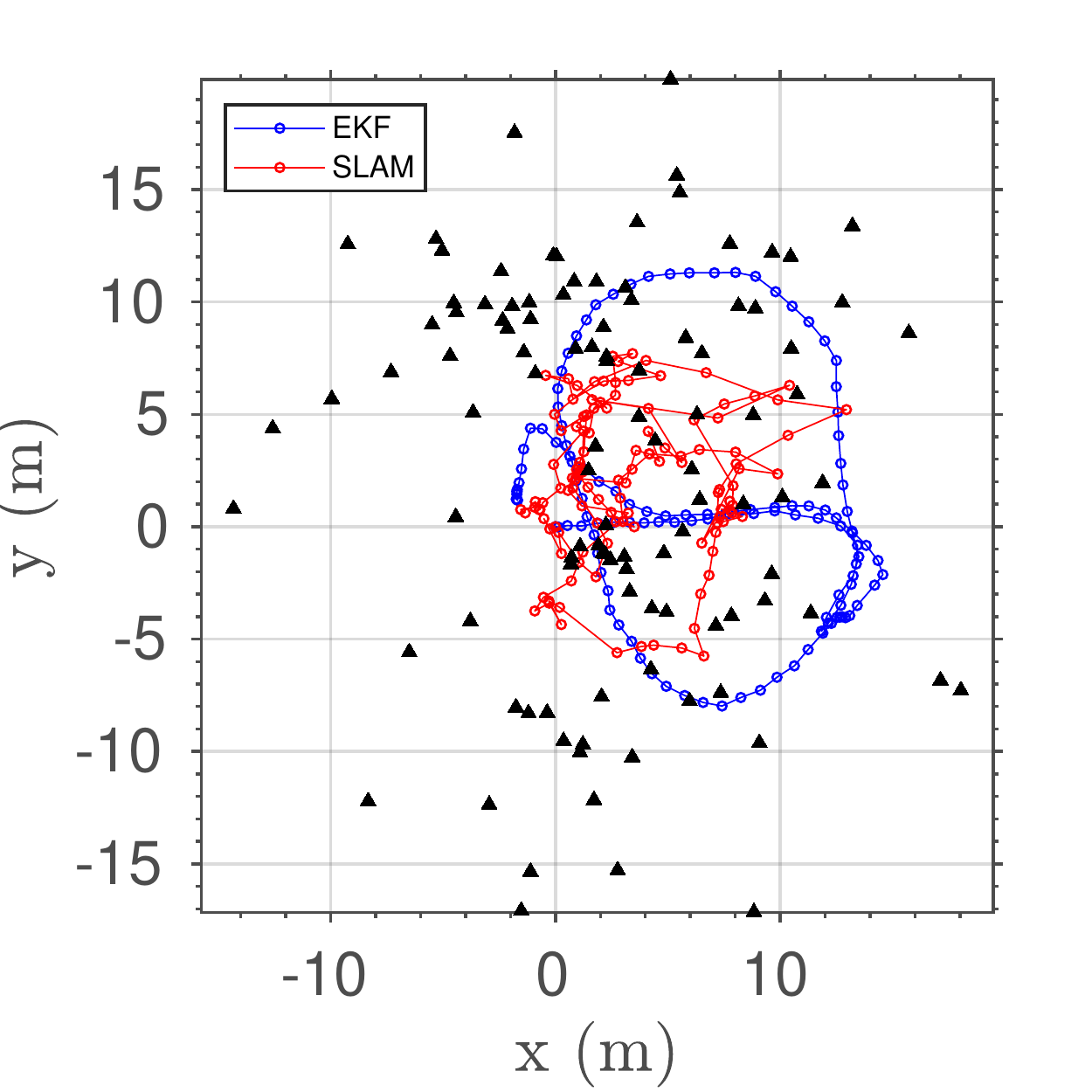}
		\caption{NMFSync (post-processed)}
		\label{fig:fused_NMF}
	\end{subfigure}
	~
	\begin{subfigure}{0.23\textwidth}
		\includegraphics[width=1.0\textwidth]{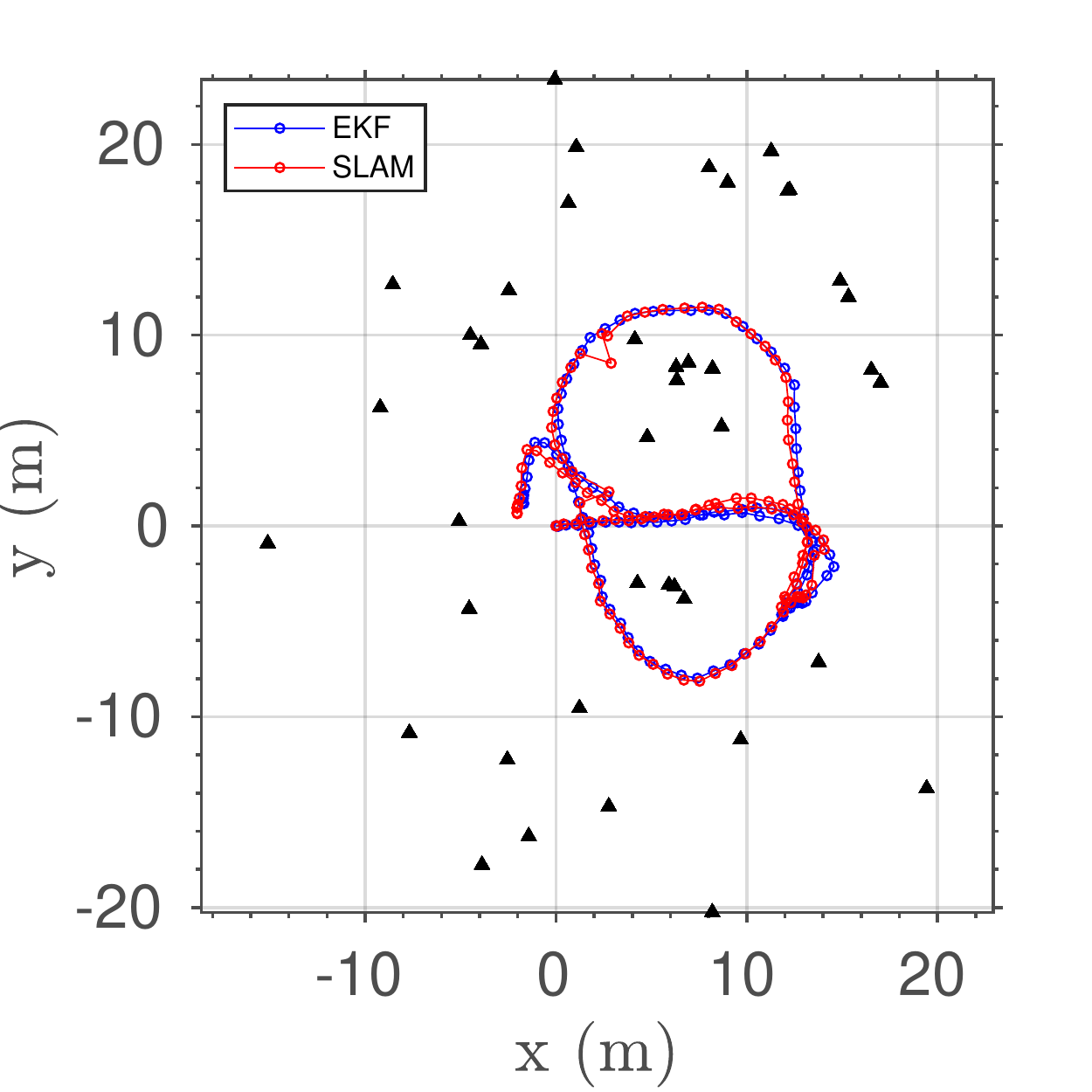}
		\caption{MatchALS (post-processed)}
	\end{subfigure}
	~
	\begin{subfigure}{0.23\textwidth}
		\includegraphics[width=1.0\textwidth]{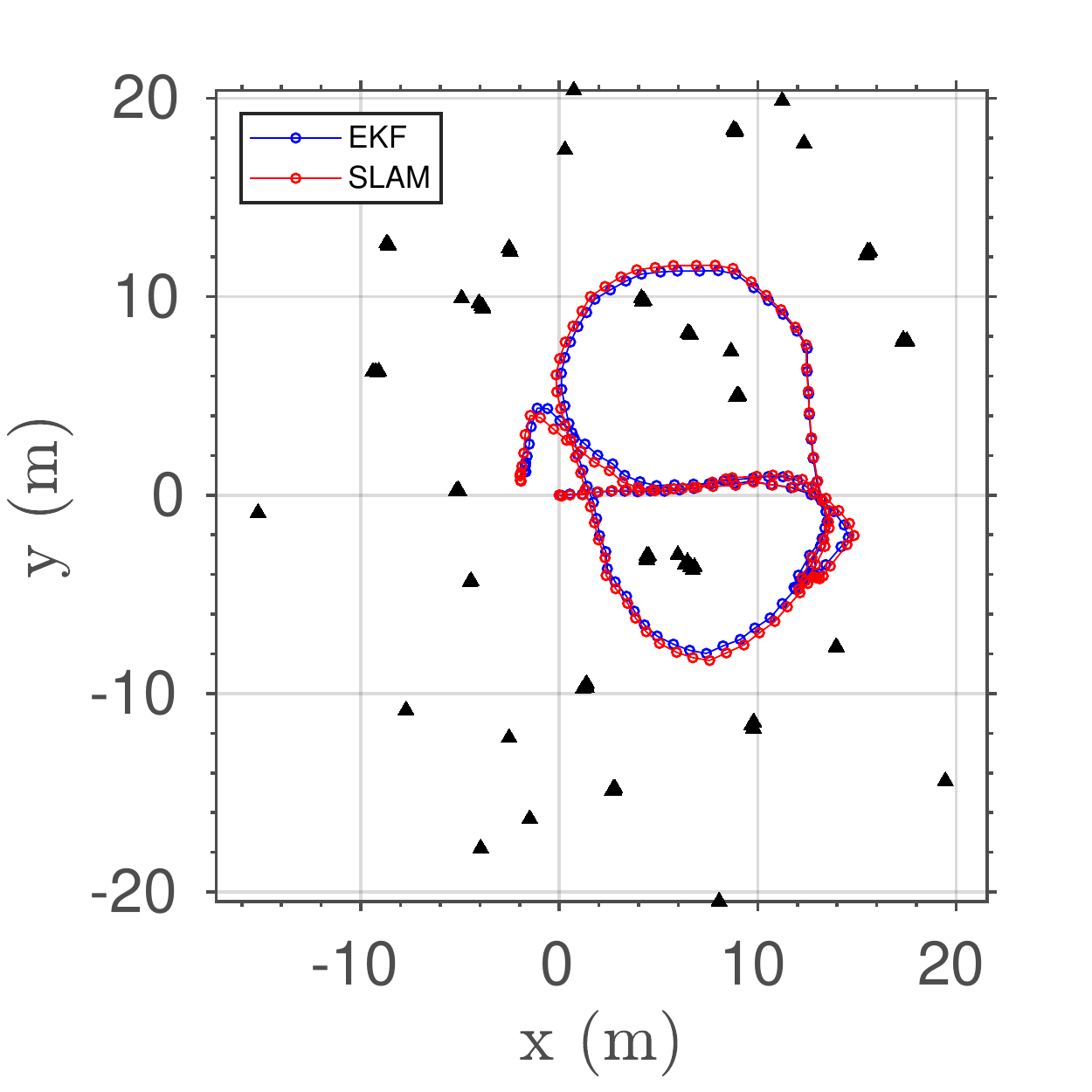}
		\caption{MatchLift (post-processed)}
	\end{subfigure}
	\\
	\begin{subfigure}{0.23\textwidth}
		\includegraphics[width=1.0\textwidth]{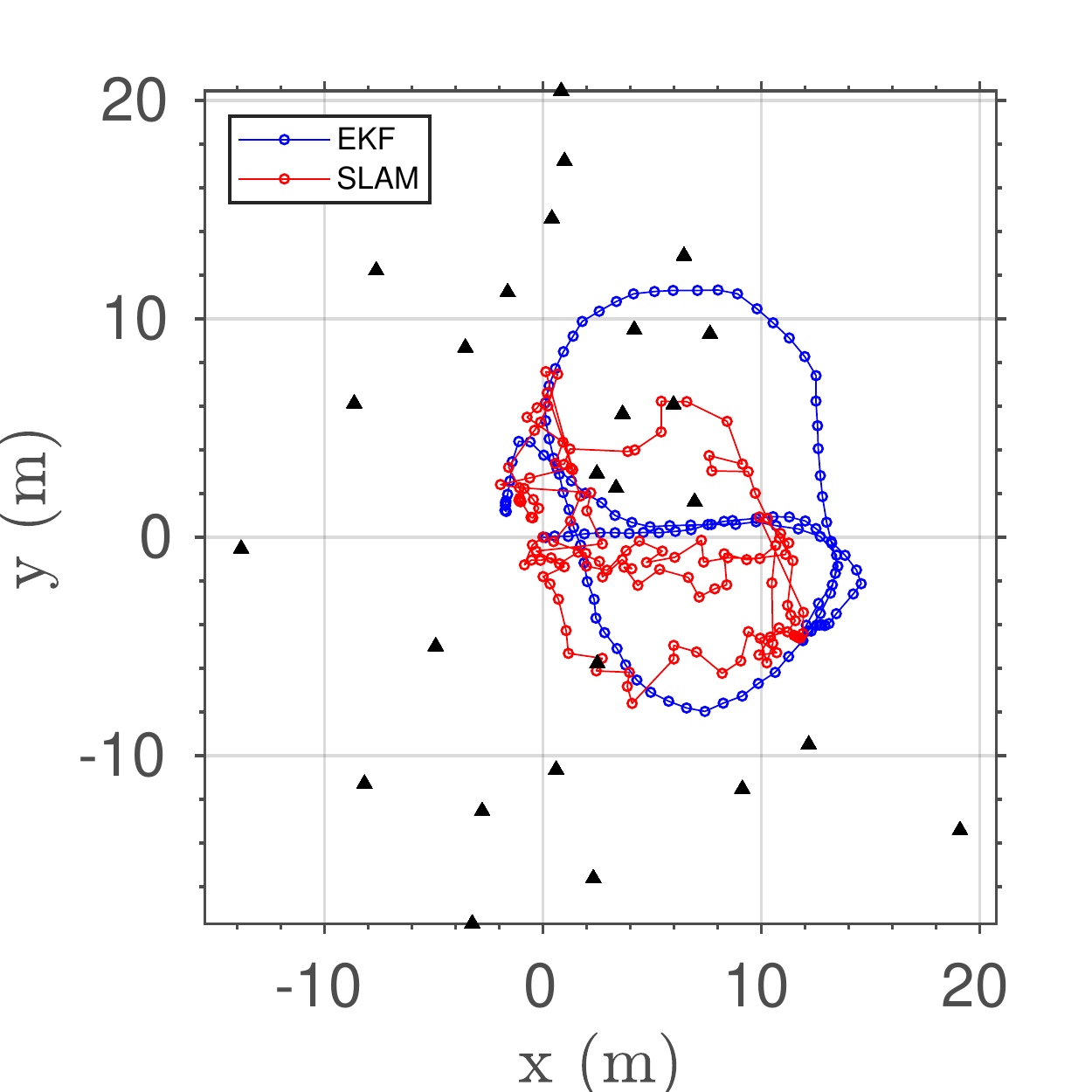}
		\caption{MatchEIG (post-processed)}
	\end{subfigure}
	~
	\begin{subfigure}{0.23\textwidth}
		\includegraphics[width=1.0\textwidth]{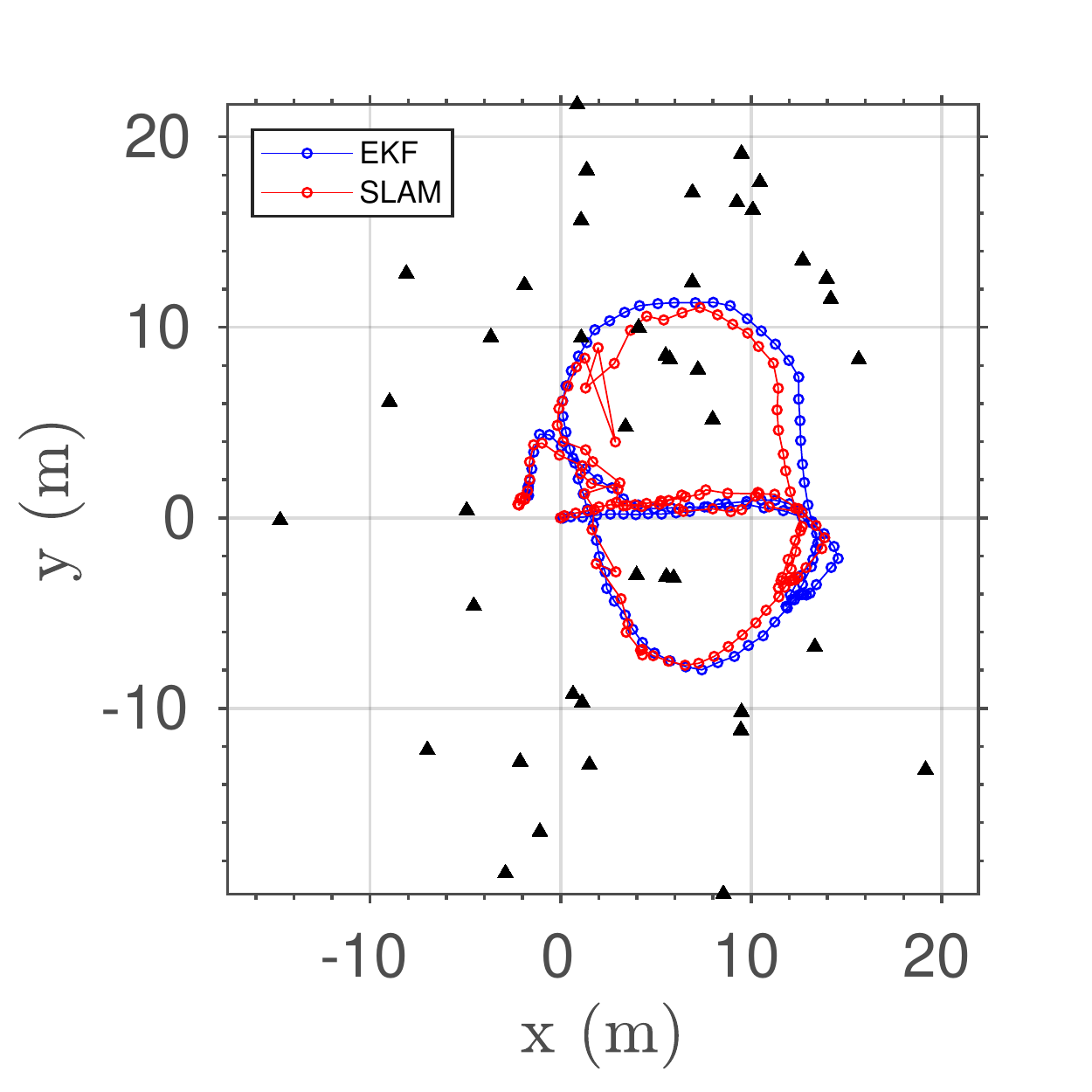}
		\caption{QuickMatch (post-processed)}
	\end{subfigure}
	~
	\begin{subfigure}{0.23\textwidth}
		\includegraphics[width=1.0\textwidth]{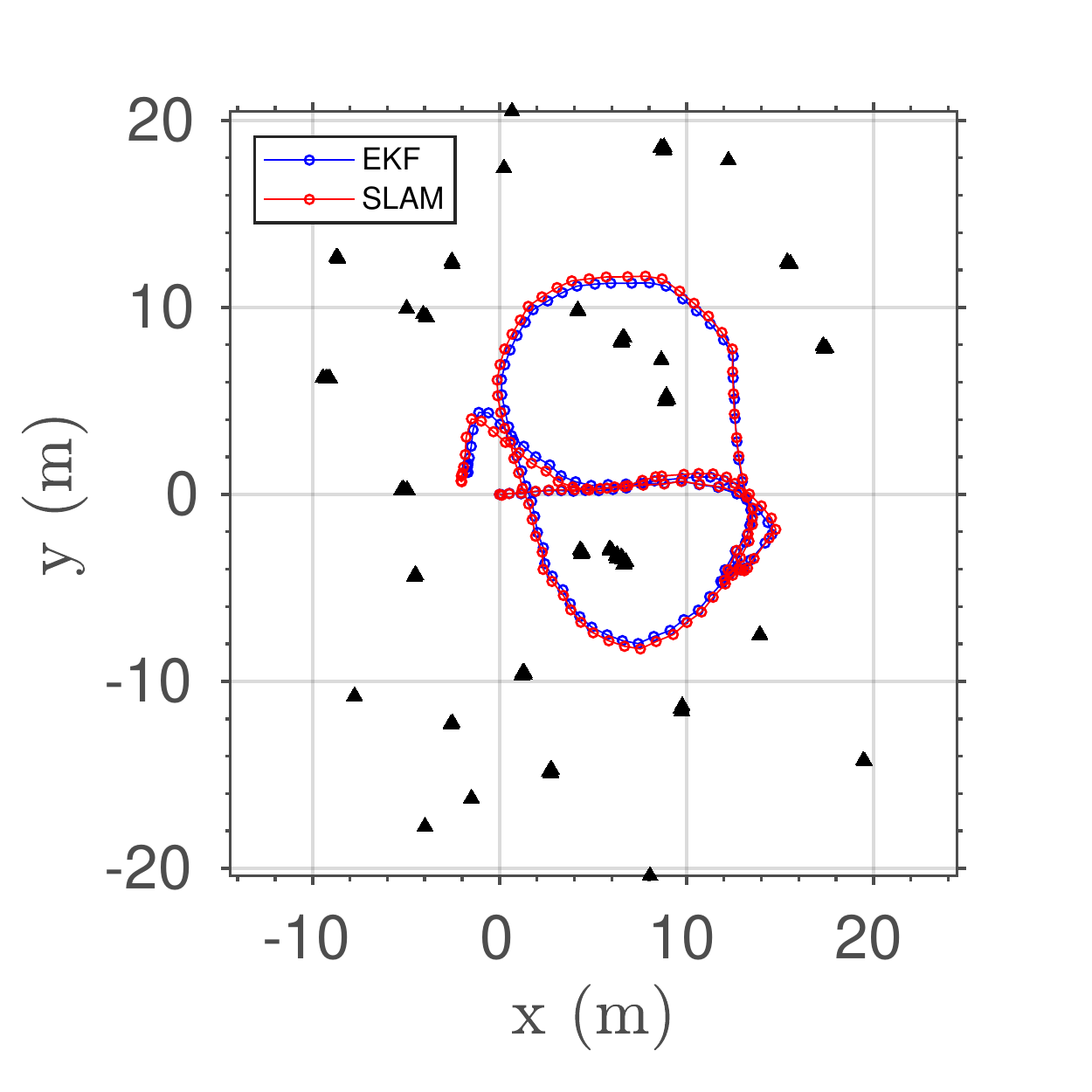}
		\caption{CLEAR (post-processed)}
		\label{fig:fused_CLEAR}
	\end{subfigure}
	\caption{Fused map after optimization with g2o. The solutions of MatchALS and MatchEIG are made cycle consistent by completing each connected components in the induced association graph. Each association is post-processed to remove any clusters of size smaller than four. Each black  triangle represents a single tree in the fused map. Trajectory estimates from EKF-based LIDAR-inertial odometry and after landmark-based SLAM are shown in blue and red, respectively. }
	\label{fig:slam_fused}
\end{figure*}

Fig.~\ref{fig:slam_fused} demonstrates the results of landmark-based SLAM
using the data association returned by each algorithm. Prior to optimization, we
apply the same post-processing procedure described earlier, by removing clusters
of size smaller than four from each data association. Subsequently, we
initialize a single tree for each remaining cluster in the fused map. All tree
positions and vehicle poses are then jointly optimized using g2o
\cite{kuemmerle2011g2o}. 
We note that Fig.~\ref{fig:slam_fused} mainly provides a qualitative comparison of the trajectory estimates.
Intuitively, we expect that SLAM trajectories that are discontinuous are likely to be wrong due to false data associations. 
These include the trajectories optimized with the baseline, NMFSync, MatchALS, MatchEIG, and QuickMatch.
While CLEAR and MatchLift produce similar results, CLEAR is more than $1000$ times faster as indicated by the results in Table~\ref{tab:slam_table}.

Finally, we note that some quantitative results presented in this section are
different from those obtained in earlier comparisons based on synthetic data.
For example, we observed that algorithms such as NMFSync perform better in
simulation. A major cause of this discrepancy is the noise model. In our
synthetic data, the input is solely corrupted by mismatches that reassign
correct matches to wrong ones. In the forest dataset, however, the input is
corrupted by both mismatches and a significant number of \emph{missing} correct
associations, thus further reducing the signal-to-noise ratio.

\section{Conclusion}

Data association across multiple views is a fundamental problem in robotic applications.
Traditionally, this problem is decomposed into a sequence of pairwise
subproblems. Multi-view matching algorithms can leverage observation
redundancy to improve the accuracy of pairwise associations. 
However, the use of these algorithms in robotic applications is often prohibited by their high computational complexity, as well as critical issues such as cycle
inconsistency and high number of mismatches which may have catastrophic
consequences.

To address these critical challenges, we presented CLEAR, an algorithm that
leverages the natural graphical representation of the multi-view association
problem. CLEAR uses a spectral graph clustering technique, which is uniquely tailored to solve this problem in a computationally efficient manner. 
Empirical results based on extensive synthetic and experimental evaluations
demonstrated that CLEAR outperforms the state-of-the-art algorithms
in terms of both accuracy and speed.  
This general framework can  provide significant improvements in the accuracy
and efficiency of data association in many applications such as
metric/semantic SLAM, multi-object tracking, and multi-view point cloud registration that traditionally rely on pairwise matchings.

\section*{Acknowledgments}

This work was supported in part by NASA Convergent Aeronautics Solutions project Design Environment for Novel Vertical Lift Vehicles (DELIVER), by ONR under BRC award N000141712072, and by ARL DCIST under Cooperative Agreement Number W911NF-17-2-0181.
We would also like to thank Florian Bernard, Kostas Daniilidis, Spyros Leonardos,  Stephen Phillips, Roberto Tron, and Xiaowei Zhou for their valuable insights that considerably improved this work.

\balance 

\bibliographystyle{IEEEtran}
\bibliography{Bibs}

\onecolumn
\appendix 


\begin{proof} [\textbf{Proof of Proposition~\ref{prop:reform}}]
Consider the optimization problem \eqref{eq:optim}. Since trace is invariant under cyclic permutations\footnote{e.g., $\tr(A\,B\,C) = \tr(B\,C\,A) = \tr(C\,A\,B)$.}, we obtain
%
\begin{subequations}
	\begin{align} 
	\underset{P = V\, V^\top}{\mathrm{max}}\, \langle P_{\text{nrm}},\,
	\tilde{P}_{\text{nrm}} \rangle & 
	= \underset{P = V\, V^\top}{\mathrm{max}}\, \tr (P_{\text{nrm}}^\top
	\tilde{P}_{\text{nrm}}) && {\color{gray} (\text{from definition of inner product } \langle \cdot, \cdot \rangle) }\\
	&= \underset{V \in \bv}{\mathrm{max}}~ \tr (C^{-\frac{1}{2}}\, V \, V^\top
	C^{-\frac{1}{2}}\, \tilde{P}_{\text{nrm}}) && {\color{gray} (\text{since }
		P_{\text{nrm}} \eqdef C^{-\frac{1}{2}}\, P\, C^{-\frac{1}{2}} \text{ and }  P = V \, V^\top) }\\
	&= \underset{V \in \bv}{\mathrm{max}}~ \tr (V^\top C^{-\frac{1}{2}}\,
	\tilde{P}_{\text{nrm}}\, C^{-\frac{1}{2}}\, V). && {\color{gray} (\text{from cyclic permutation}) }
	\label{eq:p1}
	\end{align}
\end{subequations}
%
As discussed in Section~\ref{sec:GraphFrom}, in the graph formulation of the problem, $V$ corresponds to partitions of the association graph $\sg$ into clusters $\sa_1, \dots, \sa_m$, where $(V)_{ij} = 1$ if and only if vertex $v_i \in \sa_j$.
This implies that $\sum_{i=1}^l{(V)_{ij}} = |\sa_j|$, and diagonal entries of $C$ are $c_i = |\sa_j|$ for each vertex $v_i \in \sa_j$.
Consequently, $V^\top C^{-1} V = I$.
Since solution of \eqref{eq:p1} is invariant to adding/subtracting a constant to the objective function, by subtracting $\tr(V^\top C^{-1} V) = \tr(I) = l$ from \eqref{eq:p1} and defining $U \eqdef C^{-\frac{1}{2}}\, V$ we obtain the equivalent program
%
\begin{subequations}
	\begin{align}
	\underset{V \in \bv}{\mathrm{max}}~ &  \tr (V^\top C^{-\frac{1}{2}}\,
	\tilde{P}_{\text{nrm}}\, C^{-\frac{1}{2}}\, V) - \tr(V^\top C^{-1} V)  \\
	& = \underset{V \in \bv}{\mathrm{max}}~ \tr (V^\top C^{-\frac{1}{2}}\,
	\tilde{P}_{\text{nrm}}\, C^{-\frac{1}{2}}\, V) - \tr(V^\top C^{-\frac{1}{2}}\, C^{-\frac{1}{2}}\, V) && {\color{gray} (C^{-1} = C^{-\frac{1}{2}}\, C^{-\frac{1}{2}}) }\\
	& = \underset{U \in \bu}{\mathrm{max}}~ \tr (U^\top \tilde{P}_{\text{nrm}}\, U) - \tr(U^\top  U) && {\color{gray} (\text{replacing } U \eqdef C^{-\frac{1}{2}}\, V) }\\
	& = \underset{U \in \bu}{\mathrm{max}}~ \tr (U^\top \tilde{C}^{-\frac{1}{2}} \, \tilde{P} \, \tilde{C}^{-\frac{1}{2}}  \, U) - \tr(U^\top \tilde{C}^{-\frac{1}{2}} \, \tilde{C} \, \tilde{C}^{-\frac{1}{2}} U) && {\color{gray} (\text{since } \tilde{C}^{-\frac{1}{2}} \, \tilde{C} \, \tilde{C}^{-\frac{1}{2}} = I ) }\\
	& = \underset{U \in \bu}{\mathrm{max}}~ \tr (U^\top \tilde{C}^{-\frac{1}{2}} \, (\tilde{P} - \tilde{C}) \, \tilde{C}^{-\frac{1}{2}}  \, U) && {\color{gray} (\text{by factoring terms}) }\\
	&= \underset{U \in \bu}{\mathrm{min}}~ \tr (U^\top \tilde{C}^{-\frac{1}{2}} \, \tilde{L} \, \tilde{C}^{-\frac{1}{2}}  \, U) && {\color{gray} (\text{since } \tilde{P} - \tilde{C} = -\tilde{L}) }\\
	&= \underset{U \in \bu}{\mathrm{min}}~ \tr (U^\top \, \Lnt \, U). && {\color{gray} (\text{using definition } \Lnt \eqdef\tilde{C}^{-\frac{1}{2}} \, \tilde{L} \, \tilde{C}^{-\frac{1}{2}} ) }
	\end{align}
\end{subequations}
%
From the definition $U \eqdef C^{-\frac{1}{2}}\, V$ and since $V^\top C^{-1} V = I$, it follows that $U^\top U = I$.
\end{proof}

\begin{proof}[\textbf{Proof of Lemma~\ref{lem:EigLhat}}]
The spectrum of a \textit{complete} graph with $l_i$ vertices and Laplacian $L_i \in \br^{l_i \times l_i}$ consists of eigenvalues $0$ and $l_i$, with multiplicities 1 and $l_i-1$, respectively \cite[Chap. 1]{Chung1997}. Since in this case the diagonal matrix $C_i = D_i + I$ has diagonal entries $l_i$, eigenvalues of the normalized Laplacian $C_i^{-\frac{1}{2}} \, L_i \,  C_i^{-\frac{1}{2}} = \frac{1}{l_i} L_i$ are $0$ and $1$, with multiplicities 1 and $l_i-1$, respectively.
By definition, a cluster graph is a disjoint union of complete graphs. Since spectrum of a graph is the union of its connected components' spectra \cite{Chung1997}, the conclusion follows. 
\end{proof}

\begin{proof} [\textbf{Proof of Lemma~\ref{lem:m}}]
Let $\lambda_1 \leq \lambda_2 \leq \dots \leq \lambda_l$ denote ordered eigenvalues of $\Ln$, where from Lemma~\ref{lem:EigLhat} we have $\lambda_1 = \lambda_2 =\dots = \lambda_m = 0$ and $\lambda_{m+1} = \lambda_{m+2} =\dots = \lambda_l = 1$. If $\tilde{\lambda}_1 \leq \tilde{\lambda}_2 \leq \dots \leq \tilde{\lambda}_l$ are the ordered eigenvalues of  $\Lnt = \Ln + N$, from the Weyl's eigenvalue theorem \cite{Horn2012} we have $|\tilde{\lambda}_i - \lambda_i| < \| N\|$ for all $i \in \bn_l$. This implies, if $\|N\| < 0.5$, that $\left| \{ \tilde{\lambda} \,:\,  \lambda < 0.5  \} \right| = m$, which shows the correct number of clusters is recovered. 
\end{proof}

\begin{proof} [\textbf{Proof of Proposition~\ref{prop:m}}]
We have
\begin{subequations}
\begin{align} 
\|\Lnt - \Ln \| &
= \| \tilde{C}^{-\frac{1}{2}}\, \tilde{L} \, \tilde{C}^{-\frac{1}{2}} - C^{-\frac{1}{2}}\,  L \, C^{-\frac{1}{2}} \|  && {\color{gray} (\text{from definitions of } \Lnt, \Ln)} \\
&= \| C^{-\frac{1}{2}}\, (\tilde{L} - L ) \, C^{-\frac{1}{2}} \|  && {\color{gray} (\text{since by assumption } \tilde{C} = C)} \\
&\leq \| C^{-1} \| \, \| \tilde{L} - L  \| && {\color{gray} (\text{since 2-norm is submultiplicative})} \\
&=  \| C^{-1} \| \, \| (\tilde{D} - \tilde{A}) - (D - A)  \| && {\color{gray} (\text{since } L \eqdef D - A)} \\
&= \| C^{-1} \| \, \| \tilde{A} - A  \|  && {\color{gray} (\text{since } \tilde{C} = C \text{ and } D = C - I)} \\
&= \| C^{-1} \| \, \| E  \| && {\color{gray} (\text{since } E \eqdef \tilde{A} - A)}\\
&\leq \frac{1}{c_\text{min}} \| E  \| && {\color{gray} (\text{since } C \text{ is diagonal}) }\\
&\leq e_{\text{max}}/c_{\text{min}}, && {\color{gray} (\text{since } \| E  \|
\leq e_{\text{max}}) }
\end{align}
\end{subequations}
%
where the last inequality follows from the Gershgorin circle theorem \cite[Sec. 6.1]{Horn2012}.
The conclusion follows from Lemma~\ref{lem:m} and observing that $ \|\Lnt - \Ln
\| = \| N \| < 0.5$ implies $e_\text{max} < 0.5 \, c_\text{min}$.
\end{proof}

\end{document}